\pdfoutput=1

\documentclass[format=acmsmall, nonacm]{acmart}

\usepackage{
    adjustbox,
    algorithm,
    amsfonts,
    amsmath,
    amsthm,
    booktabs,
    colortbl,
    diagbox,
    etoolbox,
    footnote,
    graphics,
    hyperref,
    makecell,
    microtype,
    multirow,
    nicefrac,
    pgfplots,
    pifont,
    schemabloc,
    setspace,
    siunitx,
    soul,
    subcaption,
    threeparttable,
    tikz,
    url,
    wrapfig,
    xcolor,
}

\usepackage[noend]{algpseudocode}
\usepackage[shortcuts]{extdash}
\usepackage[inline]{enumitem}

\usetikzlibrary{arrows,shapes,circuits.logic.US,shapes.gates.logic.IEC,calc,decorations.markings,patterns,matrix,shapes.misc, backgrounds}

\usepgfplotslibrary{groupplots}

\definecolor{commentgreen}{RGB}{2,112,10}
\definecolor{eminence}{RGB}{108,48,130}
\definecolor{weborange}{RGB}{255,165,0}
\definecolor{frenchplum}{RGB}{129,20,83}
\usepackage{listings}
\lstdefinelanguage{MLIR}{
    keywords=[1]{
        to, step, iter_args, init, async, deps, id, scf.for, scf.for_all, scf.yield, scf.parallel, memref.dealloc, memref.alloc, memref.subview, memref.load, memref.store, memref.copy, air.execute, air.execute_terminator, air.channel.put, air.channel.get, air.herd, air.segment, air.launch, air.wait_all, air.memcpy, arith.constant, arith.addf, arith.addi, arith.mulf, arith.muli, func.call, func.func
    },
    keywordstyle=[1]{\color{blue}},
    emph={memref, arith, scf, index, i32, f32, !air.async.token},
    emphstyle=\color{eminence},
    sensitive=false, 
    numbers=left,
    xleftmargin=2em,
    morecomment=[l]{//}, 
    morecomment=[s]{/*}{*/}, 
    commentstyle=\color{commentgreen},
    morestring=[b]", 
    moredelim=**[is][\color{red}]{?}{?},
    moredelim=**[is][\color{black!25!blue}]{|}{|},
    moredelim=**[is][\color{gray}]{^}{^},
    moredelim=**[s][\color{gray}]{<}{>},
    alsoletter={.!},
} %

\newcommand{\eddie}[1]{\textcolor{orange}{[EDDIE: #1]}}
\newcommand{\erika}[1]{\textcolor{violet}{[ERIKA: #1]}}
\newcommand{\zhewen}[1]{\textcolor{blue}{[ZHEWEN: #1]}}
\newcommand{\sangeeta}[1]{\textcolor{red}{[SANGEETA: #1]}}
\newcommand{\jz}[1]{\textcolor{teal}{[JINMING: #1]}}

\usepackage{tikz}
\definecolor{ddgreen}{rgb}{0.00, 0.50, 0.00}
\newcommand{\gs}[1]{{\color{ddgreen}{#1}}}
\newcommand{\gsf}[1]{{\color{ddgreen}\textbf{\textit{GAGAN: #1}}}}
\usepackage{xspace}
\newcommand{\tech}{MLIR-AIR\xspace}
\newcommand{\air}{AIR\xspace}
\newcommand{\head}[1]{{\noindent\textbf{#1.}\xspace}} 
\newcommand{\launch}{\texttt{\color{gray} air.launch}\xspace}

\newcommand{\herd}{\texttt{\color{gray} air.herd}\xspace}

\newcommand{\segment}{\texttt{\color{gray} air.segment}\xspace}
\newcommand{\segments}{\texttt{\color{gray} air.segment} operations\xspace}
\newcommand{\channel}{\texttt{\color{gray} air.channel}\xspace}
\newcommand{\channels}{\texttt{\color{gray} air.channel} operations\xspace}
\newcommand{\token}{\texttt{\color{gray} air.token}\xspace}

\newcommand*\circled[1]{\tikz[baseline=(char.base)]{
            \node[shape=circle,draw,inner sep=0pt,fill=black, text=white] (char) {#1};}}
	\makeatletter
	\g@addto@macro{\normalsize}{%
	  \setlength{\abovedisplayskip}{2pt plus 1pt minus 1pt}
	  \setlength{\belowdisplayskip}{2pt plus 1pt minus 1pt}
	  \setlength{\abovedisplayshortskip}{0pt}
	  \setlength{\belowdisplayshortskip}{0pt}
	  \setlength{\intextsep}{2pt plus 1pt minus 1pt}
	  \setlength{\textfloatsep}{4pt plus 1pt minus 1pt}
	  \setlength{\skip\footins}{5pt plus 1pt minus 1pt}}
	\makeatother
	\usepackage{titlesec}

\titlespacing\section{0pt}{3pt plus 2pt minus 2pt}{0pt plus 2pt minus 2pt}
\titlespacing\subsection{0pt}{3pt plus 1pt minus 1pt}{0pt plus 1pt minus 1pt}

\newcommand{\memref}{MemRef\xspace}
\newcommand{\affine}{Affine\xspace}
\newcommand{\linalg}{LinAlg\xspace}
\newcommand{\scf}{SCF\xspace}
\newcommand{\memcpy}{\texttt{\color{gray} air.memcpy}\xspace}

\newcommand{\scffor}{\texttt{\color{gray} scf.for}\xspace}
\newcommand{\scfpar}{\texttt{\color{gray} scf.parallel}\xspace}
\newcommand{\scfforall}{\texttt{\color{gray} scf.for\_all}\xspace}
\newcommand{\channelput}{\texttt{\color{gray} air.channel.put}\xspace}
\newcommand{\channelget}{\texttt{\color{gray} air.channel.get}\xspace}
\newcommand{\waitall}{\texttt{\color{gray} air.wait\_all}\xspace}
\newcommand{\memrefcopy}{\texttt{\color{gray} memref.copy}\xspace}
\newcommand{\affineif}{\texttt{\color{gray} affine.if}\xspace}

\newcommand*\coretile[0]{\tikz[anchor=base, baseline]{
            \node[shape=rectangle, pattern=north west lines, pattern color=black!10!green, rounded corners=1mm, minimum height=1em, minimum width=1em, draw] at (0em,0.3em) {};}}
\newcommand*\coremem[0]{\tikz[anchor=base, baseline]{
            \node[shape=rectangle, pattern=north east lines, pattern color=black!20!yellow, rounded corners=1mm, minimum height=1em, minimum width=1em, draw] at (0em,0.3em) {};}}
\definecolor{ballblue}{rgb}{0.13, 0.67, 0.8}
\newcommand*\memtile[0]{\tikz[anchor=base, baseline]{
            \node[shape=rectangle, pattern=vertical lines, pattern color=ballblue, rounded corners=1mm, minimum height=1em, minimum width=1em, draw] at (0em,0.3em) {};}}
\definecolor{cadetgrey}{rgb}{0.57, 0.64, 0.69}
\newcommand*\shimtile[0]{\tikz[anchor=base, baseline]{
            \node[shape=rectangle, pattern=horizontal lines, pattern color=cadetgrey, rounded corners=1mm, minimum height=1em, minimum width=1em, draw] at (0em,0.3em) {};}}

\newcommand\yup{\ding{52}}
\newcommand\nope{\ding{56}}

\usepackage{todonotes}
\newcommand\ew[1]{\todo[inline,backgroundcolor=green!25]{EW: #1}}

\soulregister\ref{7}
\soulregister\subref{7}
\soulregister\cite{7}

\begin{document}

\title[From Loop Nests to Silicon: Mapping AI Workloads onto AMD NPUs with MLIR-AIR]{From Loop Nests to Silicon: Mapping AI Workloads onto AMD NPUs with MLIR-AIR}

\author{Erwei Wang}
\email{erwei.wang@amd.com}
\author{Samuel Bayliss}
\email{samuel.bayliss@amd.com}
\author{Andra Bisca}
\email{andra.bisca@amd.com}
\author{Zachary Blair}
\email{zachary.blair@amd.com}
\author{Sangeeta Chowdhary}
\email{sangeeta.chowdhary@amd.com}
\author{Kristof Denolf}
\email{kristof.denolf@amd.com}
\author{Jeff Fifield}
\email{jeff.fifield@amd.com}
\author{Brandon Freiberger}
\email{brandon.freiberger@amd.com}
\author{Erika Hunhoff}
\email{erika.hunhoff@amd.com}
\author{Phil James-Roxby}
\email{phil.james-roxby@amd.com}
\author{Jack Lo}
\email{jack.lo@amd.com}
\author{Joseph Melber}
\email{joseph.melber@amd.com}
\author{Stephen Neuendorffer}
\email{stephen.neuendorffer@amd.com}
\author{{Eddie~Richter}}
\email{eddie.richter@amd.com}
\author{Andr\'e R\"osti}
\email{andre.rosti@amd.com}
\author{Javier Setoain}
\email{javier.setoain@amd.com}
\author{Gagandeep Singh}
\email{gagandeep.singh@amd.com}
\author{Endri Taka}
\email{endri.taka@utexas.edu}
\author{Pranathi Vasireddy}
\email{pranathi.vasireddy@amd.com}
\author{Zhewen Yu}
\email{zhewen.yu@amd.com}
\author{Niansong Zhang}
\email{nz264@cornell.edu}
\author{Jinming Zhuang}
\email{jinming_zhuang@brown.edu}
\affiliation{%
    \institution{Research and Advanced Development, AMD}
    \city{San Jose}
    \streetaddress{Logic Drive}
    \postcode{95124}
    \country{USA}
}

\renewcommand{\shortauthors}{Wang et al.}

\begin{abstract}

General-purpose compilers abstract away parallelism, locality, and synchronization, limiting their effectiveness on modern spatial architectures. As modern computing architectures increasingly rely on fine-grained control over data movement, execution order, and compute placement for performance, compiler infrastructure must provide explicit mechanisms for orchestrating compute and data to fully exploit such architectures.
We introduce \tech, a novel,  open-source compiler stack built on MLIR that bridges the semantic gap between high-level workloads and fine-grained spatial architectures such as AMD's NPUs. \tech defines the \air dialect, which provides structured representations for asynchronous and hierarchical operations across compute and memory resources. AIR primitives allow the compiler to orchestrate spatial scheduling, distribute computation across hardware regions, and overlap communication with computation without relying on \emph{ad hoc} runtime coordination or manual scheduling.
We demonstrate \tech's capabilities through two case studies: matrix multiplication and the multi-head attention block from the LLaMA 2 model. 
For matrix multiplication, \tech achieves up to 78.7\% compute efficiency and generates implementations with performance almost identical to state-of-the-art, hand-optimized matrix multiplication written using the lower-level, close-to-metal MLIR-AIE framework.
For multi-head attention, we demonstrate that the AIR interface supports fused implementations using approximately 150 lines of code, enabling tractable expression of complex workloads with efficient mapping to spatial hardware.
\tech transforms high-level structured control flow into spatial programs that efficiently utilize the compute fabric and memory hierarchy of an NPU, leveraging asynchronous execution, tiling, and communication overlap through compiler-managed scheduling.

\end{abstract}

\maketitle

\section{Introduction}
\label{sec:intro}

Modern computing architectures are increasingly spatial and asynchronous. They consist of many distributed compute units, partitioned memory hierarchies, and message-passing interconnects. Achieving high performance on such architectures requires precise control over computation, data movement, and execution of tasks.

Mainstream CPUs and GPUs rely on a thread-centric parallel compute model that assumes many threads will be scheduled onto a limited set of hardware resources. Programmers describe large numbers of threads, and the system (hardware for GPUs, software for CPUs) maps them to available compute units. This model has scaled effectively for decades as semiconductor technology has delivered more cores and vector units. Hardware support for more concurrent threads improves throughput and reduces compute latency.



The effectiveness of the thread-centric model depends on two assumptions: (1) that threads operate independently, and (2) that shared resources, particularly memory bandwidth, scale with compute. When these assumptions break down, due to synchronization, data dependencies, or resource contention, execution stalls. Hardware schedulers, particularly in GPUs, respond by context-switching to another group of threads (wavefronts) to maintain forward progress. However, GPUs do not guarantee \emph{independent} forward progress across all threads. Their schedulers may allocate compute resources to a subset of runnable wavefronts, while deferring others indefinitely. This behavior introduces nondeterminism and limits software visibility into execution dynamics, an increasingly critical shortcoming for latency-sensitive or tightly coupled workloads. Despite these limitations, the thread-centric model remains widely adopted because it simplifies software development: applications are decomposed into independent tasks that communicate through shared memory, while the hardware handles data reuse and execution interleaving. However, this abstraction comes at a significant cost. Maintaining the illusion of shared memory and uniform execution requires dense interconnects, deep cache hierarchies, and complex runtime mechanisms, all of which consume energy, silicon area, and increase design complexity.

An emerging alternative is to return control to the software. A programming model that enables explicit expression of task placement, scheduling order, and inter-task data sharing allows software to better exploit spatial and temporal locality. Rather than relying on implicit reuse via caches, such a model supports deliberate coordination between compute units that are scheduled close in space and time, reducing hardware overhead while improving predictability and efficiency.

To this end, we introduce AIR,  a compiler intermediate representation (IR) that exposes spatial and temporal execution structure as explicit, programmable constructs.
\air captures high level user-described data-movement and compute scheduling intent, including concurrent execution. Implemented as a multi-level intermediate representation (MLIR)~\cite{mlir} dialect, \air bridges the gap between high-level programs and spatial architectures. It supports transformations that lower structured control flow into statically scheduled spatial programs, optimized for GPUs and domain-specific neural processing units (NPUs). We demonstrate \air's effectiveness on two representative AI workloads: matrix multiplication and the multi-head attention (MHA) block from the LLaMA 2 model~\cite{LLAMA2}. Our results demonstrate that \air produces spatially distributed schedules that overlap communication with computation, exploit locality, and minimize runtime control overhead. 

\subsection{Contributions}
The rapid advance of artificial intelligence (AI) models, algorithms, and accelerators has driven the adoption of diverse programming tools. Some tools focus on end-user productivity, while others are aimed at optimizing the efficient implementation of AI applications on an increasingly diverse range of specialized accelerators. MLIR is a flexible compiler abstraction designed to bridge this gap by allowing progressive lowering of designs through an extensible set of dialects~\cite{lattner2020mlircompilerinfrastructureend}. Users can compose operations from a range of dialects and, in general, select transformations to achieve the goal of lowering high-level programmer intent to low-level optimized implementation\footnote{In some applications, MLIR is used to analyze and \emph{raise} the abstraction of operations, rather than lower them for execution.}. 

\air is an MLIR dialect that contains operations to express compute scheduling and memory allocation in spatial architectures. It operates at a level of abstraction that enables portable expression of compute kernels by avoiding explicit support for vendor-specific features. Cross-generational portability and performance scalability are supported by splitting the responsibilities for scheduling compute tasks between the compiler and runtime. This enables the compiler to define tightly coupled and concurrent \textit{herds} of execution, while giving the runtime flexibility to schedule those herds on devices whose sizes vary within and across generations of accelerator hardware. 

A design expressed using the \air dialect can use a vendor-specific lowering for implementation on an accelerator as the operations included 
in \tech are intended to support common features that we observe emerging in a class of spatial hardware accelerators. Programmers or compilers can use \air to express compute groupings and data allocations spatially, and see those decisions honored in the subsequent lowering to vendor-specific implementations. 



In sum, this work makes the following key contributions:
\begin{itemize}
    \item We present \air, a new IR implemented as an MLIR dialect that exposes spatial and temporal structure in programs. \air enables the compiler to coordinate computation, data movement, and synchronization --- capabilities that traditional thread-centric models obscure or defer to hardware.
    \air is developed as a set of spatially aware abstractions that enable lowerings from high-level programs to tiled spatial hardware. \air models spatial partitioning with \herd, point-to-point communication with \channel, and explicit synchronization with \token. These abstractions enable the compiler to control spatial execution, without compelling the user to drop down to lower-levels of vendor-specific abstractions. 
    \item We build a complete end-to-end compiler flow that uses AIR to lower workloads written using high-level Python frameworks to low-level code for AMD NPUs. \tech compiles structured loop nests into efficient, spatial programs dispatched using the NPU runtime.\footnote{\url{https://github.com/Xilinx/mlir-air}} 
    \item We demonstrate \tech's effectiveness on two representative AI workloads. \tech produces statically scheduled programs that exploit locality, parallelism, and pipelining on tiled hardware.
\end{itemize}

\tech is open source and modular by design. It integrates into, and composes with other dialects with the MLIR ecosystem and provides a foundation for targeting a wide range of spatial accelerators beyond AMD's NPU.

\section{Background}
\label{sec:related_works}

This section surveys recent trends in spatial hardware that inform the architectural design of modern accelerators, which motivate key requirements on modern compilers.

\subsection{Trends in Spatial Hardware}
\label{subsec:trendsinspatialhardware}
In Table~\ref{tab:spatialhardwarecharacteristics}, we describe six key trends in efficient compute hardware. Taken together, these trends define a general direction in parallel hardware design, where efficient \emph{data movement} is the driving design philosophy. Control over \emph{where} compute operations are dispatched and \emph{where} data is allocated are fundamental in such systems. Emphasizing the importance of physical placement in these systems, we refer to this direction in hardware design as a movement towards \textit{spatial hardware}. 
These six hardware trends collectively motivate a corresponding set of \emph{compiler} features necessary to effectively target spatial hardware.

\begin{table*}
    \centering
    \begin{threeparttable}
        \begin{tabular}{ll}
            \toprule
            \textbf{Trend} & \textbf{Description} \\
            \midrule
            Complex System Hierarchy & Design reuse introduces arbitrary boundaries in systems. \\
            \midrule
            Dispatch Placement & Schedulers guaranteeing locality enable resource-sharing. \\
            \midrule
            Multi-root Memory Hierarchy & \makecell[l]{Devices have independent, physically distant memory\\channels.}\\
            \midrule
            Peer Memory Movement & \makecell[l]{Efficient designs are not limited to data transfer through main\\memory.} \\
            \midrule
            Data Movement Offload & Specialized DMAs coordinate efficient data movement. \\
            \midrule
            Asynchronous Execution & \makecell[l]{Distinct hardware scheduled to execute independently via\\dependencies.}\\
            \bottomrule
        \end{tabular}
    \end{threeparttable}

    \caption{\label{tab:spatialhardwarecharacteristics} Six hardware trends of spatial architectures.}
\end{table*}

\subsubsection{Complex System Hierarchy}
Design reuse is extensive in semiconductor manufacturing because of the high cost of verification. Larger chip designs are often composed of multiple chiplets, that may themselves be composed of pre-verified hardware building blocks. Non-uniform performance for similar workloads can occur if a workload is assigned resources that cross spatial or hierarchical boundaries, or if the workload uses components shared at a cluster level. Interactions between spatially arranged components can be positive (e.g., components within a cluster share a level of cache hierarchy) or negative (e.g., components arbitrate for access to a limited number of ports). In order to maximize performance and minimize negative interactions, compilers and schedulers must be aware of spatial boundaries within the chip. 

\subsubsection{Dispatch Placement} 

Compilers and schedulers share control of decisions over placement within a spatial architecture. As such, in order to holistically optimize placement, both compilers and runtimes must be able to control or query where a scheduler allocates compute or where a memory allocator places memory. This knowledge or control would allow a compiler optimized for spatial architectures to note the desired spatial affinity of dispatched compute elements as optional or mandatory constraints on the behavior of the runtime scheduler. 

\subsubsection{Multi-root Memory Hierarchy}

Traditional compilers treat main memory as a unifying single \textit{root} of coherency. However, many modern devices use multiple independent memory channels to increase aggregate bandwidth. The transparent hardware-based interleaving of data across these channels offers one simple mechanism for accessing this bandwidth, but in a large device, it is likely that there is a non-uniform energy and latency cost for access to these separate memory channels. These NUMA effects have previously been observed in large multi-socket CPU systems, but compilers and runtimes now have a role to play in ensuring physical affinity between memory allocation within channels and compute scheduling even within a single package. 

\subsubsection{Peer Memory Movement} 

CPUs and GPUs incorporate large amounts of on-chip SRAM memory that is used as caches and/or scratchpads. Data transfer between on-chip memories can occur implicitly in the case of coherent caches, or may be explicitly orchestrated. Effective use of on-chip memories can offer lower interconnect energy, and achieve higher realized bandwidth compared to when data is fetched multiple times from external memory. 

\subsubsection{Data Movement Offload} 

GPUs and NPUs increasingly feature Direct Memory Access (DMA) engines capable of offloading complex address generation from the compute datapath to improve data movement efficiency. This enables efficient pipelined use of the interconnect fabric as well as in-line reshaping and transposition of data for efficient computation.

\subsubsection{Asynchronous Execution}
Memory and communication operations often have considerable latency. To achieve the most efficient performance, independent actors in the system (e.g., DMAs, compute units, etc.) are kept busy, using techniques to avoid stalling during the round-trip time necessary to synchronize two concurrent components. Increasingly sophisticated hardware schedulers close to those actors interpret explicitly encoded dependencies and select the next suitable thread of work for the actor to perform. 




\subsection{Identified Needs in Compilers}

Taken together, these trends in hardware construction motivate a desire for a software model that enables user control over scheduling and memory allocation. Specifically, we see a need for a framework that:

\begin{enumerate}
    \item \emph{Exposes hardware controls over memory allocation}, allowing users to allocate memory in different levels of the memory hierarchy and in different non-uniform memory access (NUMA) domains at each level of the memory hierarchy. 
    \item \emph{Exposes hardware controls over compute placement}, enabling users to describe units of compute that should be scheduled concurrently, enabling tightly-coupled compute elements to optimize data-sharing and local synchronization. 
    \item \emph{Separates data movement from computation explicitly in the IR}, enabling independent scheduling and optimization of each. This decoupling allows the compiler to overlap communication with computation, and apply architecture-specific optimizations when supported by hardware.
    \item \emph{Enables dependency resolution close to hardware} to minimize the time taken to observe completion of a predecessor operation. This can be achieved by expressing dependencies explicitly, and supporting lowerings that target platform-specific synchronization capabilities. 

\end{enumerate}

\section{Related Work}
\label{sec:related_works}

The past decade has seen rapid evolution in accelerator architectures for machine learning. Many of these accelerators share the key characteristics outlined in Section~\ref{subsec:trendsinspatialhardware}: explicit spatial compute and memory hierarchies, high-throughput interconnects, and programmable DMA subsystems. Examples include Google's TPU~\cite{GOOGLE_TPU_V4}, AMD's and Intel's Neural Processing Units (NPUs)~\cite{RYZEN_XDNA, INTEL_NPU}, Qualcomm's AI Engine~\cite{QUALCOMM_AI_ENGINE}, GroqChip~\cite{GROQCHIP}, Cerebras' Wafer Scale Engine~\cite{CEREBRAS_WAFER_SCALE_ENGINE}, and platforms from SambaNova~\cite{SAMBANOVA}. A defining common feature of these accelerators is the spatial allocation of compute kernels to fixed hardware regions (e.g., tiles or cores), where data is communicated via explicitly programmed on-chip data-paths, often decoupled from compute~\cite{singh2023sparta,singh2024rubicon}. While architectural designs vary, a common challenge remains: enabling compilers to map high-level programs to these platforms by spatial locality, data movement, and synchronization~\cite{singh2022designing}.

In response, the compiler community has developed a range of spatially aware compilation frameworks that aim to bridge the gap between abstract algorithm specification and low-level hardware control. These works largely focus on flexible frontends for compiler frameworks, compile transformations that enable efficient computation, compiler techniques for targeting a broad range of accelerators, or any combination thereof. The remainder of this section highlights notable works in each category.



\head{Frontends for Accelerator Programming} 
There is a large diversity of frontends for accelerator programming frameworks. IRON provides a close-to-metal interface that allows detailed and customized performance tuning~\cite{IRON_FCCM}. In contrast, frontends that capture intent at a higher level of abstraction are useful for flexibility, reusability, and quick adaptation to new algorithms and emerging programming models. Consequentially, \tech and other works have focused on this higher level of abstraction.
For instance, Union introduces a unified hardware-software co-design ecosystem within the MLIR infrastructure~\cite{jeong2021union}, which supports TensorFlow, ONNX, and COMET~\cite{comet} as inputs. 
Similarly, SODA-OPT supports various high-level languages as inputs, including Python, C++, and Fortran~\cite{agostini2022mlir}. XLA, while originally a standalone compiler for TensorFlow and JAX, has increasingly adopted MLIR components to enhance its modularity and extensibility~\cite{sabne2020xla}.
Both Union and SODA-OPT use MLIR internally to increase front-end flexibility; while XLA was originally a standalone compiler, it has increasingly adopted MLIR components to enhance its modularity and extensibility.
ARIES~\cite{zhuang2025aries} provides an MLIR-based flow targeting AMD AI Engines, with a focus on providing tile-granularity programming interface.
Unlike these frameworks, \tech defines a spatially explicit intermediate representation that directly models hardware concurrency, locality, and asynchronous execution inline of MLIR IRs, uniquely striking a balance between fine-grained compiler-managed scheduling and frontend and backend flexibility.

\head{Polyhedral Compilation for Mapping Tasks to Resources} The extraction and spatial mapping of parallelism implicit in algorithms are central to delivering high quality of results (QoR), especially for accelerators which are often composed of many parallel compute units.
The polyhedral model~\cite{polyhedral_model} provides a formal framework for analyzing and transforming loop nests through affine access relations and schedule functions.
Early efforts in this space include Vivado High-level Synthesis, which demonstrated how affine loop transformations could be applied to high-level code to generate efficient FPGA implementations~\cite{AUTOPILOT,singh2021fpga,singha2022leaper,singh2020nero}. AutoSA advanced this direction by introducing a full-stack polyhedral compilation flow targeting systolic arrays on FPGAs~\cite{AUTOSA}. It applies space-time transformations and loop tiling to generate parallel accelerator kernels that maximize throughput while respecting hardware resource constraints. More recent tools extend these capabilities across broader architectural targets.
Tools like Diesel~\cite{DIESEL_POLYHEDRAL} and PLUTO~\cite{PLUTO} utilize the polyhedral model to automatically parallelize and optimize loop nests across multiple hardware architectures, including multicore CPUs, GPUs, and FPGAs. Polygeist further enhances the applicability of polyhedral compilation by translating C to MLIR's \affine and \scf dialects, enabling integration with modern compiler infrastructure and reuse of polyhedral analyses with MLIR-based workflows~\cite{POLYGEIST}.
In contrast, \tech leverages polyhedral analyses not only for loop transformations but also to guide asynchronous scheduling and data movement, integrating these capabilities within a structured, token-based IR.

\head{Compiler Frameworks for Diverse Spatial Accelerators} Alongside the polyhedral model, many tools such as Marvel~\cite{chatarasi2021marvel} and AMOS~\cite{zheng2022amos} offer plug-and-play mechanisms for diverse spatial accelerator architectures. 
By abstracting device-specific optimizations and code generation, these tools focus on compute patterns and memory hierarchies common to spatial accelerators, facilitating seamless integration across diverse hardware generations and platforms.
Moreover, when targeting reconfigurable FPGA devices, frameworks like HIDA~\cite{ye2024hida} and Revet~\cite{rucker2024revet} enable automatic generation of Register Transfer Level (RTL) code, streamlining the hardware design process without requiring extra manual effort.
In contrast, \tech emphasizes explicit modeling of spatial scheduling and asynchronous execution within the IR itself, enabling precise control over task placement without relying on external runtime coordination or fixed hardware templates.

\section{\tech: A Novel Compiler Framework for Spatial Architectures}
\label{sec:air}

\begin{figure}[h]
  \centering
  \begin{tikzpicture}[thick, node distance=2.8cm]
    \newcommand{\sbToCore}{0.6}
    \newcommand{\xsep}{1.5}
    \newcommand{\ysep}{1.2}
    \tikzstyle {interconnect} = [<->, gray]
    \tikzstyle {sb} = [rectangle, rounded corners=0.8mm, minimum width=3mm, minimum height=3mm, align=center, draw, black, fill=black]
    \tikzstyle {core} = [rectangle, pattern=north west lines, pattern color=black!10!green, rounded corners, minimum width=5mm, minimum height=5mm, align=center, draw]
    \tikzstyle {mem} = [rectangle, pattern=north east lines, pattern color=black!20!yellow, rounded corners=0.8mm, minimum width=3mm, minimum height=5mm, align=center, draw]
    \definecolor{ballblue}{rgb}{0.13, 0.67, 0.8}
    \tikzstyle {memtile} = [rectangle, pattern=vertical lines, pattern color=ballblue, rounded corners, minimum width=8mm, minimum height=5mm, align=center, draw]
    \definecolor{cadetgrey}{rgb}{0.57, 0.64, 0.69}
    \tikzstyle {shimtile} = [rectangle, pattern=horizontal lines, pattern color=cadetgrey, rounded corners, minimum width=8mm, minimum height=5mm, align=center, draw]
    \tikzstyle{TextBox} = [rectangle, text centered, minimum height=1mm, minimum width=1mm, text width=2em]


    \node (topleft) [sb, white, opacity=100] at (0,0) {};
    \node (core04) [core, white, opacity=100] at ($(topleft.center)+(\sbToCore,-\sbToCore)$) {};
    \node (mem04) [mem, anchor=west, white, opacity=100] at (core04.east) {};

    \foreach \j [evaluate=\j as \y using \j*\ysep] in {0,1,2,3} {
        \foreach \i [evaluate=\i as \x using \i*\xsep] in {0,1,2,3} {
            \node (currSb) [sb] at ($(topleft.center)+(\x,-\y)$){};
            \node (currCore) [core] at ($(core04.center)+(\x,-\y)$){};
            \node (currMem) [mem] at ($(mem04.center)+(\x,-\y)$){};
            \draw [interconnect] ($(currSb.east)+(0.2,0)$) -- ($(currSb.east)+(1.0,0)$);
            \draw [interconnect] ($(currSb.south)+(0,-0.2)$) -- ($(currSb.south)+(0, -0.7)$);
            \draw [interconnect] ($(currSb.south east)+(0.01,-0.01)$) -- ($(currCore.north west)+(-0.01, 0.01)$);
        }
    }

    
    \node (memtile01) [memtile, white, opacity=100, anchor=west] at ($(core04.west)+(0,-4.8)$) {};
    \foreach \i [evaluate=\i as \x using \i*\xsep] in {0,1,2,3} {
        \node (currSb) [sb] at ($(topleft.center)+(\x,-4.8)$){};
        \node (currCore) [memtile, anchor=west] at ($(core04.west)+(\x,-4.8)$){};
        \draw [interconnect] ($(currSb.east)+(0.2,0)$) -- ($(currSb.east)+(1.0,0)$);
        \draw [interconnect] ($(currSb.south)+(0,-0.2)$) -- ($(currSb.south)+(0, -0.7)$);
        \draw [interconnect] ($(currSb.south east)+(0.01,-0.01)$) -- ($(currCore.north west)+(-0.01, 0.01)$);
    }

    
    \node (shimtile00) [shimtile, white, opacity=100, anchor=west] at ($(core04.west)+(0,-6.0)$) {};
    \foreach \i [evaluate=\i as \x using \i*\xsep] in {0,1,2} {
        \node (currSb) [sb] at ($(topleft.center)+(\x,-6.0)$){};
        \node (currCore) [shimtile, anchor=west] at ($(core04.west)+(\x,-6.0)$){};
        \draw [interconnect] ($(currSb.east)+(0.2,0)$) -- ($(currSb.east)+(1.0,0)$);
        \draw [interconnect] ($(currSb.south east)+(0.01,-0.01)$) -- ($(currCore.north west)+(-0.01, 0.01)$);
    }
    \node (currSb) [sb] at ($(topleft.center)+(4.5,-6.0)$){};
    \node (currCore) [shimtile, anchor=west] at ($(core04.west)+(4.5,-6.0)$){};
    \draw [interconnect] ($(currSb.south east)+(0.01,-0.01)$) -- ($(currCore.north west)+(-0.01, 0.01)$);

        
    \path ([xshift=2mm,yshift=0mm]current bounding box.east) node[matrix,anchor=west,cells={nodes={font=\tiny,anchor=west}}, draw,thick,inner sep=1ex, row sep=0mm]{
        \node [core]{}; & \node{Core tile.};\\
        \node [mem, minimum width=5mm, rounded corners]{}; & \node{Memory.};\\
        \node [memtile, minimum width=5mm]{}; & \node{Memory tile.};\\
        \node [shimtile, minimum width=5mm]{}; & \node{Shim tile.};\\
        \node [sb]{}; & \node{Switch box.};\\
        \draw[interconnect](0,0) -- ++ (0.5,0); & \node{\makecell[l]{Streaming\\interconnect.}};\\
    };

\end{tikzpicture}
  \caption{
      AMD NPU architecture.
  }
  \label{fig:npu_arch}
\end{figure}
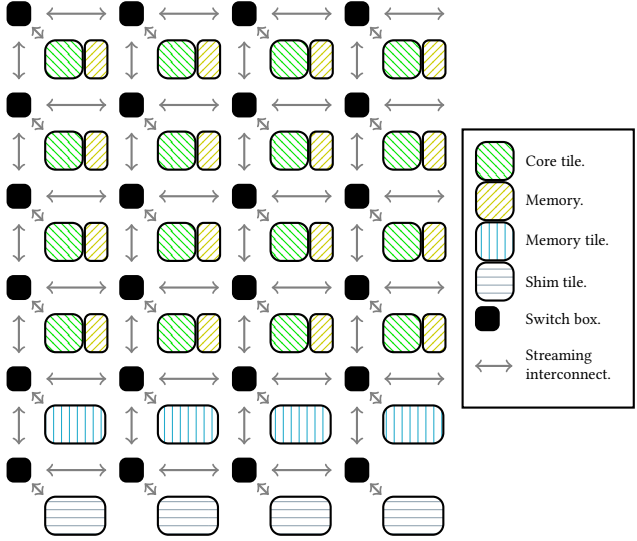

Modern spatial accelerators require the compiler to do more than expose parallelism, they require explicit control over placement, communication, and execution order. \tech is built to provide that control natively.

\tech is a novel, platform-agnostic compiler framework designed to target a wide range of spatial architectures.
In this work, we focus on its instantiation for AMD NPUs --- a tiled architecture optimized for high-throughput and low-latency AI computations.



As shown in Figure~\ref{fig:npu_arch}, the AMD NPU architecture consists of a two-dimensional grid of compute (\coretile), memory (\memtile), and shim tiles (\shimtile). 
Shim tiles form the interfacing row which connects the NPU to host memory and I/O systems. These are the only tiles that can initiate a memory transaction to the SoC memory system. Memory tiles, located adjacent to the shim tiles, provide shared memory resources accessible by compute tiles throughout the array.
Compute tiles comprise the majority of the array, each integrating local memory buffers with scalar and vector engines.  Every tile features a dedicated DMA engine for block data transfers (represented in buffer descriptors, or BDs) over a reconfigurable streaming interconnect.
This enables localized compute-memory communication, via the streaming interconnects---and peer-to-peer data movement, via either the dedicated cascade connections between cores or local memory shared by neighbors (\coremem). The absence of caches, either for data or instructions, and the emphasis on computing on local tiles of data ( eliminating memory access latency variation) means the architecture is characterized by extremely deterministic behaviour. Compilers designed to tile up work to fit into local memories can use the predictable behavior to construct efficient data-flow achieving high utilization. 

\begin{figure}[h]
    \centering
    \includegraphics[width=\textwidth]{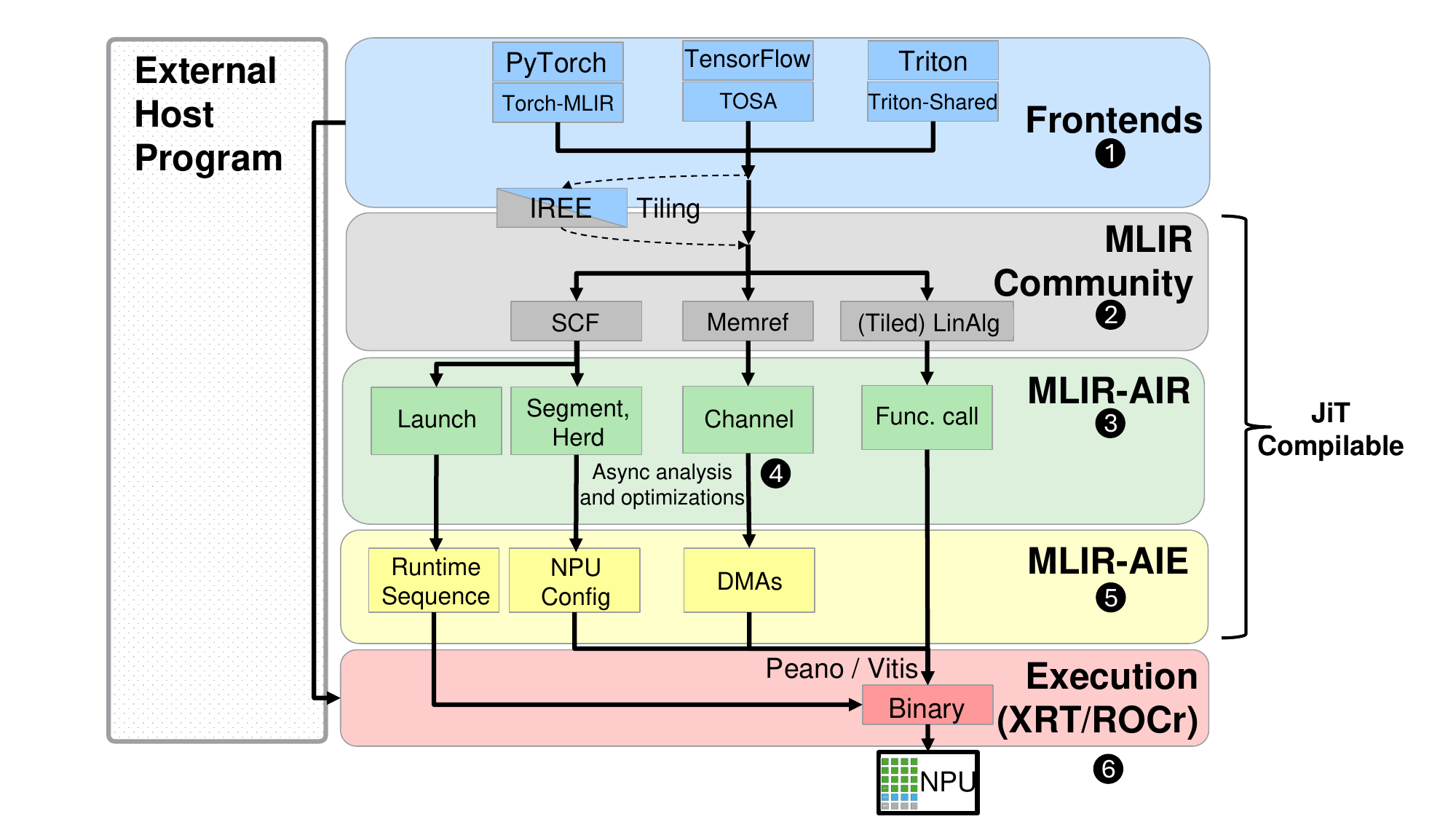}
    \caption{\tech stack overview.} 
    \label{fig:air_toolflow}
\end{figure}

\tech is designed to bridge high-level algorithmic representations with 
the low-level spatial execution requirements of modern accelerators, such as the AMD NPU. 
It provides the abstractions and transformations necessary to translate structured programs into tiled, explicitly scheduled implementations. 
Figure~\ref{fig:air_toolflow} illustrates the \tech compilation flow for the AMD NPU, highlighting its integration with MLIR's ecosystem and spatial backend tools.

\head{Algorithm-Level Programming Interface}
At the compiler frontend (\circled{1}), \tech interfaces with high-level AI frameworks such as PyTorch, TensorFlow, and Triton through MLIR-compatible frontends including Torch-MLIR, TOSA, and Triton-Shared. These frameworks emit programs using structured MLIR representations that preserve loop nests, tensor operations, and affine indexing.

Frontend dialects are lowered into MLIR common components (\circled{2}), including structured control flow (SCF) and linear algebra (LinAlg) dialects to provide an algorithm-friendly interface. These dialects offer C-like generic programming abstractions that preserve loop structure and tensor semantics, making AI workloads analyzable by non-domain experts. 
Unlike traditional low-level compilation flows that are tightly coupled to specific frontends or hardware backends, \tech decouples emerging AI frontends from new accelerator architectures, defining a common compute model fit for the new era of spatial hardware. 

\head{Representation of Asynchronous Parallelism}
At the core of \tech is the \air dialect (\circled{3}), a set of compiler abstractions that explicitly model hardware scheduling, asynchronous execution, and interactions with the memory hierarchy. Unlike conventional IRs that assume shared memory and centralized scheduling, AIR models the constraints and opportunities of spatial systems directly in the compiler.
\tech captures fine-grained asynchronous parallelism through an asynchronous control and dataflow graph (ACDG), a directed acyclic graph that encodes MLIR operation-level dependencies sequencing computation and data movement. 
The ACDG is embedded directly in the \tech IR via the production and consumption of \token values, which are static single assignment (SSA) values encoding the read-after-write (RAW), write-after-read (WAR), and write-after-write (WAW) dependency types. 
This token-based mechanism integrates with MLIR's native SSA dominance and verification infrastructure, enabling automatic correctness checks and transformation legality throughout the compilation process.

ACDG constructs are composable with MLIR's structured control flow, supporting structured parallelism through tokens yielded by \scfpar and AIR spatial operations (see Section~\ref{sec:air_concepts}). 
This allows explicit encoding of both inter- and intra-loop parallelism. 
Furthermore, loop-carried dependencies and pipelining opportunities are represented explicitly via tokens passed through \scffor iteration arguments, enabling the compiler to reason about and optimize fine-grained execution schedules, including pipeline stages and resource contention.

\head{Decoupled Data Movement Primitives}
Unlike conventional memory copy operations that couple source and destination within a single construct, \tech introduces decoupled data movement \channelput and \channelget operations (\circled{4}) to model unidirectional data transfers localized to their respective memory hierarchies. 
These operations are linked via a globally declared symbolic \channel, which abstracts the communication path and enforces backpressure-based synchronization between asynchronous code regions. 
This decoupling enables fine-grained control over dataflow, allowing communication to be scheduled alongside computation when desired, or independently when beneficial for performance or modularity.
The design closely aligns \channel operations with their target memory scopes, allowing the compiler to reason about and optimize data movement via pattern matching of simple and localized codes.

\head{Optimization and Performance Feedback}
Beyond compilation, \tech enhances the optimization and debugging process by providing execution traces that capture key performance metrics during hardware execution. 
These traces are visualized using tools like Chrome Tracing or Perfetto UI~\cite{PERFETTO}, allowing developers to analyze the runtime parallelism between computation and data movement. 
This profiling capability enables fine-grained performance evaluation, helping developers identify bottlenecks and inefficiencies in execution.


\head{Platform-agnostic Implementation and Runtime} 
\tech supports integration with multiple hardware implementation backends and runtime systems, enabling platform-agnostic compilation. 
The final lowered IR is consumed by platform-specific tools such as MLIR-AIE (\circled{5}) for NPUs~\cite{IRON_FCCM} or LLVM-based pipelines for CPUs and GPUs, which generate hardware-specific control and dataflow representations. 
These are subsequently compiled and deployed using runtime frameworks (\circled{6}) such as XRT~\cite{XRT} or ROCr~\cite{ROCR}, ensuring compatibility with diverse hardware platforms. This modular backend integration facilitates scalable and efficient deployment while preserving the architectural flexibility of \tech across spatially heterogeneous systems.

\section{{\air} Concepts}
\label{sec:air_concepts}

The \air dialect provides the core primitives for expressing spatial execution semantics in \tech. 
These primitives are designed to give the compiler fine-grained control over execution, concurrency, communication, and synchronization at various levels of granularity. These primitives can then be targeted by architecture-specific backends.  

\air integrates within existing compilation stacks, and transforms allow designs to be ingested from several different frontends. The \air dialect is designed to compose with standard MLIR dialects, reusing existing dialects to describe computation and kernel. 
This modularity and flexibility allow developers to describe fine-grained control over execution, communication, and memory behavior while maintaining portability by decoupling these abstractions from vendor-specific hardware implementations.


We group the new operations in the \air dialect into three categories:
\begin{itemize}
    \item \textbf{Scheduling Constructs}, which express spatial and hierarchical parallelism across compute resources (Section~\ref{subsec:scheduling_constructs}).
    \item \textbf{Data Locality Constructs}, which represent explicit and decoupled data transfers aligned with memory hierarchy and DMA affinity (Section~\ref{subsec:data_locality_constructs}).
    \item \textbf{Synchronization Constructs}, which captures explicit operation-level and loop-carried dependencies for asynchronous execution and pipelining (Section~\ref{subsec:sync_constructs}).
\end{itemize}

Together, these abstractions allow \air to represent concurrency control, memory placement and movement, and synchronization between many concurrent actors as explicit compiler-visible constructs.

\subsection{Scheduling Constructs}
\label{subsec:scheduling_constructs}

\air introduces scheduling constructs that express how computation is distributed and executed across a spatial accelerator. 
These constructs define task placement, launch behavior, and hardware resource partitioning, forming the foundation of spatial scheduling in \air.
The dialect includes \launch, \segment, and \herd operations, which are hierarchically composed: an \launch may contain multiple \segments, each of which may dispatch one or more \herd operations.

\subsubsection{\texttt{\textbf{air.launch}}}

The \launch operation defines a region of computation to be offloaded from the host processor to the accelerator. It is designed as a construct to support portability and scaling by selecting groups of operations whose dispatch may be deferred to a \emph{runtime} scheduler. 
It groups together compute, communication, and synchronization operations into a single launch unit, which are scheduled at runtime.

The optional iteration space attribute attached to an \launch operation describes a set of unique instances of the region body that the runtime scheduler is delegated to manage. Those unique instances must be permutable, \emph{i.e.}, a completely parallel schedule is legal, and instances within the \launch iteration space must not rely on observing the effects of any other instance.

To improve the effectiveness of compile-time optimization, we assume the compiler is free to hand off multiple different variants of the compiled \launch operation to the runtime, each enabling optimized dispatch of a parameterizable subset of the iteration space. Our lowering for AMD NPU architecture uses this freedom to offer different opportunistic granularities (e.g., one column or whole array) for the runtime to schedule work. \launch also manages the lifetime of the bound resources that implement the operations hierarchically nested within its body. Once all nested operations within a launch iteration are scheduled and begin execution, the launch is able to release unused resources back to the runtime. This hierarchical management of resources ensures efficient resource utilization, especially when tasks are nested within larger compute tasks.

\subsubsection{\texttt{\textbf{air.segment}}}
The body of an \segment encapsulates the reservation of pool(s) of compute resources for use in scheduling the operations nested inside them.
Segments can be optionally annotated with architecture-specific attributes describing the pool of resources they are reserving, when targeting backends that benefit from resource-aware scheduling.
An architecture might want to define two pools of resources that have physical affinity (e.g., resources in one chiplet) so that they can ensure that the scheduler dispatches \herd operations within that segment exclusively using the segment resources. 

Segments have an optional grid space. This allows easy replication of resource pools. Segment instances within that space are dispatched concurrently. Other relationships between the scheduling of segments in time and space can be controlled using the synchronization constructs in Section~\ref{subsec:sync_constructs}.

\subsubsection{\texttt{\textbf{air.herd}}}

The \herd operation defines a group of work units that execute concurrently on a grid of physical compute units and their local memories.
It contains an index space which, expressed as an affine set of worker indices, generalizes the notion of thread IDs found in traditional parallel programming models (e.g., CUDA~\cite{cuda} or OpenMP~\cite{openmp}).
Each \textit{worker} in the \herd executes the same region body, but specialization is enabled by indexing: control flow and memory access patterns may diverge based on each worker's coordinates in the \herd.
\herd operations are scheduled atomically : they are only scheduled when resources for \textit{all} the workers are available, and must enable independent forward progress for their individual workers. It is implied that workers are allocated as a physically local contiguous block, and lowerings may make use of the grid dimensions to lower to architecture-specific features that make use of that physical locality. The size of \herd operations indicates the granularity of (concurrent) dispatch; users should be aware that certain architectures may place limits on the size of resources it can guarantee run concurrently (e.g., lowerings may fail during backend compilation if unimplementable \herd operations are created). 

Where multiple \herd operations are included in a \launch, their default behavior is to run sequentially. Programmers may use the advanced synchronization constructions in Section~\ref{subsec:sync_constructs} to set additional constraints on \herd operations to guarantee sequential or concurrent execution and consequently modify the resource requirements of the surrounding \launch.

\subsection{Data Locality Constructs}
\label{subsec:data_locality_constructs}

Spatial architectures rely heavily on local memory hierarchies and explicit DMA engines to achieve high efficiency.
\tech introduces constructs that make data locality explicit, enabling the compiler to reason about and optimize data movement across compute tiles and memory spaces. 

When ingesting code from an AI framework, progressive lowerings are supported by use of MLIR \memref types---which represent typed memory references to structured data and are assumed to reside in global memory if not explicitly annotated.
Existing lowerings support explicit memory allocation and movement into at least two further levels of addressable memory hierarchy (shared cluster scratchpads and private memory local to a worker).  

We support two levels of abstraction in our data movement constructs. First, to support lowering from higher-level dialects, \tech supports an \memcpy operation. Second, to provide further control over memory movement, \tech provides an \channel operation that abstracts architecture-specific optimizations in device interconnect and specialized tensor-DMAs. 

\subsubsection{\texttt{\textbf{air.memcpy}}}
To progressively bridge the gap between high-level memory transfer specifications and spatial hardware implementations, \tech introduces an intermediate \memcpy construct.
\memcpy enhances the conventional \texttt{memcpy} operation with explicit attributes for data layout and memory spaces. This allows users to indicate which levels of hierarchy they are transferring from, and to express the desire for an in-flight physical reshaping of the data, decoupling the logical layout of a \memref from its physical representation in memory. This is useful because data transfer operations offer an opportunity to specialize data layout on the fly.
In subsequent lowering paths, \texttt{memcpy} operations are lowered to make use of \channel operations. 

\subsubsection{\texttt{\textbf{air.channel}}}
Many modern spatial accelerators, including AMD NPUs and NVIDIA Hopper GPUs, expose hierarchies of data movement engines and memory spaces that require explicit software modeling for efficient execution.
To support this, \tech introduces the \channel abstraction, which represents stream-based data transfers between distinct memory regions through paired \texttt{put} and \texttt{get} operations:
\begin{itemize}
    \item \texttt{put} operations transfer data from a source memory address in one level of the memory hierarchy onto a serialized stream, and 
    \item \texttt{get} operations retrieve data from the stream into a destination memory address, representing a buffer in a particular level of the memory hierarchy.
\end{itemize}
These operations are placed in the code regions local to their respective memory allocations, enabling the compiler to express and optimize DMA-to-memory affinity.
It captures both endpoints of the communication via a globally scoped symbolic \channel, enabling an ordered and streamed data transfer.
Subsequent compiler passes, detailed in Section~\ref{subsec:inferring_dataflow_channel}, then decouple \memcpy (Listing~\ref{lst:memcpy}) into discrete \channelput and \channelget operations (Listing~\ref{lst:channel}).

Notably, \channel operation references can cross levels of the \air construct hierarchy. For example, an \channelput operation that is the immediate child of an \launch operation may push data into an \channel whose consumers are more deeply nested \channelget operations inside \herd and \segment operations.

The \channel abstraction integrates naturally with the \memref dialect by adopting the same offset, size, and stride specifications for describing structured memory accesses.
As shown in Listing~\ref{lst:channel}, \channel operations operate over structured \memref views, allowing tensor access patterns to remain analyzable and composable with other MLIR transformations.



\begin{figure}[ht]
\centering
\begin{minipage}{0.48\textwidth}
\begin{lstlisting}[language=mlir, caption={\memcpy operation\protect\footnotemark.}, label=lst:memcpy, escapechar=|]
air.memcpy (%y, %x)
\end{lstlisting}
\end{minipage}%
\hfill
\begin{minipage}{0.48\textwidth}
\begin{lstlisting}[language=mlir, caption={Equivalent \channel pair.}, label=lst:channel, escapechar=|]
air.channel.put @chan1 (%x)
air.channel.get @chan1 (%y)
\end{lstlisting}
\end{minipage}
\label{fig:air_memcpy_channel_comparison}
\end{figure}

\tech also supports broadcasting within \channel operations, enabling a single source to supply data to multiple consumers without redundant resource usage.
Broadcasting is explicitly specified through affine integer sets, which define mappings from input indices to sets of output indices and are specialized into \affineif conditions at lower levels.
The detection and lowering of broadcast patterns is further discussed in Section~\ref{subsec:bcast_detect_and_lowering}.

Named bundles of \channel symbols are supported to allow users to select a specific \channel to put/get buffers (using a numerical index). 

\footnotetext{For simplicity, we omit the offsets, sizes, and strides lists from this code snippet.}

\subsubsection{\texttt{\textbf{air.herd}}}

The \herd operation not only defines parallel execution but also plays a crucial role in data locality. 
Since all workers in an \herd run concurrently on allocated local resources---including compute units, local memories, and tile-local data movers---we can optimize data movement between them using methods such as double buffering to minimize memory latency (see Section~\ref{subsubsec:pingpong}). 
The lowering of \herd to hardware platforms such as AMD NPUs ensures spatial contiguity of worker placement, allowing architecture-specific features to be exploited for efficient implementation of communications.
For example, in its default setting, the physical lowering of \herd operations to NPU AI Engine arrays guarantees that neighboring workers can write to the local memory of their neighboring cores. This allows an efficient specialized lowering of certain patterns of channel communication within the \herd. 
By constraining data exchange to local resources, \herd operations improve the dataflow efficiency within their scope.

\subsubsection{\texttt{\textbf{air.segment}}}

As a resource management construct, \segment also contributes to data locality by controlling memory partitioning and affinity constraints. 
Since \segments dictate how resources are allocated and shared, they help ensure that accesses to the shared memories remain localized within their data producers and consumers, including the data movers and any \herd operations within an \segment.
This reduces unnecessary data movement, keeping computation and memory access spatially co-located for better efficiency.

\subsection{Synchronization Constructs}
\label{subsec:sync_constructs}

\subsubsection{\texttt{\textbf{air.token}}}

The \token construct provides fine-grained control over execution dependencies in asynchronous workloads, with the granularity tunable from coarse-grained synchronization over regions of code to fine-grained per-operation scheduling. 
When an operation is marked with the \emph{async} keyword, it returns an \token that signals its completion status; this \token can be used to constrain the relative scheduling or placement of operations in time or space. 

Synchronization using \token is managed through two mechanisms. Firstly, explicit \waitall operations, can be inserted into synchronous control flow. This allow us to prevent further operation dispatch until tokens specified in the \waitall operation have signaled completion. Secondly, synchronization lists, that explicitly specify the relative scheduling of operations in time and space, describe a scheduling graph. Backend lowerings can use this graph to push dependency resolution to distributed dispatchers in hardware devices, allowing offload of groups of operations. 

The compute model envisages three types of relationships between operations controlled by synchronization lists. 

\begin{itemize}
    \item \emph{Dependency} lists encode directed edges between an operation and the predecessor operations in which it's inputs are defined.
    It requires that all inputs to a operation that are modified by a source operation in the dependency list are visible before the sink operation is scheduled. This is often implemented as a \textit{happens-before} relationship to control the scheduling of operations in time. 
    \item \emph{Concurrency} lists constrain the scheduling of operations in space and time. Each \token in a concurrency list defines an undirected edge between two operations that indicates they must be scheduled at the same time. This implies that each operation must use exclusive resources.  
    \item \emph{Affinity} lists constrain the scheduling of operations in space. These are lists of tokens that define undirected edges between operations that must execute using the same resources. In practice, this means these operations' time-slots must be disjoint, but the edge does not describe which operation must be scheduled first. The edges provide information on where spatial affinity could be exploited by the compiler or runtime.
\end{itemize}

Details on how \tech automatically detects and lowers data dependencies are included in Section~\ref{subsec:async_dep_analysis}.\air's parallelism constructs, such as \launch, signal \token completion by behaving as a grouped asynchronous task: an \launch's \token is released only when all operations within the \launch have completed.

\subsubsection{\texttt{\textbf{air.channel}}}

While primarily a data movement construct, \channel also plays an essential role in synchronization across data movers operating on discrete memory spaces, where an \channelget operation is synchronized to an \channelput with the same \channel by back pressure, as shown in Listing~\ref{lst:channel}. 

This synchronization abstraction---combined with asynchronous dependencies on \channel actors to enforce synchronization local to each memory space---is both simple and effective: enforcement of dependencies between distributed data movers does not require complex control-flow dependencies across code regions. 

\section{\air Dialect Constructs in Use}
\label{sec:air_example}

To concretely illustrate how \air dialect constructs appear in practice, we present a simplified example of an element-wise vector addition program. 
This example bridges the conceptual descriptions of Section~\ref{sec:air_concepts} and the compiler transformations detailed in Section~\ref{sec:opts}.

The input program, shown in Appendix~\ref{appendix:vecadd_expanded}, expresses a tiled vector addition in generic SCF using \scfpar and \scffor. 
We use explicit \memrefcopy operations to move data between \memref objects scheduled in a loop iteration. 
The program is agnostic to the target hardware and does not yet reflect spatial execution, memory locality, or asynchronous scheduling.


The corresponding AIR-transformed IR, shown in Listing~\ref{lst:vecadd_output}, makes the spatial and asynchronous execution explicit. In this transformed version:
\begin{itemize}
    \item The outer \scfpar loop is replaced by an \launch enclosing an \herd, assigning each iteration to a spatial tile in a 2D compute grid.
    \item Temporary buffer allocations are restructured with explicit memory hierarchy annotations, and all data movement operations are rewritten using \memcpy or decoupled \channelput and \channelget.
    \item Execution dependencies across asynchronous regions are made explicit with \token values, enabling pipelined execution between data transfer and compute stages.
\end{itemize}

The next section describes the key compilation stages in this IR transform.

\begin{minipage}{\linewidth}
\begin{lstlisting}[language=mlir, caption={Element-wise vector add described in \air dialect. The syntax has been simplified and reformatted for clarity of presentation; some MLIR dialect annotations and attributes are omitted.}, label={lst:vecadd_output}, basicstyle=\ttfamily\tiny, xleftmargin=0.5em, xrightmargin=1.8em,] 
air.channel @channel_0 [1, 2]
air.channel @channel_1 [1, 2]
air.channel @channel_2 [1, 2]
func.func @eltwise_add(%arg0: memref<65536xf32>, %arg1: memref<65536xf32>, %arg2: memref<65536xf32>) {
  %0 = air.launch async (%arg3, %arg4) in (1, 1) args(%arg7=%arg0, %arg8=%arg1, %arg9=%arg2) {
    %1 = air.segment @eltwise_add_0 async  args(%arg10=%arg7, %arg11=%arg8, %arg12=%arg9) {
      %2 = air.channel.put async  @channel_0[0, 0] (%arg10[0, 0, 0] [32, 2, 512] [2048, 512, 1])
      %3 = air.channel.put async  @channel_0[0, 1] (%arg10[0, 0, 1024] [32, 2, 512] [2048, 512, 1])
      %4 = air.channel.put async  @channel_1[0, 0] (%arg11[0, 0, 0] [32, 2, 512] [2048, 512, 1])
      %5 = air.channel.put async  @channel_1[0, 1] (%arg11[0, 0, 1024] [32, 2, 512] [2048, 512, 1])
      %6 = air.channel.get async  @channel_2[0, 0] (%arg12[0, 0, 0] [32, 2, 512] [2048, 512, 1])
      %7 = air.channel.get async  @channel_2[0, 1] (%arg12[0, 0, 1024] [32, 2, 512] [2048, 512, 1])
      %8 = air.herd @herd_0 async  tile (%arg13, %arg14) in (1, 2) {
        %async_token, %results = memref.alloc() async
        %async_token_4, %results_5 = memref.alloc() async
        %async_token_6, %results_7 = memref.alloc() async
        %async_token_8, %results_9 = memref.alloc() async
        %async_token_10, %results_11 = memref.alloc() async
        %async_token_12, %results_13 = memref.alloc() async
        %9 = air.wait_all async deps=[%async_token_10, %async_token_12] 
        %10:3 = scf.for %arg17 = 0 to 65536 step 4096 iter_args(%arg18 = %9, %arg19 = %async_token_12, %arg20 = %async_token_12) {
          %11 = air.channel.get async deps=[%arg18, %arg20]  @channel_0[%arg13, %arg14] (%results_9[] [] [])
          %12 = air.channel.get async deps=[%arg18, %arg20]  @channel_1[%arg13, %arg14] (%results_13[] [] [])
          %13 = air.wait_all async deps=[%11, %12] 
          %14 = scf.for %arg21 = 0 to 1024 step 1 iter_args(%arg22 = %13) {
            %async_token_20, %results_21 = memref.load async deps=[%arg22] %results_9[%arg21]
            %async_token_22, %results_23 = memref.load async deps=[%arg22] %results_13[%arg21]
            %22 = arith.addf %results_21, %results_23
            %async_token_24 = memref.store async deps=[%arg22] %22, %results_11[%arg21]
            %23 = air.wait_all async deps=[%async_token_20, %async_token_22, %async_token_24] 
            scf.yield %23
          }
          %15 = air.channel.put async deps=[%arg19, %arg18, %14]  @channel_2[%arg13, %arg14] (%results_11[] [] [])
          %16 = air.channel.get async deps=[%arg19]  @channel_0[%arg13, %arg14] (%results[] [] [])
          %17 = air.channel.get async deps=[%arg19]  @channel_1[%arg13, %arg14] (%results_5[] [] [])
          %18 = air.wait_all async deps=[%16, %17, %arg18] 
          %19 = scf.for %arg21 = 0 to 1024 step 1 iter_args(%arg22 = %18) {
            %async_token_20, %results_21 = memref.load async deps=[%arg22] %results[%arg21]
            %async_token_22, %results_23 = memref.load async deps=[%arg22] %results_5[%arg21]
            %22 = arith.addf %results_21, %results_23
            %async_token_24 = memref.store async deps=[%arg22] %22, %results_7[%arg21]
            %23 = air.wait_all async deps=[%async_token_20, %async_token_22, %async_token_24] 
            scf.yield %23
          }
          %20 = air.channel.put async deps=[%19, %arg18]  @channel_2[%arg13, %arg14] (%results_7[] [] [])
          %21 = air.wait_all async deps=[%16, %17] 
          scf.yield %15, %20, %21
        }
        /* Memref deallocations were omitted. */
      }
      air.wait_all [%2, %3, %4, %5, %6, %7, %8] 
    }}
  return
}
\end{lstlisting}
\end{minipage}

\section{Compilation of \air dialect to Spatial Hardware}
\label{sec:opts}

This section builds upon the \air constructs introduced in Section~\ref{sec:air_concepts}, and describes the core compiler optimizations in \tech that transform high-level loop-based programs into efficient spatial implementations. 
These passes progressively transform generic operations into tiled subproblems, resolve dependencies for correct asynchronous execution, optimize data reuse and communication locality, and finally lower the program into hardware-executable IRs targeting AMD NPUs. 
The following subsections describe each stage in this process:
\begin{itemize}
    \item \textbf{Tiling and Parallelism Mapping:} Maps high-level operations to distinct hardware tiles via loop tiling and parallel loop conversion (Section~\ref{subsec:tiling_opt}).
    \item \textbf{Broadcast Detection and Lowering:} Identifies and lowers data reuse patterns as affine-mapped broadcasts to reduce redundant transfers (Section~\ref{subsec:bcast_detect_and_lowering}).
    
    \item \textbf{Asynchronous Dependency Analysis:} Constructs fine-grained control and data dependencies represented using ACDG (Section~\ref{subsec:async_dep_analysis}).
    
    \item \textbf{Inferring Dataflow via \channel:} Decouples memory transfers into local operations, exposing DMA-to-memory affinity for scheduling (Section~\ref{subsec:inferring_dataflow_channel}).
    
    \item \textbf{Lowering to AMD NPU Targets:} Generates spatial hardware IR and runtime code for deployment on AMD NPUs (Section~\ref{subsec:lowering_to_targets}).
\end{itemize}


\subsection{Tiling and Parallelism Mapping}
\label{subsec:tiling_opt}

Tiling identifies parallel subregions of computation and introduces structured iteration constructs to represent them. 
These parallel tiles form the basis for spatial parallelism analysis and schedule optimization.
In \tech, tiling is performed through a compiler pass pipeline that lowers implicit parallelism in high-level operations (e.g., from the \linalg dialect) into explicit \scfpar or \scfforall loops. 
These are subsequently mapped to \air spatial constructs such as \launch and \herd. 
The pipeline leverages upstream MLIR tiling utilities  while also ensuring flexibility and compatibility when plugged into broader compiler ecosystems, including IREE~\cite{IREE} and Triton~\cite{TRITON}.

\begin{figure}[h]
  \centering
  \begin{subfigure}[t]{0.20\textwidth}
    \centering
    \begin{tikzpicture}[thick, node distance=2.8cm]
    \tikzstyle {arrow} = [->, >=stealth]
    \tikzstyle {buffer} = [minimum width=2mm, minimum height=5mm, align=center, draw]
    \tikzstyle {dot} = [circle, fill=black, inner sep=0.1mm, draw]
    \tikzstyle{TextBox} = [rectangle, text centered, minimum height=1em, minimum width=10em, text width=9em, scale=0.8]
    \tikzstyle{APattern} = [pattern=north west lines, pattern color=black!25!green]
    \tikzstyle{BPattern} = [pattern=horizontal lines, pattern color=blue, rotate=90]

    \newcommand{\xseparation}{1.6}
    \newcommand{\yseparation}{1.2}


    \node (c11a) [buffer, APattern] at (0,0) {};
    \node (c11b) [buffer, BPattern] at ($(c11a.center)+(0.5, 0)$) {};

    \node (c12a) [buffer, APattern] at ($(c11a.center)+(0.0, \yseparation)$) {};
    \node (c12b) [buffer, BPattern] at ($(c11b.center)+(0.0, \yseparation)$) {};

    \node (c13a) [buffer, APattern] at ($(c12a.center)+(0.0, \yseparation)$) {};
    \node (c13b) [buffer, BPattern] at ($(c12b.center)+(0.0, \yseparation)$) {};

    \node (c14a) [buffer, APattern] at ($(c13a.center)+(0.0, \yseparation)$) {};
    \node (c14b) [buffer, BPattern] at ($(c13b.center)+(0.0, \yseparation)$) {};

    \node (m1a) [buffer, APattern] at ($(c11a.center)+(-0.2, -1.4)$) {};
    \node (m1b) [buffer, BPattern] at ($(m1a.center)+(0.8, 0.4)$) {};
    \node (m2b) [buffer, BPattern] at ($(m1b.center)+(0.0, -0.25)$) {};
    \node (m3b) [buffer, BPattern] at ($(m2b.center)+(0.0, -0.25)$) {};
    \node (m4b) [buffer, BPattern] at ($(m3b.center)+(0.0, -0.25)$) {};

    \node (pt1) at ($(m1a.north)+(0.0, 0.2)$) {};
    \node (pt2) at ($(c12a.south)+(0.0, -0.2)$) {};
    \node (pt3) at (pt1.center |- pt2.center) {};
    \node (pt4) at ($(c11a.south)+(0.0, -0.2)$) {};
    \node (pt5) [dot, black!25!green] at (pt1.center |- pt4.center) {};
    
    \draw [arrow, black!25!green] (m1a.north) |- (pt2.center) -- (c12a.south);
    \draw [arrow, black!25!green] (pt5.center) -| (c11a.south);
    
    \node (pt6) at ($(c13a.south)+(0.0, -0.2)$) {};
    \node (pt7) [dot, black!25!green] at (pt1.center |- pt2.center) {};
    \draw [arrow, black!25!green] (pt7.center) |- (pt6.center) -- (c13a.south);
    
    \node (pt8) at ($(c14a.south)+(0.0, -0.2)$) {};
    \node (pt9) [dot, black!25!green] at (pt7.center |- pt6.center) {};
    \draw [arrow, black!25!green] (pt9.center) |- (pt8.center) -- (c14a.south);
    
    \node (pt10) at ($(m1b.south)+(0.10, 0.0)$) {};
    \draw [arrow, blue] (m1b.south) -- (pt10.center) |- (c11b.south);
    
    \node (pt11) at ($(m2b.south)+(0.15, 0.0)$) {};
    \draw [arrow, blue] (m2b.south) -- (pt11.center) |- (c12b.south);
    
    \node (pt12) at ($(m3b.south)+(0.20, 0.0)$) {};
    \draw [arrow, blue] (m3b.south) -- (pt12.center) |- (c13b.south);
    
    \node (pt13) at ($(m4b.south)+(0.25, 0.0)$) {};
    \draw [arrow, blue] (m4b.south) -- (pt13.center) |- (c14b.south);

    \newcommand{\borderleft}{0.4}
    \newcommand{\borderright}{0.8}
    \newcommand{\borderabove}{0.5}
    \newcommand{\borderbelow}{0.5}
    \draw [dashed] ($(c11a.center)+(-\borderleft,-\borderbelow)$) |- ($(c11b.center)+(\borderright,\borderabove)$) |- ($(c11a.center)+(-\borderleft,-\borderbelow)$);
    \draw [dashed] ($(c12a.center)+(-\borderleft,-\borderbelow)$) |- ($(c12b.center)+(\borderright,\borderabove)$) |- ($(c12a.center)+(-\borderleft,-\borderbelow)$);
    \draw [dashed] ($(c13a.center)+(-\borderleft,-\borderbelow)$) |- ($(c13b.center)+(\borderright,\borderabove)$) |- ($(c13a.center)+(-\borderleft,-\borderbelow)$);
    \draw [dashed] ($(c14a.center)+(-\borderleft,-\borderbelow)$) |- ($(c14b.center)+(\borderright,\borderabove)$) |- ($(c14a.center)+(-\borderleft,-\borderbelow)$);
    \node [TextBox, right, anchor=center, rotate=90] at ($(c12a.west)+(-0.8,0.5)$) {Core tiles};
    \draw [dashed] ($(m1a.center)+(-0.3,-0.6)$) |- ($(m1a.center)+(1.6,0.6)$) |- ($(m1a.center)+(-0.3,-0.6)$);
    
    \node [TextBox, right, anchor=center, rotate=90] at ($(m1a.west)+(-0.6,0.0)$) {Mem. tile};

    \draw [arrow, black!25!green] ($(m1a.south)+(0,-0.41)$) -- (m1a.south);
    \draw [arrow, blue] ($(m4b.west)+(0,-0.2)$) -- (m4b.west);


\end{tikzpicture}
    \vspace{-4mm}
    \caption{$1\times 4$ herd.}
    \label{fig:tiling_1x4}
  \end{subfigure}
  \hfill
  \hspace{-20mm}
  \begin{subfigure}[t]{0.49\textwidth}
    \centering
    \begin{tikzpicture}[thick, node distance=2.8cm]
    \tikzstyle {arrow} = [->, >=stealth]
    \tikzstyle {buffer} = [minimum width=2mm, minimum height=5mm, align=center, draw]
    \tikzstyle {dot} = [circle, fill=black, inner sep=0.1mm, draw]
    \tikzstyle{TextBox} = [rectangle, text centered, minimum height=1em, minimum width=10em, text width=9em, scale=0.8]
    \tikzstyle{APattern} = [pattern=north west lines, pattern color=black!25!green]
    \tikzstyle{BPattern} = [pattern=horizontal lines, pattern color=blue, rotate=90]

    \newcommand{\xseparation}{1.4}
    \newcommand{\yseparation}{1.2}


    \node (c11a) [buffer, APattern] at (0,0) {};
    \node (c11b) [buffer, BPattern] at ($(c11a.center)+(0.5, 0)$) {};

    \node (c21a) [buffer, APattern] at ($(c11a.center)+(\xseparation, 0.0)$) {};
    \node (c21b) [buffer, BPattern] at ($(c11b.center)+(\xseparation, 0.0)$) {};

    \node (c31a) [buffer, APattern] at ($(c21a.center)+(\xseparation, 0.0)$) {};
    \node (c31b) [buffer, BPattern] at ($(c21b.center)+(\xseparation, 0.0)$) {};

    \node (c41a) [buffer, APattern] at ($(c31a.center)+(\xseparation, 0.0)$) {};
    \node (c41b) [buffer, BPattern] at ($(c31b.center)+(\xseparation, 0.0)$) {};

    \node (m1a) [buffer, APattern] at ($(c31a.center)+(-1.2, -4.9)$) {};
    \node (m2a) [buffer, APattern] at ($(m1a.center)+(0.25, 0.0)$) {};
    \node (m3a) [buffer, APattern] at ($(m2a.center)+(0.25, 0.0)$) {};
    \node (m4a) [buffer, APattern] at ($(m3a.center)+(0.25, 0.0)$) {};
    \node (m1b) [buffer, BPattern] at ($(m1a.center)+(1.3, 0.0)$) {};

    \node (pt1) at ($(m1a.north)+(0.0, 0.2)$) {};
    
    \draw [arrow, black!25!green] (m1a.north) -- (pt1.center) -| (c11a.south);
    
    \node (pt2) at ($(m2a.north)+(0.0, 0.25)$) {};
    \draw [arrow, black!25!green] (m2a.north) -- (pt2.center) -| (c21a.south);
    
    \node (pt3) at ($(m3a.north)+(0.0, 0.25)$) {};
    \draw [arrow, black!25!green] (m3a.north) -- (pt3.center) -| (c31a.south);
    
    \node (pt4) at ($(m4a.north)+(0.0, 0.20)$) {};
    \draw [arrow, black!25!green] (m4a.north) -- (pt4.center) -| (c41a.south);
    
    \node (pt5) at ($(m1b.south)+(0.2, 0.0)$) {};
    \node (pt6) at ($(c41b.south)+(0.2, 0.0)$) {};
    \draw [arrow, blue] (m1b.south) -| (pt6.center) -- (c41b.south);
    
    \node (pt7) [dot, blue] at ($(pt6.center)+(0.0, -0.5)$) {};
    \node (pt8) at ($(c41b.south)+(0.0, -0.5)$) {};
    \draw [arrow, blue] (pt7.center) -| (pt6.center) -- (c41b.south);
    
    \node (pt9) at ($(c31b.south)+(0.2, 0.0)$) {};
    \draw [arrow, blue] (pt7.center) -| (pt9.center) -- (c31b.south);
    
    \node (pt10) [dot, blue] at (pt7.center -| pt9.center) {};
    \node (pt11) at ($(c21b.south)+(0.2, 0.0)$) {};
    \draw [arrow, blue] (pt10.center) -| (pt11.center) -- (c21b.south);
    
    \node (pt12) [dot, blue] at (pt10.center -| pt11.center) {};
    \node (pt13) at ($(c11b.south)+(0.2, 0.0)$) {};
    \draw [arrow, blue] (pt12.center) -| (pt13.center) -- (c11b.south);
    
    \newcommand{\borderleft}{0.2}
    \newcommand{\borderright}{0.6}
    \newcommand{\borderabove}{0.6}
    \newcommand{\borderbelow}{0.6}
    \draw [dashed] ($(c11a.center)+(-\borderleft,-\borderbelow)$) |- ($(c11b.center)+(\borderright,\borderabove)$) |- ($(c11a.center)+(-\borderleft,-\borderbelow)$);
    \draw [dashed] ($(c21a.center)+(-\borderleft,-\borderbelow)$) |- ($(c21b.center)+(\borderright,\borderabove)$) |- ($(c21a.center)+(-\borderleft,-\borderbelow)$);
    \draw [dashed] ($(c31a.center)+(-\borderleft,-\borderbelow)$) |- ($(c31b.center)+(\borderright,\borderabove)$) |- ($(c31a.center)+(-\borderleft,-\borderbelow)$);
    \draw [dashed] ($(c41a.center)+(-\borderleft,-\borderbelow)$) |- ($(c41b.center)+(\borderright,\borderabove)$) |- ($(c41a.center)+(-\borderleft,-\borderbelow)$);
    \node (core_tiles) [TextBox, right, anchor=center, rotate=90, text=white] at ($(c11a.west)+(-0.6,0.0)$) {Core tiles};
    \draw [dashed] ($(m1a.center)+(-0.3,-0.6)$) |- ($(m1a.center)+(1.6,0.6)$) |- ($(m1a.center)+(-0.3,-0.6)$);
    
    \node [TextBox, right, anchor=center, rotate=90, text=white] at (core_tiles |- m1a.center) {Mem. tile};

    \draw [arrow, black!25!green] ($(m2a.south)+(0.125,-0.41)$) -- ($(m2a.south)+(0.125,0.0)$);
    \draw [arrow, blue] ($(m1b.west)+(0,-0.55)$) -- (m1b.west);
        
    \path ([xshift=-10mm,yshift=10mm]current bounding box.center) node[matrix,anchor=north west,cells={nodes={font=\tiny,anchor=west}}, draw,thick,inner sep=1ex, fill=white]{
        \node(ALegend)[buffer, APattern, anchor=east] at (0,0){};\draw[arrow, black!25!green](ALegend.east) -- ++ (0.6,0); & \node{$A$ dataflow.};\\
        \node(BLegend)[buffer, BPattern, anchor=south] at (0,0){};\draw[arrow, blue](BLegend.south) -- ++ (0.6,0); & \node{$B$ dataflow.};\\
    };


\end{tikzpicture}
    \vspace{-4mm}
    \caption{$4\times 1$ herd.}
    \label{fig:tiling_4x1}
  \end{subfigure}
  \hfill
  \hspace{-20mm}
  \begin{subfigure}[t]{0.30\textwidth}
    \centering
    \begin{tikzpicture}[thick, node distance=2.8cm]
    \tikzstyle {arrow} = [->, >=stealth]
    \tikzstyle {buffer} = [minimum width=2mm, minimum height=5mm, align=center, draw]
    \tikzstyle {dot} = [circle, fill=black, inner sep=0.1mm, draw]
    \tikzstyle{TextBox} = [rectangle, text centered, minimum height=1em, minimum width=10em, text width=9em, scale=0.8]
    \tikzstyle{APattern} = [pattern=north west lines, pattern color=black!25!green]
    \tikzstyle{BPattern} = [pattern=horizontal lines, pattern color=blue, rotate=90]

    \newcommand{\xseparation}{1.6}
    \newcommand{\yseparation}{1.2}


    \node (c11a) [buffer, APattern] at (0,0) {};
    \node (c11b) [buffer, BPattern] at ($(c11a.center)+(0.5, 0)$) {};

    \node (c12a) [buffer, APattern] at ($(c11a.center)+(0.0, \yseparation)$) {};
    \node (c12b) [buffer, BPattern] at ($(c11b.center)+(0.0, \yseparation)$) {};

    \node (c21a) [buffer, APattern] at ($(c11a.center)+(\xseparation, 0.0)$) {};
    \node (c21b) [buffer, BPattern] at ($(c11b.center)+(\xseparation, 0.0)$) {};

    \node (c22a) [buffer, APattern] at ($(c12a.center)+(\xseparation, 0.0)$) {};
    \node (c22b) [buffer, BPattern] at ($(c12b.center)+(\xseparation, 0.0)$) {};

    \node (m1a) [buffer, APattern] at ($(c11a.center)+(-0.2, -3.8)$) {};
    \node (m2a) [buffer, APattern] at ($(m1a.center)+(0.25, 0.0)$) {};
    \node (m1b) [buffer, BPattern] at ($(m1a.center)+(0.8, 0.125)$) {};
    \node (m2b) [buffer, BPattern] at ($(m1b.center)+(0.0, -0.25)$) {};

    \node (pt1) at ($(m1a.north)+(0.0, 0.2)$) {};
    \node (pt2) at ($(c12a.south)+(0.0, -0.2)$) {};
    \node (pt3) at (pt1.center |- pt2.center) {};
    \node (pt4) at ($(c11a.south)+(0.0, -0.2)$) {};
    \node (pt5) [dot, black!25!green] at (pt1.center |- pt4.center) {};
    
    \draw [arrow, black!25!green] (m1a.north) -- (pt1.center) -- (pt3.center) -- (pt2.center) -- (c12a.south);
    \draw [arrow, black!25!green] (pt5.center) -- (pt4.center) -- (c11a.south);
    
    \node (pt6) at ($(m2a.north)+(0.0, 0.2)$) {};
    \node (pt7) at ($(c22a.south)+(0.0, -0.2)$) {};
    \node (pt8) at ($(pt7.center)+(-0.2, 0.0)$) {};
    \node (pt9) [dot, black!25!green] at (pt6.center -| pt8.center) {};
    \node (pt10) at ($(c21a.south)+(0.0, -0.2)$) {};
    \node (pt11) at (pt6.center |- pt10.center) {};
    
    \draw [arrow, black!25!green] (m2a.north) -- (pt6.center) -- (pt9.center) -- (pt8.center) -- (pt7.center) -- (c22a.south);
    \draw [arrow, black!25!green] (pt9.center) -| (c21a.south);
    
    \node (pt12) at ($(m1b.south)+(0.10, 0.0)$) {};
    \draw [arrow, blue] (m1b.south) -- (pt12.center) |- (c11b.south);
    \node (pt13) [dot, blue] at (pt12.center |- c11b.south) {};
    \draw [arrow, blue] (pt13.center)  |- (c12b.south);
    
    \node (pt14) [dot, blue] at ($(c21b.south)+(0.20, 0.0)$) {};
    \draw [arrow, blue] (m2b.south) -| (pt14.center) -- (c21b.south);
    \draw [arrow, blue] (pt14.center)  |- (c22b.south);

    \newcommand{\borderleft}{0.4}
    \newcommand{\borderright}{0.6}
    \newcommand{\borderabove}{0.5}
    \newcommand{\borderbelow}{0.5}
    \draw [dashed] ($(c11a.center)+(-\borderleft,-\borderbelow)$) |- ($(c11b.center)+(\borderright,\borderabove)$) |- ($(c11a.center)+(-\borderleft,-\borderbelow)$);
    \draw [dashed] ($(c12a.center)+(-\borderleft,-\borderbelow)$) |- ($(c12b.center)+(\borderright,\borderabove)$) |- ($(c12a.center)+(-\borderleft,-\borderbelow)$);
    \draw [dashed] ($(c21a.center)+(-\borderleft,-\borderbelow)$) |- ($(c21b.center)+(\borderright,\borderabove)$) |- ($(c21a.center)+(-\borderleft,-\borderbelow)$);
    \draw [dashed] ($(c22a.center)+(-\borderleft,-\borderbelow)$) |- ($(c22b.center)+(\borderright,\borderabove)$) |- ($(c22a.center)+(-\borderleft,-\borderbelow)$);
    \node [TextBox, right, anchor=center, rotate=90, text=white] at ($(c11a.west)+(-0.8,0.5)$) {Core tiles};
    \draw [dashed] ($(m1a.center)+(-0.3,-0.6)$) |- ($(m1a.center)+(1.4,0.6)$) |- ($(m1a.center)+(-0.3,-0.6)$);
    
    \node [TextBox, right, anchor=center, rotate=90, text=white] at ($(m1a.west)+(-0.6,0.0)$) {Mem. tile};

    \draw [arrow, black!25!green] ($(m1a.south)+(0.125,-0.41)$) -- ($(m1a.south)+(0.125,0.0)$);
    \draw [arrow, blue] ($(m2b.west)+(0,-0.41)$) -- (m2b.west);


\end{tikzpicture}
    \vspace{-4mm}
    \caption{$2\times 2$ herd.}
    \label{fig:tiling_2x2}
  \end{subfigure}

  \vspace{6mm}
  
  \begin{subfigure}[t]{0.33\textwidth}
    \centering
    \begin{tikzpicture}[thick,xscale=0.83, yscale=0.5, every node/.style={scale=0.7}]

\definecolor{darkgreen}{RGB}{27,94,32}
\definecolor{lightgreen}{RGB}{174,213,129}

\tikzstyle{APattern} = [pattern=north west lines, pattern color=black!25!green]
\tikzstyle{BPattern} = [pattern=horizontal lines, pattern color=blue]

\foreach \y in {1, 2, 3, 4, 5, 6, 7, 8} {
    \draw[gray!30, thin] (-0.5,\y) -- (5,\y);
}

\pgfmathsetmacro{\yPORTzero}{7}
\pgfmathsetmacro{\yPORTone}{6}
\pgfmathsetmacro{\yPORTtwo}{5}
\pgfmathsetmacro{\yPORTthree}{4}
\pgfmathsetmacro{\yPORTfour}{3}
\pgfmathsetmacro{\yPORTfive}{2}
\pgfmathsetmacro{\yPORTsix}{1}

\def\barheight{0.7}

\draw[APattern] (0,\yPORTzero) rectangle (1,\yPORTzero+\barheight);

\draw[APattern] (1,\yPORTone) rectangle (2,\yPORTone+\barheight);

\draw[BPattern] (0,\yPORTtwo) rectangle (1,\yPORTtwo+\barheight);
\draw[BPattern] (1.05,\yPORTtwo) rectangle (2,\yPORTtwo+\barheight);
\draw[BPattern] (2.05,\yPORTtwo) rectangle (3,\yPORTtwo+\barheight);
\draw[BPattern] (3.05,\yPORTtwo) rectangle (4,\yPORTtwo+\barheight);

\draw[BPattern] (4,\yPORTthree) rectangle (5,\yPORTthree+\barheight);
\draw[BPattern] (4,\yPORTfour) rectangle (5,\yPORTfour+\barheight);
\draw[BPattern] (4,\yPORTfive) rectangle (5,\yPORTfive+\barheight);
\draw[BPattern] (4,\yPORTsix) rectangle (5,\yPORTsix+\barheight);

\node[anchor=east] at (0,\yPORTzero+0.3) {\small W0};
\node[anchor=east] at (0,\yPORTone+0.3) {\small R0};
\node[anchor=east] at (0,\yPORTtwo+0.3) {\small W1};
\node[anchor=east] at (0,\yPORTthree+0.3) {\small R1};
\node[anchor=east] at (0,\yPORTfour+0.3) {\small R2};
\node[anchor=east] at (0,\yPORTfive+0.3) {\small R3};
\node[anchor=east] at (0,\yPORTsix+0.3) {\small R4};

\end{tikzpicture}
    \vspace{-2mm}
    \caption{Memory tile trace for (\subref{fig:tiling_1x4}).}
    \label{fig:mm_tiling_trace_1x4}
  \end{subfigure}
  \hfill
  \hspace{-5mm}
  \begin{subfigure}[t]{0.33\textwidth}
    \centering
    \begin{tikzpicture}[thick,xscale=0.83, yscale=0.5, every node/.style={scale=0.7}]

\definecolor{darkgreen}{RGB}{27,94,32}
\definecolor{lightgreen}{RGB}{174,213,129}

\tikzstyle{APattern} = [pattern=north west lines, pattern color=black!25!green]
\tikzstyle{BPattern} = [pattern=horizontal lines, pattern color=blue]

\foreach \y in {1, 2, 3, 4, 5, 6, 7, 8} {
    \draw[gray!30, thin] (-0.5,\y) -- (5,\y);
}

\pgfmathsetmacro{\yPORTzero}{7}
\pgfmathsetmacro{\yPORTone}{6}
\pgfmathsetmacro{\yPORTtwo}{5}
\pgfmathsetmacro{\yPORTthree}{4}
\pgfmathsetmacro{\yPORTfour}{3}
\pgfmathsetmacro{\yPORTfive}{2}
\pgfmathsetmacro{\yPORTsix}{1}

\def\barheight{0.7}

\draw[APattern] (0,\yPORTzero) rectangle (1,\yPORTzero+\barheight);
\draw[APattern] (1.05,\yPORTzero) rectangle (2,\yPORTzero+\barheight);
\draw[APattern] (2.05,\yPORTzero) rectangle (3,\yPORTzero+\barheight);
\draw[APattern] (3.05,\yPORTzero) rectangle (4,\yPORTzero+\barheight);

\draw[APattern] (4,\yPORTone) rectangle (5,\yPORTone+\barheight);

\draw[APattern] (4,\yPORTtwo) rectangle (5,\yPORTtwo+\barheight);

\draw[APattern] (4,\yPORTthree) rectangle (5,\yPORTthree+\barheight););

\draw[APattern] (4,\yPORTfour) rectangle (5,\yPORTfour+\barheight);

\draw[BPattern] (0,\yPORTfive) rectangle (1,\yPORTfive+\barheight);

\draw[BPattern] (1,\yPORTsix) rectangle (2,\yPORTsix+\barheight);

\node[anchor=east] at (0,\yPORTzero+0.3) {\small W0};
\node[anchor=east] at (0,\yPORTone+0.3) {\small R0};
\node[anchor=east] at (0,\yPORTtwo+0.3) {\small R1};
\node[anchor=east] at (0,\yPORTthree+0.3) {\small R2};
\node[anchor=east] at (0,\yPORTfour+0.3) {\small R3};
\node[anchor=east] at (0,\yPORTfive+0.3) {\small W1};
\node[anchor=east] at (0,\yPORTsix+0.3) {\small R4};

\end{tikzpicture}
    \vspace{-2mm}
    \caption{Memory tile trace for (\subref{fig:tiling_4x1}).}
    \label{fig:mm_tiling_trace_4x1}
  \end{subfigure}
  \hfill
  \hspace{-5mm}
  \begin{subfigure}[t]{0.33\textwidth}
    \centering
    \begin{tikzpicture}[thick,xscale=0.83, yscale=0.5, every node/.style={scale=0.7}]

\definecolor{darkgreen}{RGB}{27,94,32}
\definecolor{lightgreen}{RGB}{174,213,129}

\tikzstyle{APattern} = [pattern=north west lines, pattern color=black!25!green]
\tikzstyle{BPattern} = [pattern=horizontal lines, pattern color=blue]

\foreach \y in {1, 2, 3, 4, 5, 6, 7} {
    \draw[gray!30, thin] (-0.5,\y) -- (5,\y);
}

\pgfmathsetmacro{\yPORTzero}{6}
\pgfmathsetmacro{\yPORTone}{5}
\pgfmathsetmacro{\yPORTtwo}{4}
\pgfmathsetmacro{\yPORTthree}{3}
\pgfmathsetmacro{\yPORTfour}{2}
\pgfmathsetmacro{\yPORTfive}{1}

\def\barheight{0.7}

\draw[APattern] (0,\yPORTzero) rectangle (1,\yPORTzero+\barheight);
\draw[APattern] (1.05,\yPORTzero) rectangle (2,\yPORTzero+\barheight);

\draw[APattern] (2,\yPORTone) rectangle (3,\yPORTone+\barheight);

\draw[APattern] (2,\yPORTtwo) rectangle (3,\yPORTtwo+\barheight);

\draw[BPattern] (0,\yPORTthree) rectangle (1,\yPORTthree+\barheight);
\draw[BPattern] (1.05,\yPORTthree) rectangle (2,\yPORTthree+\barheight);

\draw[BPattern] (2,\yPORTfour) rectangle (3,\yPORTfour+\barheight);

\draw[BPattern] (2,\yPORTfive) rectangle (3,\yPORTfive+\barheight);

\node[anchor=east] at (0,\yPORTzero+0.3) {\small W0};
\node[anchor=east] at (0,\yPORTone+0.3) {\small R0};
\node[anchor=east] at (0,\yPORTtwo+0.3) {\small R1};
\node[anchor=east] at (0,\yPORTthree+0.3) {\small W1};
\node[anchor=east] at (0,\yPORTfour+0.3) {\small R2};
\node[anchor=east] at (0,\yPORTfive+0.3) {\small R3};

\end{tikzpicture}
    \vspace{-2mm}
    \caption{Memory tile trace for (\subref{fig:tiling_2x2}).}
    \label{fig:mm_tiling_trace_2x2}
  \end{subfigure}
  \caption{
    Impact of tiling strategy on data movement schedule in an output-stationary matrix multiplication.
    See Listing~\ref{lst:matmul_loop_nest} for its schedule described using loop nests, where the \texttt{for\_all} loops \texttt{ii} and \texttt{jj} were mapped to horizontal and vertical directions in the two-dimensional \herd operations, respectively.
  }
  \label{fig:matmul_tiling}
\end{figure}
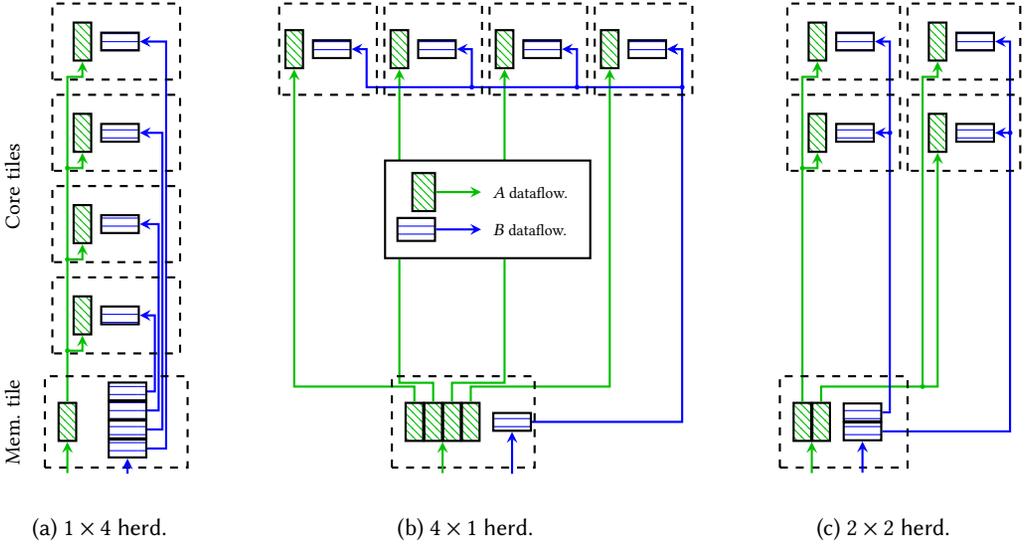
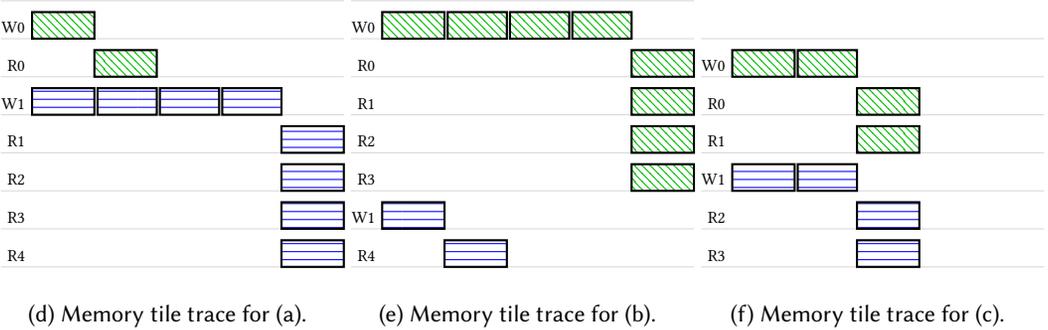

In Figure~\ref{fig:matmul_tiling}, we examine how tiling strategies in \tech influence the scheduling efficiency for a tiled matrix multiplication ($A \times B = C$, where $A\in \mathbb{R}^{M{\times}K}$, $B \in \mathbb{R}^{K{\times}N}$, and $C\in \mathbb{R}^{M{\times}N}$) mapped to AMD NPUs.
We consider a matrix multiplication problem tiled along the $M$ and $N$ dimensions and mapped to three spatial layouts: 1$\times$4, 2$\times$2, and 4$\times$1 \herd operations. 
These configurations are selected to cover a range of aspect ratios and communication patterns.
In the 1$\times$4 layout, matrix $A$ is broadcast across the four column-aligned cores, while matrix $B$ is privately transferred to each core via separate dataflows. 
The 4$\times$1 layout exhibits the complementary pattern: $B$ is broadcast across rows, but $A$ must be duplicated. In contrast, the 2$\times$2 layout enables two-dimensional reuse: $A$ is broadcast column-wise, and $B$ row-wise, minimizing redundant transfers.
Data reuse patterns are automatically inferred by \tech (detailed in Section~\ref{subsec:bcast_detect_and_lowering}).

Figure~\ref{fig:matmul_tiling} presents the runtime traces of each strategy, with an assumption of equal tile sizes for $A$ and $B$.
In the 1$\times$4 and 4$\times$1 cases, imbalance in data streaming---due to one matrix requiring separate transfers---results in core stalls as execution waits for both inputs to arrive. 
By contrast, the 2$\times$2 configuration shows reduced stalls and improved throughput owing to symmetric broadcast reuse on both $A$ and $B$ paths.

Performance bottlenecks in asymmetric tilings can be mitigated by rebalancing tile sizes to equalize data transfer volumes or by allocating additional DMA channels to heavier dataflows. 
The latter can be automated through \tech's dataflow-aware bufferization (see Section~\ref{subsubsec:memref_splitting}).

This case study highlights how \tech enables fast tile-shape-aware schedule selection through explicit representation of data dependencies and broadcast opportunities, guiding design-space exploration for spatial platforms.

\subsection{Broadcast Detection and Lowering}
\label{subsec:bcast_detect_and_lowering}

Once tiling has spatially partitioned the computational workload, the next optimization opportunity lies in minimizing off-chip data movement through on-chip reuse. 
To achieve this, \tech introduces a systematic way to identify and optimize data broadcasting patterns in the tiled problem space.
Compiler passes work in tandem to both detect opportunities and generate optimized \air code that explicitly captures such patterns using affine maps.

\subsubsection{Broadcast Detection}

The broadcast detection pass performs static analysis on the iteration domain of the program to discover replication patterns in data movements. 
When detected, the pass annotates the data movement operation with an affine set representing a projection in the spatial iteration domains from the sources to broadcast destinations.
For example, an affine set $S0$ representing a broadcast on two-dimensional spatial iterations, e.g., an \herd, where an array of $4\times1$ \memcpy sources broadcast to $4\times4$ destinations, has the form $\{\left(d_0,d_1\right) \in \mathbb{Z}^2 | \exists s_0 \in \mathbb{Z} : d_0 = s_0, 0 \leq s_0 \leq 3, 0 \leq d_1 \leq 3\}$, where the symbol $s_0$ and dimensions $d_0$ and $d_1$ represent the source and destination spaces, respectively.

By expressing this as an affine set in MLIR's \affine dialect, the \air dialect retains a precise and analyzable description of the communication pattern, which remains composable with other open-source MLIR dialects thanks to the community-developed \affine dialect utilities.

\subsection{Asynchronous Dependency Analysis}
\label{subsec:async_dep_analysis}

\tech captures asynchronous parallelism using ACDG, represented inline of an MLIR codebase via the SSA \token, which tracks the execution ordering and ensures correctness.

\subsubsection{Capturing Dependencies using ACDGs}
\label{subsubsec:capturing_acdg}


In the synchronous code snippet shown in Listing~\ref{lst:sync}, each data movement is implicitly blocked by the previous one, leading to a sequential schedule.
However, both DMA operations could theoretically execute simultaneously, assuming independent DMA resources.
\tech automatically analyzes memory references, identifies these implicit dependencies, and explicitly annotates synchronization tokens as shown in Listing~\ref{lst:async}.


\begin{figure}[ht]
\centering
\begin{minipage}{0.48\textwidth}
\begin{lstlisting}[language=mlir, caption={Synchronous (sequential) execution.}, label=lst:sync, escapechar=|]
air.memcpy (%v1, %v2)|\label{line:dep_sync_begin}|
air.memcpy (%v3, %v4)
func.call @func(%v1, %v3)|\label{line:dep_sync_end}|
\end{lstlisting}
\end{minipage}%
\hfill
\begin{minipage}{0.48\textwidth}
\begin{lstlisting}[language=mlir, caption={Explicit asynchronous dependencies.}, label=lst:async, escapechar=|]
%t1 = air.memcpy async (%v1, %v2)|\label{line:dep_async_begin}|
%t2 = air.memcpy async (%v3, %v4)
%t3 = func.call async
    deps=[%t1, %t2] @func(%v1, %v3)|\label{line:dep_async_end}|
\end{lstlisting}
\end{minipage}
\label{fig:air_dependency_compare}
\end{figure}



This explicit representation clearly indicates that the compute operation must wait for \textit{both DMA transfers} to complete before proceeding, preserving correctness. 
Furthermore, it also makes evident that the two DMA operations can execute in parallel, effectively leveraging multiple discrete DMA resources. 
In \tech, we provide compiler passes which, driven by MLIR's native SSA representation and dominance analysis, automatically capture the ACDG arising from the read-after-write, write-after-read and write-after-write dependencies between MLIR operations, providing robust guarantees of correctness in \tech's scheduling optimizations.

\head{Loop-Carried Dependency in ACDG}
In conventional compiler analysis, loop-carried dependencies are often represented using dependence polyhedra of the form ${\left(i, j, k\right) \to \left(i', j', k'\right)}$, capturing legal source and destination iteration pairs that must respect data dependence across loops.
Compilers such as PLUTO~\cite{PLUTO} and work by Baskaran~\emph{et al.}~\cite{POLYOPT_GPGPU} typically model tiling and execution at the level of atomic tiles, where dependencies across tiles dictate scheduling order, while dependencies within a tile are assumed to be resolved independently through external memory accesses.
This treatment simplifies global scheduling but leaves intra-tile parallelism and fine-grained asynchronous scheduling opportunities underexplored.

Extending beyond this model, \tech's ACDG captures dependencies at the operation level---both within and across loop iterations.
As illustrated in Figure~\ref{fig:scf_for_loop_carried}, loop-carried dependencies are explicitly represented by passing \token values (explicit synchronization handles) through the iteration arguments of \scffor loops, tracking per-iteration execution states.
An arbitrary number of \token values can be carried across iterations, each representing an independently progressing thread of execution.
This enables precise modeling of parallel pipelines, race conditions, and shared resource usage, as illustrated in Figure~\ref{fig:scf_for_race}.
This mechanism is particularly valuable in hardware pipelining, where producer and consumer stages can overlap in time (e.g., using ping-pong buffering; see Section~\ref{subsubsec:pingpong}).

\head{Representation of Asynchronous Dependencies via \token in \scfpar}
Similarly, \tech supports asynchronous dependencies within structured parallel execution constructs such as \scfpar. 
Visualized by Figure~\ref{fig:scf_par_loop_reduced}, \tech explicitly handles the synchronized initialization of parallel threads via an \token passed into the initialization argument of the loop; their synchronized termination is represented explicitly via a reduction tree of \waitall barriers.

\begin{figure}[h]
  \centering
  \hspace{-10mm}
  \begin{subfigure}[t]{\textwidth}
    \centering
    \input{diagrams/for_par_side_by_side}
    \label{fig:loop_carried_dep}
  \end{subfigure}
  \begin{minipage}[c]{0.48\textwidth}
  \begin{subfigure}[t]{\textwidth}
    \centering
    \includegraphics[width=\textwidth]{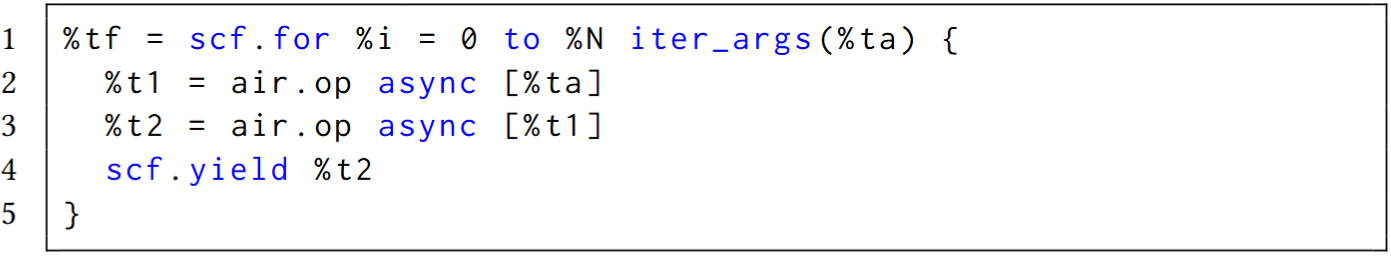}
    \caption{Code for (\subref{fig:scf_for_loop_carried}).}
    \label{fig:scf_for_loop_carried_code}
  \end{subfigure}
  \begin{subfigure}[t]{\textwidth}
    \centering
    \includegraphics[width=\textwidth]{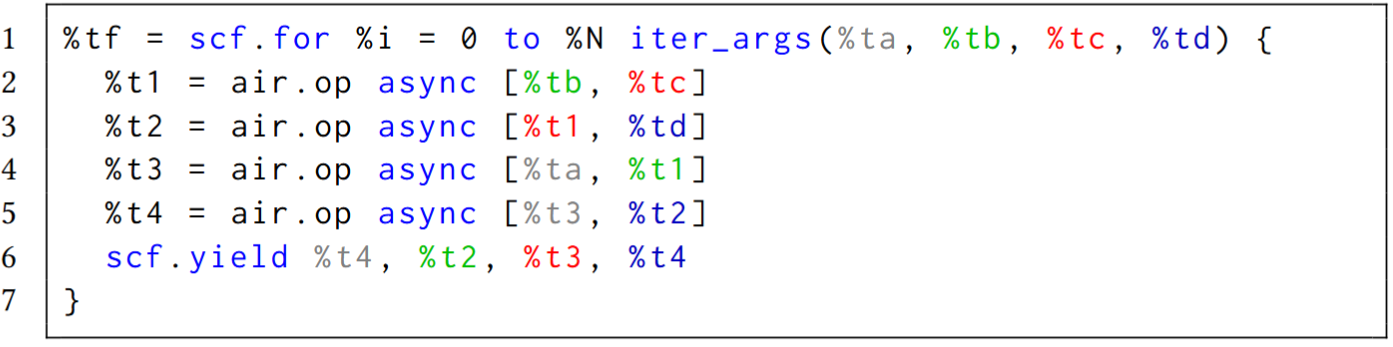}
    \caption{Code for (\subref{fig:scf_for_race}).}
    \label{fig:scf_for_race_code}
  \end{subfigure}
  \end{minipage}%
  \begin{minipage}[c]{0.48\textwidth}
  \begin{subfigure}[t]{\textwidth}
    \centering
    \includegraphics[width=\textwidth]{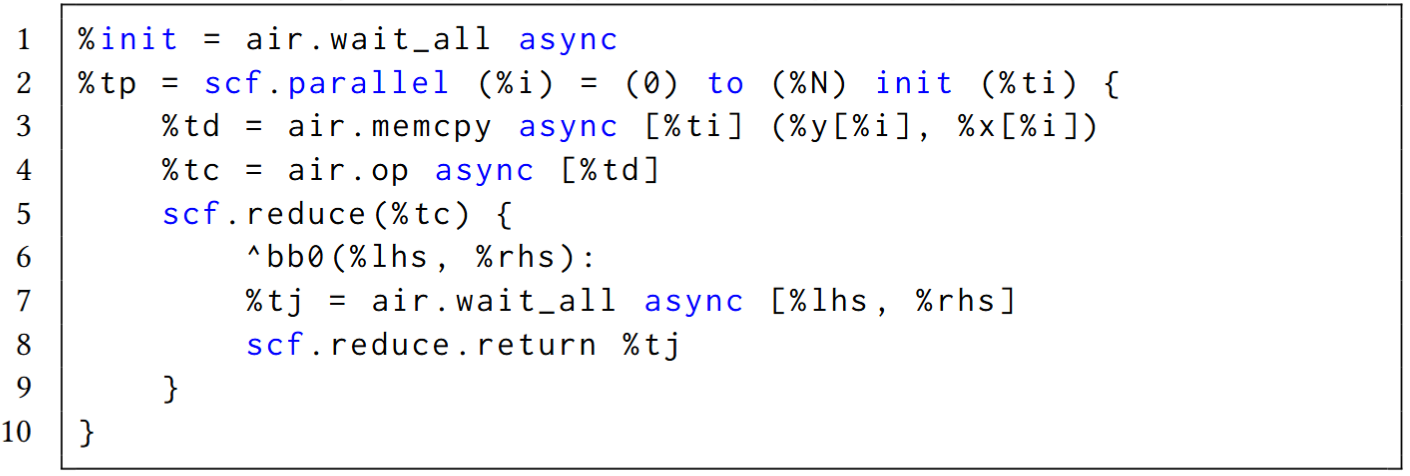}
    \caption{Code for (\subref{fig:scf_par_loop_reduced}).}
    \label{fig:scf_par_loop_reduced_code}
  \end{subfigure}
  \end{minipage}%
    \caption{
        Visualizations of ACDGs in loop iterations, including (\subref{fig:scf_for_loop_carried})~sequentialized for loop, (\subref{fig:scf_for_race})~multi-token for loop, and (\subref{fig:scf_par_loop_reduced})~parallel loop, and their respective \tech specification.
        A circle represents an \token, and a polygon represents a group of MLIR operations in the loop body.
        Listings~(\subref{fig:scf_for_loop_carried_code}---\subref{fig:scf_par_loop_reduced_code}) demonstrates how each ACDG is represented inline of MLIR code.
    }
  \label{fig:buffer_split_comparison}
\end{figure}

\subsubsection{Reasoning using ACDGs}

Building on the previous section on ACDG extraction, we now describe how \tech progressively transforms a generic loop-based program into finer-grained asynchronous schedules by analyzing and restructuring its control and data dependencies.

\begin{figure}[h]
  \centering
  \hspace{-10mm}
  \input{diagrams/acdg_split_loop_nest}
  \caption{
      Visualizations of ACDGs in loop iterations, including (\subref{fig:scf_for_loop_carried})~sequentialized for loop, (\subref{fig:scf_for_race})~multi-token for loop, and (\subref{fig:scf_par_loop_reduced})~parallel loop, and their respective \tech specification.
  }
  \label{fig:acdg_split_loop_nest}
\end{figure}

Figure~\ref{fig:acdg_split_loop_nest} illustrates this process on an imperfect loop nest. 
In the original synchronous form (Figure~\ref{fig:acdg_sync_nest}), the loop bodies imply a fully sequential dataflow in the absence of explicit parallelism annotations. 
Nevertheless, the underlying ACDG reveals opportunities for parallelism, as operations can be partitioned based on the memory buffers they access (annotated by colors).

\tech first applies asynchronous dependency analysis to construct an explicit ACDG using loop-carried \token values within each loop (Figure~\ref{fig:acdg_async_nest}). 
This step exposes parallelism between sub-graphs of each loop's body accessing distinct buffers, while preserving correctness through token synchronization.

To further expose optimization opportunities, \tech splits the asynchronous loop nest into multiple independent nests (Figure~\ref{fig:acdg_async_nest_split}), each exclusively operating on a single memory object. 
This restructuring systematically uncovers and amplifies the spatial parallelism latent in generic loop-based input programs, isolating them into dataflows which facilitate compiler optimizations.

\subsection{Inferring Dataflow via \channel}
\label{subsec:inferring_dataflow_channel}

The ACDG structure not only enables fine-grained parallelism analysis, but also serves as the foundation for identifying and scheduling data movement across disjoint memory spaces. 
\air's channel-based abstraction makes such communication patterns explicit and analyzable.

\begin{figure}[h]
  \centering
  \hspace{-10mm}
  \begin{tikzpicture}[thick, node distance=2.8cm]
    \newcommand{\inputlen}{3mm}
    \newcommand{\outputlen}{3mm}
    \tikzstyle {arrow} = [->, >=stealth]
    \tikzstyle {token} = [circle, minimum width=6mm, minimum height=6mm, align=center, draw]
    \tikzstyle {op} = [minimum width=9mm, minimum height=6mm, align=center, draw]
    \tikzstyle {body} = [chamfered rectangle, chamfered rectangle xsep=2cm, rotate=-90, draw]
    \tikzstyle{TextBox} = [rectangle, text centered, minimum height=1em, minimum width=10em, text width=9em, font=\tiny]


    \node (de_a) [op] at (0,0) {};
    \node [TextBox, anchor=center, scale=0.8] at (de_a) {{\tt dealloc} $a$};
    \node (de_cp) [op] at ($(de_a.center)+(-0.5,1.0)$) {};
    \node [TextBox, anchor=center, scale=0.8] at (de_cp) {{\tt dealloc} $c^\prime$};
    \node (de_ap) [op] at ($(de_a.center)+(0.5,1.0)$) {};
    \node [TextBox, anchor=center, scale=0.8] at (de_ap) {{\tt dealloc} $a^\prime$};
    \node (func) [op] at ($(de_a.center)+(0.0,1.8)$) {};
    \node [TextBox, anchor=center, scale=0.8] at (func) {{\tt func}};
    \node (al_cp) [op] at ($(de_a.center)+(-0.5,2.6)$) {};
    \node [TextBox, anchor=center, scale=0.8] at (al_cp) {{\tt alloc} $c^\prime$};
    \node (memcpy) [op] at ($(de_a.center)+(0.5,2.6)$) {};
    \node [TextBox, anchor=center, scale=0.8] at (memcpy) {{\tt memcpy}};
    \node (al_ap) [op] at ($(de_a.center)+(0.5,3.4)$) {};
    \node [TextBox, anchor=center, scale=0.8] at (al_ap) {{\tt alloc} $a^\prime$};
    \node (al_a) [op] at ($(de_a.center)+(0.0,4.4)$) {};
    \node [TextBox, anchor=center, scale=0.8] at (al_a) {{\tt alloc} $a$};
    
    \draw [arrow] (al_ap.south) -| (memcpy.north);
    \draw [arrow] (al_cp.south) -- (func.north);
    \draw [arrow] (memcpy.south) -- (func.north);
    \draw [arrow] (func.south) -- (de_cp.north);
    \draw [arrow] (func.south) -- (de_ap.north);

    \draw ($(de_a.center)+(-1.2,0.5)$) |- ($(al_a.center)+(1.2,-0.5)$) |- ($(de_a.center)+(-1.2,0.5)$);
    \node [TextBox, anchor=center] at ($(al_ap.north)+(-1.3,-0.2)$) {\makecell{herd\\$M\times N$}};
    
    \draw [arrow] (al_a.south) -- ($(al_a.center)+(0.0,-0.5)$);
    \draw [arrow] ($(de_a.center)+(0.0,0.5)$) -- (de_a.north);
     
    \node (to1) [right=of func, xshift=-16mm, yshift=5mm] {};
    \draw [fill=black] ([xshift=-5mm, yshift=-5mm]to1.center) -- ([xshift=-5mm, yshift=5mm]to1.center) -- ([yshift=5mm]to1.center) -- ([yshift=7.5mm]to1.center) -- ([xshift=5mm]to1.center) -- ([yshift=-7.5mm]to1.center) -- ([yshift=-5mm]to1.center) -- cycle;


    \node (base2) [op] at ($(de_a.center)+(3.6,-0.35)$) {};
    \node [TextBox, anchor=center, scale=0.8] at (base2) {{\tt dealloc} $a$};
    \node (de_cp2) [op] at ($(base2.center)+(-0.5,1.0)$) {};
    \node [TextBox, anchor=center, scale=0.8] at (de_cp2) {{\tt dealloc} $c^\prime$};
    \node (de_ap2) [op] at ($(base2.center)+(0.5,1.0)$) {};
    \node [TextBox, anchor=center, scale=0.8] at (de_ap2) {{\tt dealloc} $a^\prime$};
    \node (func2) [op] at ($(base2.center)+(0.0,1.8)$) {};
    \node [TextBox, anchor=center, scale=0.8] at (func2) {{\tt func}};
    \node (al_cp2) [op] at ($(base2.center)+(-0.5,2.6)$) {};
    \node [TextBox, anchor=center, scale=0.8] at (al_cp2) {{\tt alloc} $c^\prime$};
    \node (get2) [op, red] at ($(base2.center)+(0.5,2.6)$) {};
    \node [TextBox, anchor=center, scale=0.8] at (get2) {{\tt get}};
    \node (put2) [op, red] at ($(base2.center)+(0.5,3.4)$) {};
    \node [TextBox, anchor=center, scale=0.8] at (put2) {{\tt put}};
    \node (al_ap2) [op] at ($(base2.center)+(0.5,4.2)$) {};
    \node [TextBox, anchor=center, scale=0.8] at (al_ap2) {{\tt alloc} $a^\prime$};
    \node (top2) [op] at ($(base2.center)+(0.0,5.2)$) {};
    \node [TextBox, anchor=center, scale=0.8] at (top2) {{\tt alloc} $a$};
    
    \draw [arrow] (al_ap2.south) -| (put2.north);
    \draw [arrow] (put2.south) -- (get2.north);
    \draw [arrow] (al_cp2.south) -- (func2.north);
    \draw [arrow] (get2.south) -- (func2.north);
    \draw [arrow] (func2.south) -- (de_cp2.north);
    \draw [arrow] (func2.south) -- (de_ap2.north);

    \draw ($(base2.center)+(-1.2,0.5)$) |- ($(top2.center)+(1.2,-0.5)$) |- ($(base2.center)+(-1.2,0.5)$);
    \node [TextBox, anchor=center] at ($(al_ap2.north)+(-1.3,-0.2)$) {\makecell{herd\\$M\times N$}};

    \draw [arrow] (top2.south) -- ($(top2.center)+(0.0,-0.5)$);
    \draw [arrow] ($(base2.center)+(0.0,0.5)$) -- (base2.north);    


    \node (base) at ($(de_a.center)+(7.2,0.0)$) {};
    \node (de_cp1) [op] at ($(base.center)+(-0.5,1.0)$) {};
    \node [TextBox, anchor=center, scale=0.8] at (de_cp1) {{\tt dealloc} $c^\prime$};
    \node (de_ap1) [op] at ($(base.center)+(0.5,1.0)$) {};
    \node [TextBox, anchor=center, scale=0.8] at (de_ap1) {{\tt dealloc} $a^\prime$};
    \node (func1) [op] at ($(base.center)+(0.0,1.8)$) {};
    \node [TextBox, anchor=center, scale=0.8] at (func1) {{\tt func}};
    \node (al_cp1) [op] at ($(base.center)+(-0.5,2.6)$) {};
    \node [TextBox, anchor=center, scale=0.8] at (al_cp1) {{\tt alloc} $c^\prime$};
    \node (get) [op] at ($(base.center)+(0.5,2.6)$) {};
    \node [TextBox, anchor=center, scale=0.8] at (get) {{\tt get}};
    \node (al_ap1) [op] at ($(base.center)+(0.5,3.4)$) {};
    \node [TextBox, anchor=center, scale=0.8] at (al_ap1) {{\tt alloc} $a^\prime$};
    \node (top) at ($(base.center)+(0.0,4.4)$) {};

    \node (de_a1) [op] at ($(base.center)+(3.0,2.4)$) {};
    \node [TextBox, anchor=center, scale=0.8] at (de_a1) {{\tt dealloc} $a$};
    \node (put) [op] at ($(de_a1.center)+(0.0,1.0)$) {};
    \node [TextBox, anchor=center, scale=0.8] at (put) {{\tt put}};
    \node (al_a1) [op] at ($(put.center)+(0.0,1.0)$) {};
    \node [TextBox, anchor=center, scale=0.8] at (al_a1) {{\tt alloc} $a$};
    
    \draw [arrow] (al_ap1.south) -| (get.north);
    \draw [arrow] (al_cp1.south) -- (func1.north);
    \draw [arrow] (get.south) -- (func1.north);
    \draw [arrow] (func1.south) -- (de_cp1.north);
    \draw [arrow] (func1.south) -- (de_ap1.north);

    \draw ($(base.center)+(-1.2,0.5)$) |- ($(top.center)+(1.2,-0.5)$) |- ($(base.center)+(-1.2,0.5)$);
    \node [TextBox, anchor=center] at ($(al_ap1.north)+(-1.3,-0.2)$) {\makecell{herd\\$M\times N$}};

    \draw ($(put.center)+(-1.2,0.5)$) |- ($(put.center)+(0.7,-0.5)$) |- ($(put.center)+(-1.2,0.5)$);
    \node [TextBox, anchor=center] at ($(put.north)+(-0.82,-0.1)$) {\makecell{parallel\\$M\times N$}};
    
    \draw [arrow] (al_a1.south) -- (put.north);
    \draw [arrow] (put.south) -- (de_a1.north);
    \draw [arrow, dashed] (put.west) -- (get.east);
     
    \node (to2) [right=of func, xshift=20mm, yshift=5mm] {};
    \draw [fill=black] ([xshift=-5mm, yshift=-5mm]to2.center) -- ([xshift=-5mm, yshift=5mm]to2.center) -- ([yshift=5mm]to2.center) -- ([yshift=7.5mm]to2.center) -- ([xshift=5mm]to2.center) -- ([yshift=-7.5mm]to2.center) -- ([yshift=-5mm]to2.center) -- cycle;
	
    \captionsetup[sub]{font=Large}
    \node (cap1) [text width=15em] at ([xshift=0mm, yshift=-10mm]de_a) {\subcaption{Coupled {\tt memcpy}.\label{fig:before_channel_decouple}}};
    \node (cap2) [text width=15em, xshift=0mm] at (cap1 -| base2) {\subcaption{Decoupled {\tt put/get}.\label{fig:after_channel_decouple}}};
    \node (cap3) [text width=15em, xshift=10mm] at (cap1 -| base) {\subcaption{Decoupled {\tt put/get}, {\tt put} hoisted.\label{fig:after_channel_decouple}}};
        
    \path ([xshift=0mm,yshift=7mm]current bounding box.south east) node[matrix,anchor=south east,cells={nodes={font=\tiny,anchor=west}}, draw,thick,inner sep=1ex, row sep=-1mm]{
        \node [op, minimum width=6mm, minimum height=4mm]{}; & \node{Operation.};\\
        \draw[arrow](0,0) -- ++ (0.6,0); & \node{\makecell[l]{\token\\dependency.}};\\
        \draw[arrow,dashed](0,0) -- ++ (0.6,0); & \node{\makecell[l]{\channel\\dependency.}};\\
    };

\end{tikzpicture}
  \caption{Visualizations of ACDGs before and after \channel decoupling.}
  \label{fig:channel_decouple}
\end{figure}
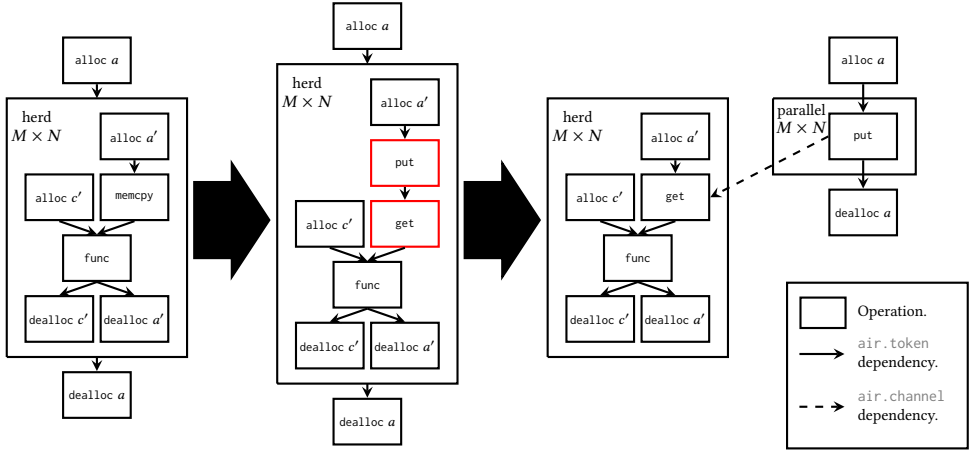

Figure~\ref{fig:channel_decouple} illustrates this transformation using ACDGs.
In the pre-transformation ACDG shown in Figure~\ref{fig:before_channel_decouple}, a \texttt{memcpy} operation moves data from a shared buffer $a$ into a local buffer $a^\prime$, which is subsequently consumed by a compute kernel.
Because \texttt{memcpy} resides within the body of the \herd, the producer and consumer of $a^\prime$ are tightly coupled within a single hierarchical region.
While this correctly expresses intra-herd dependencies, it fails to expose the fine-grained asynchronous boundary between the shared memory and local memory regions.
As a result, the external thread managing $a$ remains blocked until the entire \herd completes---despite the fact that only the \texttt{memcpy} operation requires synchronization.

The transformed ACDG in Figure~\ref{fig:after_channel_decouple} resolves this limitation by replacing \texttt{memcpy} with a decoupled pair of \channelput and \channelget operations.
These operations are hoisted to the respective regions associated with the source and destination memory, each integrated into its own ACDG subgraph via explicit \token synchronization.
To preserve the correctness of the original computation, the hoisted \texttt{put} operation must replicate the parallel semantics of the original \texttt{memcpy}; if the \texttt{memcpy} was nested within a $M\times N$ \herd, then the corresponding \texttt{put} must be nested within a matching $M\times N$ \scfpar loop.
The dashed arrow between \channelget and \channelput represents the data stream back pressure; an overlapping schedule across two sides is enabled if stream buffering is supported in hardware.
This ensures that data produced and consumed match in size across hierarchies.

This decoupling of `put' from `get' not only enables overlapping communication and execution but also allows the compiler to infer and instantiate multiple parallel dataflows---subject to available bandwidth and communication resources.
When hardware permits, the compiler may emit parallel \channel instances, increasing aggregate throughput and improving hardware utilization. 
In this setting, data movement is no longer serialized at the herd boundary, and bandwidth can scale with the degree of inferred parallelism.

Furthermore, the use of \channel operations allows multiple data movement operations to share communication resources.
This enables hardware-aware optimizations such as \channel arbitration in pipelined execution (see Section~\ref{subsubsec:pingpong}) and \channel reuse (see Section~\ref{subsubsec:channel_fusion}). 

\subsubsection{Capturing Hardware Pipelining with \channel in ACDG}
\label{subsubsec:pingpong}


Building on the fine-grained asynchronous representations introduced in Section~\ref{subsubsec:capturing_acdg} and the decoupled \channel abstraction, \tech captures hardware pipelining by leveraging the loop-carried \token semantics in ACDG. 
By allowing multiple tokens to flow independently across iterations, \tech models the three key dependencies in a hardware pipeline: \emph{(i)} producer-consumer data dependencies, \emph{(ii)} producer-side resource contention and \emph{(iii)} consumer-side resource contention, all at once. 

\begin{figure}[h]
  \centering
  \begin{tikzpicture}[thick, node distance=2.8cm]
    \newcommand{\inputlen}{3mm}
    \newcommand{\outputlen}{3mm}
    \tikzstyle {arrow} = [->, >=stealth]
    \tikzstyle {token} = [circle, minimum width=6mm, minimum height=6mm, align=center, draw]
    \tikzstyle {op} = [minimum width=6mm, minimum height=4mm, align=center, draw]
    \tikzstyle {body} = [signal, draw, fill=white, font=\tiny, signal to=west and east, align=center, minimum height=1em, minimum width=1em]
    \tikzstyle{TextBox} = [rectangle, text centered, minimum height=1em, minimum width=10em, text width=9em, font=\tiny]


    \node (ping_prod1) [body] at (0,0) {`ping' producer ops.};
    \node (ping_cons1) [body] at ($(ping_prod1.east)+(1.5,0.0)$) {`ping' consumer ops.};
    \node (ping_prod2) [body] at ($(ping_cons1.east)+(1.5,0.0)$) {`ping' producer ops.};
    \node (ping_cons2) [body] at ($(ping_prod2.east)+(1.5,0.0)$) {`ping' consumer ops.};
    \node (pong_prod1) [body] at ($(ping_prod1.east)+(1.5,-0.8)$) {`pong' producer ops.};
    \node (pong_cons1) [body] at ($(pong_prod1.east)+(1.5,0.0)$) {`pong' consumer ops.};
    \node (pong_prod2) [body] at ($(pong_cons1.east)+(1.5,0.0)$) {`pong' producer ops.};
    \node (pong_cons2) [body] at ($(pong_prod2.east)+(1.5,0.0)$) {`pong' consumer ops.};
    \draw [arrow, black!25!green, dashdotted] (ping_prod1.east) -- (ping_cons1.west);
    \draw [arrow, black!25!green, dashdotted] (ping_cons1.east) -- (ping_prod2.west);
    \draw [arrow, black!25!green, dashdotted] (ping_prod2.east) -- (ping_cons2.west);
    \draw [arrow, gray, dashed] (pong_prod1.east) -- (pong_cons1.west);
    \draw [arrow, gray, dashed] (pong_cons1.east) -- (pong_prod2.west);
    \draw [arrow, gray, dashed] (pong_prod2.east) -- (pong_cons2.west);

    \draw [arrow, red] (ping_prod1.east) -- (pong_prod1.west);
    \draw [arrow, blue, dotted] (ping_cons1.east) -- (pong_cons1.west);
    \draw [arrow, red] (ping_prod2.east) -- (pong_prod2.west);
    \draw [arrow, blue, dotted] (ping_cons2.east) -- (pong_cons2.west);
    
    \draw [arrow, red] (pong_prod1.east) -- (ping_prod2.west);
    \draw [arrow, blue, dotted] (pong_cons1.east) -- (ping_cons2.west);
    
    \draw [arrow, black!25!green, dashdotted] ($(ping_prod1.west)+(-0.2,0.0)$) -- (ping_prod1.west);
    \draw [arrow, gray, dashed] ($(pong_prod1.west)+(-0.2,0.0)$) -- (pong_prod1.west);
    \draw [arrow, black!25!green, dashdotted] (ping_cons2.east) -- ($(ping_cons2.east)+(0.2,0.0)$);
    \draw [arrow, gray, dashed] (pong_cons2.east) -- ($(pong_cons2.east)+(0.2,0.0)$);
\end{tikzpicture}
  \caption{
      Flattened ACDG showing a ping-pong buffering schedule, specialized from a generic ACDG form in Figure~\ref{fig:scf_for_race}.
  }
  \label{fig:pipeline_acdg_flattened}
\end{figure}
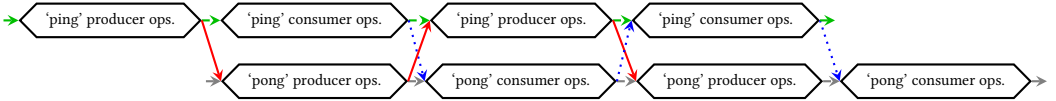

As a motivating example, we consider two-stage pipelining, commonly referred to as ping-pong buffering.
To expose pipeline stages explicitly, the loop must first be unrolled by a factor of two, corresponding to the number of stages, yielding distinct ping and pong threads for ACDG annotation. 
The resulting structure maps naturally to the generic ACDG with multiple loop-carried tokens shown in Figure~\ref{fig:scf_for_race}, where ping producer, ping consumer, pong producer, and pong consumer map to the four loop body subgraphs.
Two of the four tokens (annotated in gray and green), represent the producer-consumer dataflow for the ping and pong stages, while the other two tokens (red and blue) capture intra-stage resource contention on the producer and consumer side, respectively.
The final ACDG representing the two-stage pipeline is illustrated in Figure~\ref{fig:pipeline_acdg_flattened}, where the flattened form highlights how each token enforces correctness across iterations.

To demonstrate the pipelining transformation process, we implemented a simple case study in which a data stream traverses through an AMD NPU memory tile, featuring multiple memory banks and data ports.
Figure~\ref{fig:pingpong_compare} shows that with ping-pong enabled, the \tech compiler correctly identifies producer (write) and consumer (read) threads from the input loops and infers an overlapping schedule.
The post-transformation runtime trace, shown in Figure~\ref{fig:pingpong_enabled}, confirms the expected behavior: data reads and writes execute concurrently across two buffers, validating the correctness and effectiveness of the pipelined ACDG transformation.



\begin{figure}[h]
  \centering
  \begin{subfigure}[h]{\textwidth}
  \hspace{-5mm}
  \begin{subfigure}[h]{0.2\textwidth}
    \centering
    \begin{tikzpicture}[thick, node distance=2.8cm]
    \newcommand{\inputlen}{3mm}
    \newcommand{\outputlen}{3mm}
    \tikzstyle {arrow} = [->, >=stealth]
    \tikzstyle {buffer} = [minimum width=7mm, minimum height=6mm, align=center, draw]
    \tikzstyle{TextBox} = [rectangle, text centered, minimum height=1em, minimum width=10em, text width=9em, font=\tiny]

    \tikzstyle{APattern} = [pattern=north west lines, pattern color=black!25!green]
    \tikzstyle{BPattern} = [pattern=horizontal lines, pattern color=blue]

    \node (m1) [buffer, APattern] at (0,0) {};
    
    \node (pt1) at ($(m1.south)+(0.0, -0.3)$) {};
    \draw ($(pt1.center)+(0.0,-0.4)$) -- node [midway, right, font=\tiny, scale=0.8] {WRITE\_PORT\_0} (pt1.center);
    \draw [arrow] (pt1.center) -- (m1.south);
    \node (pt2) at ($(m1.north)+(0.0, 0.3)$) {};
    \draw [arrow] (pt2.center) -- node [midway, right, font=\tiny, scale=0.8] {READ\_PORT\_0} ($(pt2.center)+(0.0,0.4)$);
    \draw (m1.north) -- (pt2.center);

    \draw [dashed] ($(m1.north west)+(-0.2,0.3)$) -| ($(m1.south east)+(0.2,-0.3)$) -| ($(m1.north west)+(-0.2,0.3)$);
    \node [TextBox, right, anchor=center, rotate=90] at ($(m1.west)+(-0.4,0.0)$) {Mem. tile};

\end{tikzpicture}
  \end{subfigure}
  \begin{subfigure}[h]{0.79\textwidth}
    \centering
    \begin{tikzpicture}[thick,xscale=0.9, yscale=0.5, every node/.style={scale=0.7}]

\tikzstyle{APattern} = [pattern=north west lines, pattern color=black!25!green]
\tikzstyle{BPattern} = [pattern=horizontal lines, pattern color=blue]

\definecolor{darkgreen}{RGB}{27,94,32}
\definecolor{lightgreen}{RGB}{174,213,129}

\foreach \y in {1, 2, 3} {
    \draw[gray!30, thin] (-2.5,\y) -- (10,\y);
}

\pgfmathsetmacro{\yPORTzero}{2}
\pgfmathsetmacro{\yPORTfour}{1}
\pgfmathsetmacro{\yPORTthree}{0}

\def\barheight{0.7}

\draw[APattern] (1,\yPORTzero) rectangle (3,\yPORTzero+\barheight);
\draw[APattern] (5,\yPORTzero) rectangle (7,\yPORTzero+\barheight);
\draw[APattern] (9,\yPORTzero) rectangle (10,\yPORTzero+\barheight);

\draw[APattern] (3,\yPORTfour) rectangle (5,\yPORTfour+\barheight);
\draw[APattern] (7,\yPORTfour) rectangle (9,\yPORTfour+\barheight);


\node[anchor=east] at (0,\yPORTzero+0.3) {\small WRITE\_PORT\_0};
\node[anchor=east] at (0,\yPORTfour+0.3) {\small READ\_PORT\_0};

\end{tikzpicture}
  \end{subfigure}
  \vspace{-10mm}
    \caption{
        Block diagram and trace showing memtile ping-pong buffering disabled.
    }
    \label{fig:pingpong_disabled}
  \end{subfigure}
  \begin{subfigure}[h]{\textwidth}
  \hspace{-5mm}
  \begin{subfigure}[h]{0.2\textwidth}
    \centering
    \begin{tikzpicture}[thick, node distance=2.8cm]
    \newcommand{\inputlen}{3mm}
    \newcommand{\outputlen}{3mm}
    \tikzstyle {arrow} = [->, >=stealth]
    \tikzstyle {buffer} = [minimum width=7mm, minimum height=6mm, align=center, draw]
    \tikzstyle{TextBox} = [rectangle, text centered, minimum height=1em, minimum width=10em, text width=9em, font=\tiny]

    \tikzstyle{APattern} = [pattern=north west lines, pattern color=black!25!green]
    \tikzstyle{BPattern} = [pattern=horizontal lines, pattern color=blue]

    \node (m1) [buffer, APattern] at (0,0) {};
    \node (m2) [buffer, BPattern] at ($(m1.center)+(0.8, 0.0)$) {};
    
    \node (pt1) at ($(m1.south)+(0.4, -0.3)$) {};
    \draw ($(pt1.center)+(0.0,-0.4)$) -- node [midway, left, font=\tiny, scale=0.8] {WRITE\_PORT\_0} (pt1.center);
    \draw [arrow] (pt1.center) -- (m1.south);
    \draw [arrow, dashed] (pt1.center) -- (m2.south);
    \node (pt2) at ($(m1.north)+(0.4, 0.3)$) {};
    \draw [arrow] (pt2.center) -- node [midway, left, font=\tiny, scale=0.8] {READ\_PORT\_0} ($(pt2.center)+(0.0,0.4)$);
    \draw [dashed] (pt2.center) -- (m1.north);
    \draw (pt2.center) -- (m2.north);

    \draw [dashed] ($(m1.north west)+(-0.2,0.3)$) -| ($(m2.south east)+(0.2,-0.3)$) -| ($(m1.north west)+(-0.2,0.3)$);
    \node [TextBox, right, anchor=center, rotate=90] at ($(m1.west)+(-0.4,0.0)$) {Mem. tile};

\end{tikzpicture}
  \end{subfigure}
  \begin{subfigure}[h]{0.79\textwidth}
    \centering
    \begin{tikzpicture}[thick,xscale=0.9, yscale=0.5, every node/.style={scale=0.7}]

\definecolor{darkgreen}{RGB}{27,94,32}
\definecolor{lightgreen}{RGB}{174,213,129}

\tikzstyle{APattern} = [pattern=north west lines, pattern color=black!25!green]
\tikzstyle{BPattern} = [pattern=horizontal lines, pattern color=blue]

\foreach \y in {1, 2, 3} {
    \draw[gray!30, thin] (-2.5,\y) -- (10,\y);
}

\pgfmathsetmacro{\yPORTzero}{2}
\pgfmathsetmacro{\yPORTfour}{1}
\pgfmathsetmacro{\yPORTthree}{0}

\def\barheight{0.7}

\draw[APattern] (1,\yPORTzero) rectangle (3,\yPORTzero+\barheight);
\draw[BPattern] (3.05,\yPORTzero) rectangle (5,\yPORTzero+\barheight);
\draw[APattern] (5.05,\yPORTzero) rectangle (7,\yPORTzero+\barheight);
\draw[BPattern] (7.05,\yPORTzero) rectangle (9,\yPORTzero+\barheight);
\draw[APattern] (9.05,\yPORTzero) rectangle (10,\yPORTzero+\barheight);

\draw[APattern] (3,\yPORTfour) rectangle (5,\yPORTfour+\barheight);
\draw[BPattern] (5.05,\yPORTfour) rectangle (7,\yPORTfour+\barheight);
\draw[APattern] (7.05,\yPORTfour) rectangle (9,\yPORTfour+\barheight);
\draw[BPattern] (9.05,\yPORTfour) rectangle (10,\yPORTfour+\barheight);


\node[anchor=east] at (0,\yPORTzero+0.3) {\small WRITE\_PORT\_0};
\node[anchor=east] at (0,\yPORTfour+0.3) {\small READ\_PORT\_0};

\end{tikzpicture}
  \end{subfigure}
  \vspace{-10mm}
    \caption{
        Block diagram and trace showing memtile ping-pong buffering enabled.
    }
    \label{fig:pingpong_enabled}
  \end{subfigure}
  \caption{A simple data streaming case study showing the effect of enabling two-stage hardware pipelining.}
  \label{fig:pingpong_compare}
\end{figure}

\subsubsection{Time-multiplexed Data Movement via \channel Merging}
\label{subsubsec:channel_fusion}

The decoupled \channel abstraction in \tech enables time-multiplexed data movement by allowing multiple dataflows to reuse shared communication resources through \channel merging.
This is particularly valuable in scenarios where data movement hardware---such as memory ports, DMA engines, or network routing resources---is limited.

\tech provides compiler passes that automatically detect opportunities for channel merging by analyzing the ACDG structure.
Merging is controlled via compiler flags that specify the memory hierarchy at which merging is applied. 
For selected hierarchies, all merging opportunities implicit in the control flow are greedily identified and lowered.

\begin{figure}[h]
  \centering
  \begin{tikzpicture}[thick, node distance=2.8cm]
    \newcommand{\inputlen}{3mm}
    \newcommand{\outputlen}{3mm}
    \tikzstyle {arrow} = [->, >=stealth]
    \tikzstyle {token} = [circle, minimum width=6mm, minimum height=6mm, align=center, draw]
    \tikzstyle {op} = [minimum width=23mm, minimum height=6mm, align=center, draw]
    \tikzstyle {body} = [chamfered rectangle, chamfered rectangle xsep=2cm, rotate=-90, draw]
    \tikzstyle{TextBox} = [rectangle, text centered, minimum height=1em, minimum width=10em, text width=9em, font=\tiny]


    \node (put_a) [op] at (0,0) {};
    \node [TextBox, anchor=center, scale=0.8] at (put_a) {put \@chan2 $a\left[f_a\left(\boldsymbol{i}\right)\right]\left[g_a\left(\boldsymbol{j}\right)\right]$};
    \node (get_a) [op] at ($(put_a.center)+(-0.05,1.2)$) {};
    \node [TextBox, anchor=center, scale=0.8] at (get_a) {get \@chan1 $a\left[f_a\left(\boldsymbol{i}\right)\right]$};

    \draw ($(put_a.center)+(-1.4,0.6)$) |- ($(put_a.center)+(1.25,-0.5)$) |- ($(put_a.center)+(-1.4,0.6)$);
    \node [TextBox, anchor=center] at ($(put_a.north)+(-0.6,0.1)$) {for $\boldsymbol{j}$ in $\boldsymbol{0}$ to $\boldsymbol{n}$};

    \draw ($(get_a.center)+(-1.5,0.6)$) |- ($(get_a.center)+(1.4,-2.0)$) |- ($(get_a.center)+(-1.5,0.6)$);
    \node [TextBox, anchor=center] at ($(get_a.north)+(-0.7,0.1)$) {for $\boldsymbol{i}$ in $\boldsymbol{0}$ to $\boldsymbol{m}$};
    
    \draw [arrow] (get_a.south) -- ($(get_a.south)+(0.0,-0.3)$);
    \draw [arrow] ($(get_a.center)+(-0.05,0.8)$) -- ($(get_a.center)+(-0.05,0.6)$);
    \draw [arrow] ($(get_a.center)+(-0.05,-2.0)$) -- ($(get_a.center)+(-0.05,-2.2)$);


    \node (put_b) [op] at ($(put_a.center)+(3.2,0.0)$) {};
    \node [TextBox, anchor=center, scale=0.8] at (put_b) {put \@chan4 $b\left[f_b\left(\boldsymbol{i}\right)\right]\left[g_b\left(\boldsymbol{j}\right)\right]$};
    \node (get_b) [op] at ($(put_b.center)+(-0.05,1.2)$) {};
    \node [TextBox, anchor=center, scale=0.8] at (get_b) {get \@chan3 $b\left[f_b\left(\boldsymbol{i}\right)\right]$};

    \draw ($(put_b.center)+(-1.4,0.6)$) |- ($(put_b.center)+(1.25,-0.5)$) |- ($(put_b.center)+(-1.4,0.6)$);
    \node [TextBox, anchor=center] at ($(put_b.north)+(-0.6,0.1)$) {for $\boldsymbol{j}$ in $\boldsymbol{0}$ to $\boldsymbol{n}$};

    \draw ($(get_b.center)+(-1.5,0.6)$) |- ($(get_b.center)+(1.4,-2.0)$) |- ($(get_b.center)+(-1.5,0.6)$);
    \node [TextBox, anchor=center] at ($(get_b.north)+(-0.7,0.1)$) {for $\boldsymbol{i}$ in $\boldsymbol{0}$ to $\boldsymbol{m}$};
    
    \draw [arrow] (get_b.south) -- ($(get_b.south)+(0.0,-0.3)$);
    \draw [arrow] ($(get_b.center)+(-0.05,0.8)$) -- ($(get_b.center)+(-0.05,0.6)$);
    \draw [arrow] ($(get_b.center)+(-0.05,-2.0)$) -- ($(get_b.center)+(-0.05,-2.2)$);
     
    \node (to1) [right=of put_b, xshift=-19mm, yshift=5mm] {};
    \draw [fill=black] ([xshift=-5mm, yshift=-5mm]to1.center) -- ([xshift=-5mm, yshift=5mm]to1.center) -- ([yshift=5mm]to1.center) -- ([yshift=7.5mm]to1.center) -- ([xshift=5mm]to1.center) -- ([yshift=-7.5mm]to1.center) -- ([yshift=-5mm]to1.center) -- cycle;


    \node (put_b1) [op] at ($(put_b.center)+(4.5,-0.8)$) {};
    \node [TextBox, anchor=center, scale=0.8] at (put_b1) {put \textcolor{blue}{\@chan2} $b\left[f_b\left(\boldsymbol{i}\right)\right]\left[g_b\left(\boldsymbol{j}\right)\right]$};
    \node (put_a1) [op] at ($(put_b1.center)+(0.0,0.8)$) {};
    \node [TextBox, anchor=center, scale=0.8] at (put_a1) {get \textcolor{blue}{\@chan2} $a\left[f_a\left(\boldsymbol{i}\right)\right]\left[g_a\left(\boldsymbol{j}\right)\right]$};
    \node (get_b1) [op] at ($(put_a1.center)+(-0.05,1.2)$) {};
    \node [TextBox, anchor=center, scale=0.8] at (get_b1) {get \textcolor{red}{\@chan1} $b\left[f_b\left(\boldsymbol{i}\right)\right]$};
    \node (get_a1) [op] at ($(get_b1.center)+(0.0,0.8)$) {};
    \node [TextBox, anchor=center, scale=0.8] at (get_a1) {get \textcolor{red}{\@chan1} $a\left[f_a\left(\boldsymbol{i}\right)\right]$};
    
    \draw [arrow] (get_a1.south) -- (get_b1.north);
    \draw [arrow] (put_a1.south) -- (put_b1.north);

    \draw ($(put_a1.center)+(-1.4,0.6)$) |- ($(put_a1.center)+(1.25,-1.3)$) |- ($(put_a1.center)+(-1.4,0.6)$);
    \node [TextBox, anchor=center] at ($(put_a1.north)+(-0.6,0.1)$) {for $\boldsymbol{j}$ in $\boldsymbol{0}$ to $\boldsymbol{n}$};

    \draw ($(get_a1.center)+(-1.5,0.6)$) |- ($(get_a1.center)+(1.4,-3.5)$) |- ($(get_a1.center)+(-1.5,0.6)$);
    \node [TextBox, anchor=center] at ($(get_a1.north)+(-0.7,0.1)$) {for $\boldsymbol{i}$ in $\boldsymbol{0}$ to $\boldsymbol{m}$};
    
    \draw [arrow] (get_b1.south) -- ($(get_b1.south)+(0.0,-0.3)$);
    \draw [arrow] ($(get_a1.center)+(-0.05,0.8)$) -- ($(get_a1.center)+(-0.05,0.6)$);
    \draw [arrow] ($(get_a1.center)+(-0.05,-3.5)$) -- ($(get_a1.center)+(-0.05,-3.7)$);
	
    \captionsetup[sub]{font=Large}
    \node (cap1) [text width=15em] at ([xshift=14mm, yshift=-20mm]put_a) {\subcaption{Before \channel fusion.\label{fig:before_channel_fusion}}};
    \node (cap2) [text width=15em, xshift=-1mm] at (cap1 -| put_b1) {\subcaption{After \channel fusion.\label{fig:after_channel_fusion}}};

\end{tikzpicture}
  \caption{Visualizations of ACDGs before and after channel merging.}
  \label{fig:chan_fusion}
\end{figure}
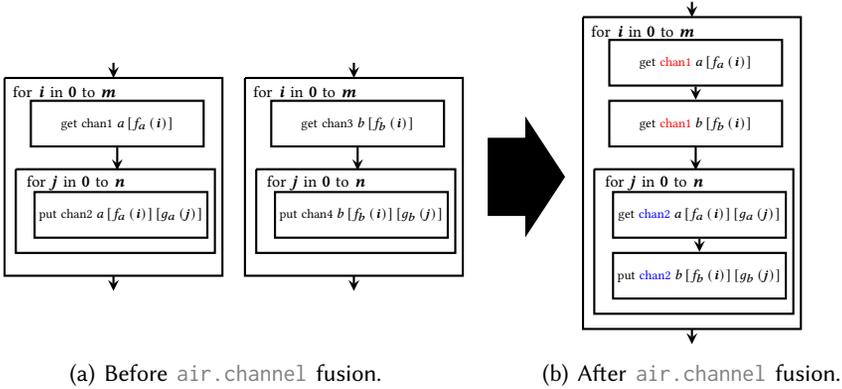

\begin{figure}[h]
  \centering
  \begin{minipage}[c]{\textwidth}
  \begin{subfigure}[t]{0.48\textwidth}
    \centering
    \begin{tikzpicture}[thick, node distance=2.8cm]
    \newcommand{\inputlen}{3mm}
    \newcommand{\outputlen}{3mm}
    \tikzstyle {arrow} = [->, >=stealth]
    \tikzstyle {buffer} = [minimum width=10mm, minimum height=8mm, align=center, draw]
    \tikzstyle{TextBox} = [rectangle, text centered, minimum height=1em, minimum width=10em, text width=9em]

    \tikzstyle{APattern} = [pattern=north west lines, pattern color=black!25!green]
    \tikzstyle{BPattern} = [pattern=horizontal lines, pattern color=blue]
    \tikzstyle {dot} = [circle, fill=black, inner sep=0.3mm, draw]


    \node (c1) [buffer, APattern] at (0,0) {};
    \node (c2) [buffer, BPattern] at ($(c1.center)+(2.0, 0)$) {};
    \node (m1) [buffer, APattern] at ($(c1.center)+(0, -2.0)$) {};
    \node [dot] at ($(m1.north)+(0.0,0.2)$){};
    \node [dot] at ($(m1.south)+(0.0,-0.2)$){};
    \draw [arrow] ($(m1.north)+(-0.0,0)$) -- node [midway, right, font=\tiny, yshift=-1mm] {READ\_PORT\_1} ($(c1.south)+(-0.0,0)$);
    \node (s1) at ($(m1.center)+(0, -1.3)$) {};
    \node [TextBox, anchor=north, scale=0.8] at (s1) {Shim port 1};
    \draw [arrow] (s1.center) -- node [midway, right, font=\tiny] {WRITE\_PORT\_1} (m1.south);
    \node (m2) [buffer, BPattern] at ($(m1.center)+(2.0, 0.0)$) {};
    \node [dot] at ($(m2.north)+(0.0,0.2)$){};
    \node [dot] at ($(m2.south)+(0.0,-0.2)$){};
    \draw [arrow] ($(m2.north)+(-0.0,0)$) -- node [midway, right, font=\tiny, yshift=-1mm] {READ\_PORT\_2} ($(c2.south)+(-0.0,0)$);
    \node (s2) at ($(m2.center)+(0, -1.3)$) {};
    \node [TextBox, anchor=north, scale=0.8] at (s2) {Shim port 2};
    \draw [arrow] (s2.center) -- node [midway, right, font=\tiny] {WRITE\_PORT\_2} (m2.south);

    \draw [dashed] ($(c1.north west)+(-0.4,0.2)$) -- ($(c2.north east)+(0.4,0.2)$);
    \node [TextBox, right, anchor=center, rotate=90] at ($(c1.west)+(-1.2,0.0)$) {Core tiles};
    \draw [dashed] ($(c1.south west)+(-0.4,-0.2)$) -- ($(c2.south east)+(0.4,-0.2)$);
    
    \draw [dashed] ($(m1.north west)+(-0.4,0.2)$) -- ($(m2.north east)+(0.4,0.2)$);
    \node [TextBox, right, anchor=center, rotate=90] at ($(m1.west)+(-1.2,0.0)$) {Mem. tiles};
    \draw [dashed] ($(m1.south west)+(-0.4,-0.2)$) -- ($(m2.south east)+(0.4,-0.2)$);
    
\end{tikzpicture}
    \caption{Before \channel merging.}
    \label{fig:chan_fusion_block_before}
  \end{subfigure}
  \hfill
  \begin{subfigure}[t]{0.48\textwidth}
    \centering
    \begin{tikzpicture}[thick, node distance=2.8cm]
    \newcommand{\inputlen}{3mm}
    \newcommand{\outputlen}{3mm}
    \tikzstyle {arrow} = [->, >=stealth]
    \tikzstyle {buffer} = [minimum width=11mm, minimum height=8mm, align=center, draw]
    \tikzstyle{TextBox} = [rectangle, text centered, minimum height=1em, minimum width=10em, text width=9em]

    \tikzstyle{APattern} = [pattern=north west lines, pattern color=black!25!green]
    \tikzstyle{BPattern} = [pattern=horizontal lines, pattern color=blue]
    \tikzstyle {dot} = [circle, fill=black, inner sep=0.3mm, draw]


    \node (c1) [buffer, APattern] at (0,0) {};
    \node (c2) [buffer, BPattern] at ($(c1.center)+(1.2, 0.0)$) {};
    
    \node (m1) [buffer, APattern] at ($(c1.center)+(0, -2.0)$) {};
    \node (m2) [buffer, BPattern] at ($(m1.center)+(1.2, 0.0)$) {};
    \node (pt1) [dot] at ($(m1.south)+(0.6, -0.3)$) {};
    \node (s1) at ($(m1.center)+(0.6,-1.3)$) {};
    \node [TextBox, anchor=north, scale=0.8] at (s1) {Shim port 1};
    \draw (s1.center) -- node [midway, left, font=\tiny] {WRITE\_PORT\_0} (pt1.center);
    \draw [arrow] (pt1.center) -- (m1.south);
    \draw [arrow, dashed] (pt1.center) -- (m2.south);
    \node (pt2) [dot] at ($(m1.north)+(0.6, 0.3)$) {};
    \draw (pt2.center) -- node [midway, left, font=\tiny] {READ\_PORT\_0} ($(pt2.center)+(0.0,0.4)$);
    \draw [dashed] (pt2.center) -- (m1.north);
    \draw (pt2.center) -- (m2.north);
    \node (pt3) at ($(c1.south)+(0.6, -0.3)$) {};
    \draw [arrow] ($(pt2.center)+(0.0,0.4)$) -- (pt3.center) -- (c1.south);
    \draw [arrow, dashed] (pt3.center) -- (c2.south);
    
    \draw [dashed] ($(m1.north west)+(-0.4,0.3)$) -| ($(m2.south east)+(0.4,-0.3)$) -| ($(m1.north west)+(-0.4,0.3)$);
    \node [TextBox, right, anchor=center, rotate=90] at ($(m1.west)+(-1.2,0.0)$) {Mem. tile};
    \draw [dashed] ($(c1.north west)+(-0.4,0.2)$) -- ($(c2.north east)+(0.4,0.2)$);
    \node [TextBox, right, anchor=center, rotate=90] at ($(c1.west)+(-1.2,0.0)$) {Core tiles};
    \draw [dashed] ($(c1.south west)+(-0.4,-0.3)$) -- ($(c2.south east)+(0.4,-0.3)$);
    
\end{tikzpicture}
    \caption{After \channel merging.}
    \label{fig:chan_fusion_block_after}
  \end{subfigure}
  \end{minipage}
  \\
  \begin{minipage}[c]{\textwidth}
  \begin{subfigure}[b]{0.48\textwidth}
    \centering
    \begin{tikzpicture}[thick,xscale=0.83, yscale=0.5, every node/.style={scale=0.7}]

\definecolor{darkgreen}{RGB}{27,94,32}
\definecolor{lightgreen}{RGB}{174,213,129}

\tikzstyle{APattern} = [pattern=north west lines, pattern color=black!25!green]
\tikzstyle{BPattern} = [pattern=horizontal lines, pattern color=blue]

\foreach \y in {0, 1, 2, 3, 4} {
    \draw[gray!30, thin] (-0.5,\y) -- (7,\y);
}

\pgfmathsetmacro{\yPORTzero}{3}
\pgfmathsetmacro{\yPORTone}{1}
\pgfmathsetmacro{\yPORTtwo}{2}
\pgfmathsetmacro{\yPORTthree}{0}

\def\barheight{0.7}

\draw[APattern] (0,\yPORTzero) rectangle (1.5,\yPORTzero+\barheight);
\draw[APattern] (3.10,\yPORTzero) rectangle (4.60,\yPORTzero+\barheight);
\draw[APattern] (6.15,\yPORTzero) rectangle (7.00,\yPORTzero+\barheight);

\draw[APattern] (1.55,\yPORTtwo) rectangle (2.30,\yPORTtwo+\barheight);
\draw[APattern] (2.35,\yPORTtwo) rectangle (3.10,\yPORTtwo+\barheight);
\draw[APattern] (4.60,\yPORTtwo) rectangle (5.35,\yPORTtwo+\barheight);
\draw[APattern] (5.40,\yPORTtwo) rectangle (6.15,\yPORTtwo+\barheight);

\draw[BPattern] (0,\yPORTone) rectangle (2,\yPORTone+\barheight);
\draw[BPattern] (4.00,\yPORTone) rectangle (6.00,\yPORTone+\barheight);

\draw[BPattern] (2.05,\yPORTthree) rectangle (2.5,\yPORTthree+\barheight);
\draw[BPattern] (2.55,\yPORTthree) rectangle (3.0,\yPORTthree+\barheight);
\draw[BPattern] (3.05,\yPORTthree) rectangle (3.5,\yPORTthree+\barheight);
\draw[BPattern] (3.55,\yPORTthree) rectangle (4.0,\yPORTthree+\barheight);
\draw[BPattern] (6.00,\yPORTthree) rectangle (6.5,\yPORTthree+\barheight);
\draw[BPattern] (6.55,\yPORTthree) rectangle (7.0,\yPORTthree+\barheight);

\node[anchor=east] at (0,\yPORTzero+0.3) {\small W0};
\node[anchor=east] at (0,\yPORTone+0.3) {\small W1};
\node[anchor=east] at (0,\yPORTtwo+0.3) {\small R0};
\node[anchor=east] at (0,\yPORTthree+0.3) {\small R1};

\draw[dashed, red] (0, 3) -- (0, 4);
\draw[dashed, red] (1.5, 3) -- (1.5, 4);
\draw[<->, red] (0, 3.8) -- node [midway, above] {$a\left[f_a\left(\boldsymbol{i}\right)\right]$} (1.5, 3.8);
\draw[dashed, black!25!green] (1.55, 2) -- (1.55, 5);
\draw[dashed, black!25!green] (2.3, 2) -- (2.3, 5);
\draw[<->, black!25!green] (1.55, 4.8) -- node [midway, above] {$a\left[f_a\left(\boldsymbol{i}\right)\right]\left[g_a\left(\boldsymbol{j}\right)\right]$} (2.3, 4.8);

\draw[dashed, blue] (0, 2) -- (0, 0);
\draw[dashed, blue] (2.0, 2) -- (2.0, 0);
\draw[<->, blue] (0, 0.2) -- node [midway, below] {$b\left[f_b\left(\boldsymbol{i}\right)\right]$} (2.0, 0.2);
\draw[dashed, black!25!gray] (2.05, 1) -- (2.05, -1);
\draw[dashed, black!25!gray] (2.5, 1) -- (2.5, -1);
\draw[<->, black!25!gray] (2.05, -0.8) -- node [midway, below] {$b\left[f_b\left(\boldsymbol{i}\right)\right]\left[g_b\left(\boldsymbol{j}\right)\right]$} (2.5, -0.8);

\end{tikzpicture}
    \caption{Pre-merge trace at memory tile.}
    \label{fig:chan_fusion_trace_disabled}
  \end{subfigure}
  \hfill
  \begin{subfigure}[b]{0.48\textwidth}
    \centering
    \begin{tikzpicture}[thick,xscale=0.83, yscale=0.5, every node/.style={scale=0.7}]

\definecolor{darkgreen}{RGB}{27,94,32}
\definecolor{lightgreen}{RGB}{174,213,129}

\tikzstyle{APattern} = [pattern=north west lines, pattern color=black!25!green]
\tikzstyle{BPattern} = [pattern=horizontal lines, pattern color=blue]

\foreach \y in {2, 3, 4} {
    \draw[gray!30, thin] (-0.5,\y) -- (7,\y);
}

\pgfmathsetmacro{\yPORTzero}{3}
\pgfmathsetmacro{\yPORTone}{2}

\def\barheight{0.7}

\draw[APattern] (0,\yPORTzero) rectangle (1.5,\yPORTzero+\barheight);
\draw[BPattern] (1.55,\yPORTzero) rectangle (3.55,\yPORTzero+\barheight);
\draw[APattern] (3.60,\yPORTzero) rectangle (5.10,\yPORTzero+\barheight);
\draw[BPattern] (5.75,\yPORTzero) rectangle (7.00,\yPORTzero+\barheight);

\draw[APattern] (1.50,\yPORTone) rectangle (2.25,\yPORTone+\barheight);
\draw[APattern] (2.30,\yPORTone) rectangle (3.05,\yPORTone+\barheight);
\draw[BPattern] (3.60,\yPORTone) rectangle (4.10,\yPORTone+\barheight);
\draw[BPattern] (4.15,\yPORTone) rectangle (4.65,\yPORTone+\barheight);
\draw[BPattern] (4.70,\yPORTone) rectangle (5.20,\yPORTone+\barheight);
\draw[BPattern] (5.25,\yPORTone) rectangle (5.75,\yPORTone+\barheight);
\draw[APattern] (5.75,\yPORTone) rectangle (6.50,\yPORTone+\barheight);
\draw[APattern] (6.55,\yPORTone) rectangle (7.00,\yPORTone+\barheight);

\node[anchor=east] at (0,\yPORTzero+0.3) {\small W0};
\node[anchor=east] at (0,\yPORTone+0.3) {\small R0};

\draw[dashed, red] (0, 3) -- (0, 4);
\draw[dashed, red] (1.5, 3) -- (1.5, 4);
\draw[<->, red] (0, 3.8) -- node [midway, above] {$a\left[f_a\left(\boldsymbol{i}\right)\right]$} (1.5, 3.8);
\draw[dashed, blue] (1.55, 3) -- (1.55, 4);
\draw[dashed, blue] (3.55, 3) -- (3.55, 4);
\draw[<->, blue] (1.55, 3.8) -- node [midway, above] {$b\left[f_b\left(\boldsymbol{i}\right)\right]$} (3.55, 3.8);

\draw[dashed, black!25!green] (1.50, 3) -- (1.50, 2);
\draw[dashed, black!25!green] (2.25, 3) -- (2.25, 2);
\draw[<->, black!25!green] (1.50, 1.8) -- node [midway, below] {$a\left[f_a\left(\boldsymbol{i}\right)\right]\left[g_a\left(\boldsymbol{j}\right)\right]$} (2.25, 1.8);
\draw[dashed, black!25!gray] (4.15, 3) -- (4.15, 2);
\draw[dashed, black!25!gray] (4.65, 3) -- (4.65, 2);
\draw[<->, black!25!gray] (4.15, 1.8) -- node [midway, below] {$b\left[f_b\left(\boldsymbol{i}\right)\right]\left[g_b\left(\boldsymbol{j}\right)\right]$} (4.65, 1.8);

\end{tikzpicture}
    \vspace{13mm}
    \caption{Post-merge trace at memory tile.}
    \label{fig:chan_fusion_trace_enabled}
  \end{subfigure}
  \end{minipage}
  \caption{
    Impact of \channel merging on data movement parallelism and resource usage.
    \subref{fig:chan_fusion_block_before} and \subref{fig:chan_fusion_block_after} show schematic diagrams before and after \channel merging.
    \subref{fig:chan_fusion_trace_disabled} and \subref{fig:chan_fusion_trace_enabled} show performance traces pre- and post-merging.
  }
  \label{fig:chan_fusion_block_and_trace}
\end{figure}
Figure~\ref{fig:chan_fusion} illustrates a generic example: in Figure~\subref{fig:chan_fusion_block_before}, two imperfect loop nests perform channel \texttt{put} and \texttt{get} operations on separate memory objects $a$ and $b$, through affine maps $f_a$, $g_a$, $f_b$, and $g_b$, respectively.
Merging is permitted when the iteration domains ${\boldsymbol{i}, \boldsymbol{j}}$ match, ensuring correctness when interleaving the data movements.

The resulting fused ACDG, shown in Figure~\subref{fig:chan_fusion_block_after}, sequentializes the data movements by interleaving the two loops, consolidating their use of the \channel operations \texttt{@chan1} and \texttt{@chan2}.

Figure~\ref{fig:chan_fusion_block_and_trace} further demonstrates the hardware mapping of the fused design onto an NPU memory tile, along with performance traces showing the data movement schedule.
Both the original and fused designs apply pipelined execution following the scheme in Section~\ref{subsubsec:pingpong}.
After merging, data movements are time-multiplexed, reducing contention for ports and buffers, thereby lowering resource utilization while preserving performance.

\subsubsection{Parallelized Data Movement via \channel Splitting}
\label{subsubsec:memref_splitting}

While channel merging enables time-multiplexed reuse of constrained DMA resources by sequentializing data transfers, such serialization may limit performance when hardware availability permits greater parallelism.
When DMA resources are abundant, \tech supports an alternative strategy: exposing and exploiting data movement parallelism through \memref splitting.

Inputs for \tech, often from high-level IRs expressed using generic tensor abstractions, do not always consider spatial memory connectivity constraints in target architectures such as AMD NPUs during bufferization, leading to degraded performance or mapping failures at implementation time.
To address this, \memref splitting performs a dataflow-aware partitioning analysis that refines buffer allocations based on the actual access patterns and hardware platform constraints.

In a common access pattern, a memory object $\boldsymbol{a}$ is read once and written to multiple outputs. Using the polyhedral representation, the read and write operations within loop nests over $\boldsymbol{i}$ and $\boldsymbol{j}$ can be represented as $\boldsymbol{a}\left[f\left(\boldsymbol{i}\right)\right]$ and $\boldsymbol{a}\left[f\left(\boldsymbol{i}\right)\right]\left[g\left(\boldsymbol{j}\right)\right]$, where $f$ and $g$ are affine maps.
A concrete example of $g$ which implies a splittable data access pattern, with $\boldsymbol{i} \in \mathbb{Z}^1$, is one with dependence polyhedron $\{ S0\left[\boldsymbol{i}\right] \to S0\left[i\ \mathrm{mod}\ 2\right] \}$, indicating two disjoint access patterns.

The affine map transformation is made possible by \tech's explicit representation of parallelism: by analyzing parallel \channel operations and the associated asynchronous dependencies, the compiler infers the implicit parallel access patterns, transforms the affine access functions to partition independent memory accesses, and bufferizes them into smaller sub-buffers---guaranteeing parallel, conflict-free access at runtime.

In the original schedule (Figure~\ref{fig:buffer_before_split}), $\boldsymbol{a}$ is naively bufferized into a single memory object, leading to sequentialized reads $\boldsymbol{a}\left[f\left(\boldsymbol{i}\right)\right]$ over time, regardless of available parallel memory tiles and DMA engines.
This limits parallelism across memory tiles and DMA engines, leading to reduced throughput and potential port over-utilization that can cause mapping failures.

After \memref splitting, \tech transforms the access maps $f \to \left<f_1, f_2\right>$, partitioning $\boldsymbol{a}\left[f\left(\boldsymbol{i}\right)\right]$ into multiple independent sub-tensors $\left<\boldsymbol{a}\left[f_1\left(\boldsymbol{i}\right)\right], \boldsymbol{a}\left[f_2\left(\boldsymbol{i}\right)\right]\right>$ that can be allocated to separate memory tiles.
This results in an optimized schedule, enabling independent and concurrent data movement across the spatial fabric.

Following this workflow, we implemented a synthetic data streaming experiment, moving data from shim ports through memory tiles to cores using a unit-stride affine access pattern.
The pre-splitting hardware trace in Figure~\ref{fig:trace_split_l2_before} shows serialization of all inbound traffic through a single tile, limiting throughput.
After \memref splitting, the post-splitting trace in Figure~\ref{fig:trace_split_l2_after} demonstrates parallel streaming through disjoint memory tiles and shim ports, significantly improving data movement efficiency.





\begin{figure}[h]
  \centering
  \begin{subfigure}[t]{0.48\textwidth}
    \centering
    \begin{tikzpicture}[thick, node distance=2.8cm]
    \newcommand{\inputlen}{3mm}
    \newcommand{\outputlen}{3mm}
    \tikzstyle {arrow} = [->, >=stealth]
    \tikzstyle {buffer} = [minimum width=10mm, minimum height=8mm, align=center, draw]
    \tikzstyle{TextBox} = [rectangle, text centered, minimum height=1em, minimum width=10em, text width=9em]

    \tikzstyle{APattern} = [pattern=north west lines, pattern color=black!25!green]
    \tikzstyle{BPattern} = [pattern=horizontal lines, pattern color=blue]
    \tikzstyle {dot} = [circle, fill=black, inner sep=0.3mm, draw]


    \node (c1) [buffer, APattern] at (0,0) {};
    \node (c2) [buffer, BPattern] at ($(c1.center)+(2.0, 0)$) {};
    \node (m1) [buffer] at ($(c1.center)+(0, -2.0)$) {};
    \node [dot] at ($(m1.north)+(-0.1,0.2)$){};
    \node [dot] at ($(m1.north)+(0.1,0.2)$){};
    \node [dot] at ($(m1.south)+(0.0,-0.2)$){};
    \node (m2) [buffer, white, fill=white] at ($(m1.center)+(2.0, 0)$) {};
    \draw[white, APattern] (m1.south west) rectangle (m1.north);
    \draw[white, BPattern] (m1.south) rectangle (m1.north east);
    \draw [arrow] ($(m1.north)+(-0.1,0)$) -- node [midway, left, font=\tiny, yshift=-1mm] {READ\_PORT\_1} ($(c1.south)+(-0.1,0)$);
    \draw [arrow] ($(m1.north)+(0.1,0)$) |- ($(m1.north)+(0.1,0.3)$) |- ($(m1.north)+(0.8,0.3)$) -| node [midway, above, font=\tiny, xshift=-8mm, yshift=-0.5mm] {READ\_PORT\_2} (c2.south);
    \node (s1) at ($(m1.center)+(0, -1.5)$) {};
    \node [TextBox, anchor=north, scale=0.8] at (s1) {Shim port 1};
    \draw [arrow] (s1.center) -- node [midway, right, font=\tiny, yshift=1mm] {WRITE\_PORT\_1} (m1.south);


    \draw [dashed] ($(c1.north west)+(-0.4,0.2)$) -- ($(c2.north east)+(0.4,0.2)$);
    \node [TextBox, right, anchor=center, rotate=90] at ($(c1.west)+(-1.2,0.0)$) {Core tiles};
    \draw [dashed] ($(c1.south west)+(-0.4,-0.2)$) -- ($(c2.south east)+(0.4,-0.2)$);
    
    \draw [dashed] ($(m1.north west)+(-0.4,0.2)$) -- ($(m2.north east)+(0.4,0.2)$);
    \node [TextBox, right, anchor=center, rotate=90] at ($(m1.west)+(-1.2,0.0)$) {Mem. tiles};
    \draw [dashed] ($(m1.south west)+(-0.4,-0.2)$) -- ($(m2.south east)+(0.4,-0.2)$);


\end{tikzpicture}
    \caption{Before \memref splitting.}
    \label{fig:buffer_before_split}
  \end{subfigure}
  \hfill
  \begin{subfigure}[t]{0.48\textwidth}
    \centering
    \begin{tikzpicture}[thick, node distance=2.8cm]
    \newcommand{\inputlen}{3mm}
    \newcommand{\outputlen}{3mm}
    \tikzstyle {arrow} = [->, >=stealth]
    \tikzstyle {buffer} = [minimum width=10mm, minimum height=8mm, align=center, draw]
    \tikzstyle{TextBox} = [rectangle, text centered, minimum height=1em, minimum width=10em, text width=9em]

    \tikzstyle{APattern} = [pattern=north west lines, pattern color=black!25!green]
    \tikzstyle{BPattern} = [pattern=horizontal lines, pattern color=blue]
    \tikzstyle {dot} = [circle, fill=black, inner sep=0.3mm, draw]


    \node (c1) [buffer, APattern] at (0,0) {};
    \node (c2) [buffer, BPattern] at ($(c1.center)+(2.0, 0)$) {};
    \node (m1) [buffer, APattern] at ($(c1.center)+(0, -2.0)$) {};
    \node [dot] at ($(m1.north)+(0.0,0.2)$){};
    \node [dot] at ($(m1.south)+(0.0,-0.2)$){};
    \draw [arrow] ($(m1.north)+(-0.0,0)$) -- node [midway, left, font=\tiny, yshift=-1mm] {READ\_PORT\_1} ($(c1.south)+(-0.0,0)$);
    \node (s1) at ($(m1.center)+(0, -1.5)$) {};
    \node [TextBox, anchor=north, scale=0.8] at (s1) {Shim port 1};
    \draw [arrow] (s1.center) -- node [midway, left, font=\tiny, yshift=1mm] {WRITE\_PORT\_1} (m1.south);
    \node (m2) [buffer, BPattern] at ($(m1.center)+(2.0, 0.0)$) {};
    \node [dot] at ($(m2.north)+(0.0,0.2)$){};
    \node [dot] at ($(m2.south)+(0.0,-0.2)$){};
    \draw [arrow] ($(m2.north)+(-0.0,0)$) -- node [midway, left, font=\tiny, yshift=-1mm] {READ\_PORT\_2} ($(c2.south)+(-0.0,0)$);
    \node (s2) at ($(m2.center)+(0, -1.5)$) {};
    \node [TextBox, anchor=north, scale=0.8] at (s2) {Shim port 2};
    \draw [arrow] (s2.center) -- node [midway, left, font=\tiny, yshift=1mm] {WRITE\_PORT\_2} (m2.south);


    \draw [dashed] ($(c1.north west)+(-0.4,0.2)$) -- ($(c2.north east)+(0.4,0.2)$);
    \node [TextBox, right, anchor=center, rotate=90] at ($(c1.west)+(-1.2,0.0)$) {Core tiles};
    \draw [dashed] ($(c1.south west)+(-0.4,-0.2)$) -- ($(c2.south east)+(0.4,-0.2)$);
    
    \draw [dashed] ($(m1.north west)+(-0.4,0.2)$) -- ($(m2.north east)+(0.4,0.2)$);
    \node [TextBox, right, anchor=center, rotate=90] at ($(m1.west)+(-1.2,0.0)$) {Mem. tiles};
    \draw [dashed] ($(m1.south west)+(-0.4,-0.2)$) -- ($(m2.south east)+(0.4,-0.2)$);

    
\end{tikzpicture}
    \caption{After \memref splitting.}
    \label{fig:buffer_after_split}
  \end{subfigure}
  \begin{subfigure}[t]{0.48\textwidth}
    \centering
    \begin{tikzpicture}[thick,xscale=0.83, yscale=0.5, every node/.style={scale=0.7}]

\definecolor{darkgreen}{RGB}{27,94,32}
\definecolor{lightgreen}{RGB}{174,213,129}

\tikzstyle{APattern} = [pattern=north west lines, pattern color=black!25!green]
\tikzstyle{BPattern} = [pattern=horizontal lines, pattern color=blue]

\foreach \y in {1, 2, 3, 4} {
    \draw[gray!30, thin] (-0.5,\y) -- (7,\y);
}

\pgfmathsetmacro{\yPORTzero}{3}
\pgfmathsetmacro{\yPORTone}{2}
\pgfmathsetmacro{\yPORTtwo}{1}

\def\barheight{0.7}

\draw[white, APattern] (0,\yPORTzero) rectangle (1,\yPORTzero+\barheight);
\draw[white, BPattern] (1,\yPORTzero) rectangle (2,\yPORTzero+\barheight);
\draw[fill opacity=0] (0,\yPORTzero) rectangle (2,\yPORTzero+\barheight);
\draw[white, APattern] (3,\yPORTzero) rectangle (4,\yPORTzero+\barheight);
\draw[white, BPattern] (4,\yPORTzero) rectangle (5,\yPORTzero+\barheight);
\draw[fill opacity=0] (3,\yPORTzero) rectangle (5,\yPORTzero+\barheight);
\draw[white, APattern] (6,\yPORTzero) rectangle (7,\yPORTzero+\barheight);
\draw[fill opacity=0] (6,\yPORTzero) rectangle (7,\yPORTzero+\barheight);

\draw[APattern] (2,\yPORTone) rectangle (3,\yPORTone+\barheight);
\draw[APattern] (5,\yPORTone) rectangle (6,\yPORTone+\barheight);

\draw[BPattern] (2,\yPORTtwo) rectangle (3,\yPORTtwo+\barheight);
\draw[BPattern] (5,\yPORTtwo) rectangle (6,\yPORTtwo+\barheight);

\node[anchor=east] at (0,\yPORTzero+0.3) {\small W0};
\node[anchor=east] at (0,\yPORTone+0.3) {\small R0};
\node[anchor=east] at (0,\yPORTtwo+0.3) {\small R1};

\draw[dashed, red] (0, 3) -- (0, 4);
\draw[dashed, red] (2, 3) -- (2, 4);
\draw[<->, red] (0, 3.8) -- node [midway, above] {$t\left[f\left(\boldsymbol{i}\right)\right]$} (2, 3.8);
\draw[dashed, black!25!green] (2, 2) -- (2, 5);
\draw[dashed, black!25!green] (3, 2) -- (3, 5);
\draw[<->, black!25!green] (2, 4.8) -- node [midway, above] {$t\left[f\left(\boldsymbol{i}\right)\right]\left[g_1\left(\boldsymbol{j}\right)\right]$} (3, 4.8);

\draw[dashed, blue] (2, 2) -- (2, 0);
\draw[dashed, blue] (3, 2) -- (3, 0);
\draw[<->, blue] (2, 0.2) -- node [midway, below] {$t\left[f\left(\boldsymbol{i}\right)\right]\left[g_2\left(\boldsymbol{j}\right)\right]$} (3, 0.2);

\end{tikzpicture}
    \caption{Mem. tile trace, pre-splitting.}
    \label{fig:trace_split_l2_before}
  \end{subfigure}
  \hfill
  \begin{subfigure}[t]{0.48\textwidth}
    \centering
    \begin{tikzpicture}[thick,xscale=0.83, yscale=0.5, every node/.style={scale=0.7}]

\definecolor{darkgreen}{RGB}{27,94,32}
\definecolor{lightgreen}{RGB}{174,213,129}

\tikzstyle{APattern} = [pattern=north west lines, pattern color=black!25!green]
\tikzstyle{BPattern} = [pattern=horizontal lines, pattern color=blue]

\foreach \y in {0, 1, 2, 3, 4} {
    \draw[gray!30, thin] (-0.5,\y) -- (7,\y);
}

\pgfmathsetmacro{\yPORTzero}{3}
\pgfmathsetmacro{\yPORTone}{2}
\pgfmathsetmacro{\yPORTtwo}{1}
\pgfmathsetmacro{\yPORTthree}{0}

\def\barheight{0.7}

\draw[APattern] (0,\yPORTzero) rectangle (1,\yPORTzero+\barheight);
\draw[APattern] (2,\yPORTzero) rectangle (3,\yPORTzero+\barheight);
\draw[APattern] (4,\yPORTzero) rectangle (5,\yPORTzero+\barheight);
\draw[APattern] (6,\yPORTzero) rectangle (7,\yPORTzero+\barheight);

\draw[APattern] (1,\yPORTone) rectangle (2,\yPORTone+\barheight);
\draw[APattern] (3,\yPORTone) rectangle (4,\yPORTone+\barheight);
\draw[APattern] (5,\yPORTone) rectangle (6,\yPORTone+\barheight);

\draw[BPattern] (0,\yPORTtwo) rectangle (1,\yPORTtwo+\barheight);
\draw[BPattern] (2,\yPORTtwo) rectangle (3,\yPORTtwo+\barheight);
\draw[BPattern] (4,\yPORTtwo) rectangle (5,\yPORTtwo+\barheight);
\draw[BPattern] (6,\yPORTtwo) rectangle (7,\yPORTtwo+\barheight);

\draw[BPattern] (1,\yPORTthree) rectangle (2,\yPORTthree+\barheight);
\draw[BPattern] (3,\yPORTthree) rectangle (4,\yPORTthree+\barheight);
\draw[BPattern] (5,\yPORTthree) rectangle (6,\yPORTthree+\barheight);

\node[anchor=east] at (0,\yPORTzero+0.3) {\small W0};
\node[anchor=east] at (0,\yPORTone+0.3) {\small R0};
\node[anchor=east] at (0,\yPORTtwo+0.3) {\small W1};
\node[anchor=east] at (0,\yPORTthree+0.3) {\small R1};

\draw[dashed, red] (0, 3) -- (0, 4);
\draw[dashed, red] (1, 3) -- (1, 4);
\draw[<->, red] (0, 3.8) -- node [midway, above] {$t\left[f_1\left(\boldsymbol{i}\right)\right]$} (1, 3.8);
\draw[dashed, black!25!green] (1, 2) -- (1, 5);
\draw[dashed, black!25!green] (2, 2) -- (2, 5);
\draw[<->, black!25!green] (1, 4.8) -- node [midway, above] {$t\left[f_1\left(\boldsymbol{i}\right)\right]\left[g_1\left(\boldsymbol{j}\right)\right]$} (2, 4.8);

\draw[dashed, blue] (0, 0) -- (0, 2);
\draw[dashed, blue] (1, 0) -- (1, 2);
\draw[<->, blue] (0, 0.1) -- node [midway, below] {$t\left[f_2\left(\boldsymbol{i}\right)\right]$} (1, 0.1);

\draw[dashed, gray] (1, -1) -- (1, 1);
\draw[dashed, gray] (2, -1) -- (2, 1);
\draw[<->, gray] (1, -0.8) -- node [midway, below] {$t\left[f_2\left(\boldsymbol{i}\right)\right]\left[g_2\left(\boldsymbol{j}\right)\right]$} (2, -0.8);

\end{tikzpicture}
    \caption{Mem. tile trace, post-splitting.}
    \label{fig:trace_split_l2_after}
  \end{subfigure}
  \caption{
    Impact of \memref splitting on data movement parallelism.
    \subref{fig:buffer_before_split} and \subref{fig:buffer_after_split} show schematic diagrams before and after \memref splitting.
    \subref{fig:trace_split_l2_before} and \subref{fig:trace_split_l2_after} show performance traces visualizing serialization and parallelism pre- and post-splitting.
  }
  \label{fig:buffer_split_comparison}
\end{figure}

\subsection{Lowering to AMD NPU Targets}
\label{subsec:lowering_to_targets}
With parallelism, data reuse, and communication patterns fully expressed, the \air IR is ready to be lowered into hardware-specific representations.
First, constructs in \tech are lowered to constructs in MLIR-AIE~\cite{MLIR_AIE}. \herd operations are lowered to per-core compute kernels.
\channel constructs are lowered to DMA engines, BDs, and stream connections between tiles.
Synchronization via \token values is implemented using tile-local locks. 

\tech supports code generation targeting multiple open-source frameworks, including LLVM IR, AMD XRT, and ROCr, enabling integration into heterogeneous systems involving CPUs and GPUs.
Synchronization between the hardware-specific IR and the runtime program is provided by tokens generated from the NPU hardware controller, which is lowered from \token values synchronizing host operations with on-device operations.

\section{Integration with AI Model Software Ecosystems}
\label{sec:ai_eco}

\tech is designed to bridge the gap between high-level AI model frameworks and low-level hardware execution platforms. 
A key feature of \tech is its flexible frontend integration, which allows it to ingest AI model specifications from multiple widely-used programming environments and IRs.
These integrations allow developers to compile high-level AI models directly into \tech's asynchronous, tiled execution model, ready for targeting spatial accelerators like GPUs and AMD NPUs.

\head{Python Integration via \air's Python Bindings}
\label{subsec:python_bindings}
\tech includes native Python bindings that expose \air dialect operations to Python-based workflows.
An example vector-add design using these bindings is shown in Appendix~\ref{appendix:vecadd_pybind}.
These bindings allow direct programmatic construction of \air IR, enabling rapid prototyping and integration with AI model preprocessing, autotuning, or interactive toolchains. 

\head{PyTorch Frontend via Torch-MLIR}
\label{subsec:pytorch}
Through Torch-MLIR, PyTorch models are lowered into MLIR dialects which are compatible with \tech's tiling, scheduling, and asynchronous lowering passes. 
This allows \tech to serve as a backend for PyTorch with no model rewriting, producing spatially executable kernels and runtime binaries for NPUs.

\head{IREE Integration for Portable Deployment}
\label{subsec:iree}
\tech interoperates with IREE by consuming its tiled intermediate MLIR representations, and producing scheduled AIR programs~\cite{IREE}. 
These can be integrated with IREE's hardware abstraction layer (HAL), enabling deployment across heterogeneous systems where AIR-based NPUs coexist with CPU and GPU targets under a unified runtime.

\head{Triton Frontend via Triton-Shared}
\label{subsec:triton}
AIR supports Triton through the Triton-Shared project~\cite{TRITON_SHARED}, which lowers Triton IR into MLIR dialects consumable by \air. 
\air's compilation pipeline then transforms these into hardware schedules and spatial mappings targeting AMD NPUs.
This enables the reuse of GPU-oriented high-level abstractions for mapping onto NPUs.
As of the date of publication,\tech is the only compiler infrastructure that enables Triton programs to target AMD NPUs, allowing the reuse of GPU-oriented high-level abstractions for spatial architectures.
The Triton-to-\air workflow is experimental and remains under active development.

\section{Design Experience and Results}
\label{sec:results}

We evaluate \tech across progressively complex AI workloads to assess its abstraction efficiency, expressiveness, and performance portability. Our analysis focuses on three main dimensions: (1) 
programming abstraction analysis of \tech using Halstead metrics~\cite{halstead_metrics}, (2) performance scaling across multiple backends and hardware configurations for matrix multiplication, and (3)
\tech's ability to express and optimize fused kernels through a case study on the LLaMA 2 MHA block. These evaluations highlight \tech’s ability to serve as a spatial compiler abstraction that balances expressiveness and analyzability, positioning it between high-level programming models such as Triton and low-level spatial backends like MLIR-AIE.

\subsection{Programming Abstraction Analysis}
    
\begin{table*}
    \centering
    \caption{
        Difference in Halstead vocabulary, difficulty, and effort among Triton, ARIES, \tech and MLIR-AIE, implementing the same set of common AI components to target AMD NPU (lower is better).
        All designs use \texttt{bfloat16} data format.
        Green shade annotates the lowest value.
        Designs implemented with $\mu$kernel called externally are annotated with \yup.
        While MLIR-AIE naturally shows higher complexity due to its explicit low-level programming model, \tech bridges the gap between Triton and MLIR-AIE, offering lower effort and difficulty while maintaining spatial expressiveness.
        }
    \begin{threeparttable}
        \begin{tabular}{lccS[table-format=2.2]S[table-format=2.3]S[table-format=2.2]S[table-format=2.2]S[table-format=5.2]S[table-format=2.2]}
            \toprule
            \multirow{2}[1]{*}{Design} & \multirow{2}[1]{*}{Abstraction} & \multirow{2}[1]{*}{\makecell{External\\$\mu$kernel}} & \multicolumn{2}{c}{Vocabulary} & \multicolumn{2}{c}{Difficulty} & \multicolumn{2}{c}{Effort} \\
            \cmidrule(lr){4-5} \cmidrule(lr){6-7} \cmidrule(lr){8-9}
             & & & {Value} & {$\times$} & {Value} & {$\times$} & {Value} & {$\times$} \\
            \midrule
            {\multirow{4}[1]{*}{\makecell[l]{\texttt{matrix\_scalar\_add}\\(single core)}}} & Triton & \nope & {\cellcolor{green!25!white}}  10 & {--} & {\cellcolor{green!25!white}}  1.25 & {--} & {\cellcolor{green!25!white}}  62.29 & {--} \\
             & ARIES & {N/A} & {N/A} & {--} & {N/A} & {--} & {N/A} & {--} \\
             & \textbf{\tech} & \nope & 14 & 1.40 & 1.64 & 1.31 & 112.14 & 1.80 \\
             & MLIR-AIE & \nope & 13 & 1.30 & 1.5 & 1.20 & 83.26 & 1.34 \\
            \midrule
            {\multirow{4}[1]{*}{\makecell{\texttt{eltwise\_binaryop}}}} & Triton & \nope & 12 & {--} &{\cellcolor{green!25!white}}  1.4 & {--} & 105.40 & {--} \\
             & ARIES & {N/A} & {N/A} & {--} & {N/A} & {--} & {N/A} & {--} \\
             & \textbf{\tech} & \yup & {\cellcolor{green!25!white}}  11 & 0.92 & 
 1.50 & 1.07 & {\cellcolor{green!25!white}}  62.27 & 0.38 \\ 
             & MLIR-AIE & \yup & 23 & 1.92 & 3.579 & 2.56 & 825.67 & 7.83 \\
            \midrule
            {\multirow{4}[1]{*}{\makecell{\texttt{softmax}}}} & Triton & \nope & 14 & {--} & 2.4 & {--} & 164.48 & {--} \\
             & ARIES & {N/A} & {N/A} & {--} & {N/A} & {--} & {N/A} & {--} \\
             & \textbf{\tech} & \yup & {\cellcolor{green!25!white}}  11 & 0.79 & {\cellcolor{green!25!white}} 
 1.50 & 0.63 & {\cellcolor{green!25!white}}  62.27 & 0.38 \\
             & MLIR-AIE & \yup & 18 & 1.29 & 4.615 & 1.92 & 692.85 & 4.21 \\
            \midrule
            {\multirow{4}[1]{*}{\makecell{\texttt{conv2d}}}} & Triton & \nope & 53 & {--} & 1.74 & {--} & 867.09 & {--} \\
             & ARIES & {N/A} & {N/A} & {--} & {N/A} & {--} & {N/A} & {--} \\
             & \textbf{\tech} & \yup & {\cellcolor{green!25!white}}  11 & 0.21 & {\cellcolor{green!25!white}} 
 1.50 & 0.86 & {\cellcolor{green!25!white}}  62.27 & 0.072 \\
             & MLIR-AIE & \yup & 36 & 0.68 & 5.0 & 2.87 & 1938.72 & 2.24 \\
            \midrule
            {\multirow{4}[1]{*}{\makecell{\texttt{matmul}}}} & Triton & \nope & 86 & {--} & 5.73 & {--} & 5410.33 & {--} \\
             & ARIES & \yup & {\cellcolor{green!25!white}}  40 & 0.47 & 4.76 & 0.83 & {\cellcolor{green!25!white}}  2079.31 & 0.38 \\
             & \textbf{\tech} & \yup & 78 & 0.91 & {\cellcolor{green!25!white}} 
 3.68 & 0.64 & 4713.03 & 0.87 \\
             & MLIR-AIE & \yup & 107 & 1.24 & 13.46 & 2.35 & 32040.15 & 5.92 \\
            \bottomrule
        \end{tabular}
    \end{threeparttable}
    \label{tab:halstead_metric}
\end{table*}

\tech provides a structured, loop-based programming interface that decouples algorithm specification from hardware mapping. 
Developers express computation at a high level while relying on the compiler to perform hardware-aware transformations. This makes \tech serves as an effective bridging layer between high-level programming models, such as Triton and PyTorch, and low-level, spatially explicit representations like MLIR-AIE, which  target fine-grained hardware configurations on spatial platforms.

To evaluate the abstraction level of \tech, we perform Halstead complexity analysis across representative AI workloads, comparing against ARIES---a compiler stack that similarly bridges high-level models to spatial hardware~\cite{zhuang2025aries}.
Halstead metrics, computed in our experiments using the open-source tool \texttt{radon}, quantify software complexity based on code structure that captures vocabulary, difficulty, and effort, to provide a language-agnostic measure of clarity and maintainability~\cite{RADON}.
The Halstead metrics were evaluated across a spectrum of representative AI designs, including matrix multiplications, strided and depth-wise convolutions, nonlinear functions such as softmax and exponentiation, and trigonometric operations used in Rotary Positional Embeddings (RoPE)~\cite{ROPE}. 


Table~\ref{tab:halstead_metric} reports the Halstead vocabulary, difficulty, and effort metrics across five representative workloads---each implemented using Triton~\cite{TRITON, TRITON_SHARED}, ARIES~\cite{zhuang2025aries}, \tech (via Python bindings), and MLIR-AIE (via IRON~\cite{IRON_FCCM})---and highlights the key trends in abstraction efficiency and programming overhead.
These examples were drawn from publicly available GitHub repositories.
The workloads span a range of complexity and include both inline and externally defined $\mu$-kernels---a templatized set of compute instructions specialized to perform a task---as indicated in the `external $\mu$kernel' column. 
Triton examples include $\mu$-kernel logic inline, which inflates vocabulary and effort metrics. 
In contrast, \tech and MLIR-AIE designs often invoke external kernels, resulting in more compact in-body control logic.

Despite this discrepancy in kernel inclusion, \tech consistently maintains Halstead difficulty and effort scores within 2$\times$ of Triton across the workloads. 
This indicates that \tech offers a similarly accessible structured parallel programming abstraction as Triton. For smaller, single-core kernels like \texttt{matrix\_scalar\_add}, \tech and MLIR-AIE show near-identical complexity.
However, for complex, multi-core designs, such as in \texttt{conv2d}, and \texttt{matmul},  \tech shows a dramatic reduction in overhead, achieving over 80\% lower difficulty and effort than MLIR-AIE.  This demonstrates \tech’s strength in managing complexity as spatial parallelism increases. 

ARIES presents matrix multiplication examples on their GitHub repository, which we used for comparisons in Table~\ref{tab:halstead_metric}.
In this example, Halstead vocabulary and effort metrics are both very low---lower than Triton---due to the reduced number of operations and operands in control logic. 
However, \tech achieves a lower difficulty score, indicating AIR-based representations use simpler and more regular constructs.
This result suggests that \tech enables structured parallelism at a comparable or lower cognitive complexity than ARIES, while supporting a broader class of workloads.




These results demonstrate that \tech effectively bridges the programming gap between the high-level Triton-style control flow and the low-level, highly explicit MLIR-AIE representation. 
By combining structured, tile-aware abstractions with token-based asynchronous scheduling, \tech enables efficient spatial hardware mapping while significantly reducing the complexity developers must manage in their source code.

\subsection{Performance Scaling: Mapping Matrix Multiplication to Spatial Hardware}
\label{subsec:matmul}

To evaluate the ability of \tech to generate efficient spatial compute kernels from generic loop-based programs, we examine its performance on matrix multiplication. 
Our experiments span a range of problem sizes (256-4096 per iteration dimension) and data formats, measured on an laptop platform featuring the AMD Ryzen AI 7840 NPU~\cite{AMD_RYZEN_7_7840U, AMD_AI_ENGINE}. 
We executed all \tech and MLIR-AIE programs using the AMD XRT runtime~\cite{XRT}, which manages binary loading, data movement, and kernel dispatch.
We evaluate our \tech-generated MLIR-AIE dialect code against MLIR-AIE's published hand-optimized matrix multiplication implementation, which has been adopted by many recent research works as the state-of-the-art baseline for spatial execution on AMD NPUs~\cite{NPU_ML_TRAINING, zhuang2025aries, IRON_FCCM}.
The goal of this evaluation is to demonstrate that \tech, starting from a naively specified nested loop for matrix multiplication, can produce performant implementations through a sequence of compiler transformations. 

Listing~\ref{lst:matmul_loop_nest} shows pseudocode for tiled matrix multiplication written in a generic loop-nest style, using \texttt{for} loops for sequential execution and \texttt{for\_all} for spatial parallelism. 
Such generic representations are beneficial for portability 
by using an algorithm specification decoupled from hardware mapping, where AI frameworks can target \tech without needing to provide device-specific code, demonstrating ease of frontend integration.
\tech compiles this form via a series of compilation passes presented in Section~\ref{sec:opts}, which optimizes the bufferization, data movement scheduling and concurrency modeling with platform awareness.

\begin{minipage}{\textwidth}
    \begin{lstlisting}[language=c, caption={Pseudocode for a tiled output-stationary matrix multiplication that drives \tech}, label=lst:matmul_loop_nest, xleftmargin=0.5em, xrightmargin=1.5em, morekeywords={for_all}]
for_all (i_outer = 0; i_outer < M; i_outer+=t_i) {
  for_all (j_outer = 0; j_outer < N; j_outer+=t_j) {
    for (k_outer = 0; k_outer < K; k_outer+=t_k) {
      for_all (ii = 0; ii < t_i; ii++) {
        for_all (jj = 0; jj < t_j; jj++) {
            for (kk = 0; kk < t_k; kk++) {
              C[ii][jj] += matmul(A[ii][kk], B[kk][jj]);
        }}}}}}
    \end{lstlisting}
\end{minipage}

The loop nest structure shown above implements an output-stationary schedule: each compute tile accumulates a portion of the output matrix locally across multiple input tile iterations (k loop). 
This is a naive but widely applicable strategy, offering high reuse of output accumulators, low communication cost for partial sums, and simple mapping to spatial arrays.
However, it is only one of many possible schedules supported by \tech, as \tech performs schedule optimizations on generic control-flow constructs, allowing for adaptability towards different compute problems and platform constraints.
        
\begin{figure*}
    \centering
    \begin{tikzpicture}
    
    \begin{groupplot}[
        scale only axis,
		width=0.28\textwidth,
		height=0.28\textwidth,
		group style={group size=3 by 1, ylabels at=edge left, yticklabels at=edge left, horizontal sep=1em},
        ylabel near ticks,
        ylabel={Throughput (GOPS)},
        xtick scale label code/.code={},
        xmin=0.1,
        xmax=140,
        ymin=100,
        ymax=10000,
        xmode=log,
        ymode=log
	]
        
        \nextgroupplot
        
        \addplot [thick, only marks, mark=x, mark options={scale=1.0, color=red}, opacity=0.8] table [y=K256TP, x=K256OPS] {data/bf16_2x2.txt}; \label{plt:matmul_bf16_2x2_k256}
        \addplot [thick, only marks, mark=+, mark options={scale=1.0, color=blue}, opacity=0.8] table [y=K1024TP, x=K1024OPS] {data/bf16_2x2.txt}; \label{plt:matmul_bf16_2x2_k1024}
        \addplot [thick, only marks, mark=o, mark options={scale=1.0, color=black!25!green}, opacity=0.8] table [y=K4096TP, x=K4096OPS] {data/bf16_2x2.txt}; \label{plt:matmul_bf16_2x2_k4096}
        \draw [thick, dashed] (axis cs:0.1,1024) -- (axis cs:140,1024);
        \draw [thick, ->, green!25!black] (axis cs:17,493.589) -- node[left, font=\tiny]{max. eff.=\%48.2} (axis cs:17,1024);
        \addplot [mark=triangle,
            mark size=2pt,
            mark options={
                draw=black,
                fill=white,
            },
            nodes near coords={1TOPS},
            every node near coord/.style={text=black, anchor=south east, font=\tiny},
            only marks,
            every mark/.append style={xshift=2, rotate=90},
        ]
        table {%
        140 1024
        };
                
        \node [text width=10em, anchor=north] at (axis description cs:0.5, 1) {\subcaption{$2\times 2$ herd\label{plot:matmul_bf16_2x2}}};
            
        \nextgroupplot [
            x label style={at={(axis description cs:0.5,-0.1)},anchor=north},
            xlabel={Compute workload (GOPs)}
        ]
        
        \addplot [thick, only marks, mark=x, mark options={scale=1.0, color=red}, opacity=0.8] table [y=K256TP, x=K256OPS] {data/bf16_2x4.txt}; \label{plt:matmul_bf16_2x4_k256}
        \addplot [thick, only marks, mark=+, mark options={scale=1.0, color=blue}, opacity=0.8] table [y=K1024TP, x=K1024OPS] {data/bf16_2x4.txt}; \label{plt:matmul_bf16_2x4_k1024}
        \addplot [thick, only marks, mark=o, mark options={scale=1.0, color=black!25!green}, opacity=0.8] table [y=K4096TP, x=K4096OPS] {data/bf16_2x4.txt}; \label{plt:matmul_bf16_2x4_k4096}
        \draw [thick, dashed] (axis cs:0.1,2048) -- (axis cs:140,2048);
        \draw [thick, ->, green!25!black] (axis cs:34,1331.30) -- node[left, font=\tiny]{max. eff.=\%65.0} (axis cs:34,2048); 
        \addplot [mark=triangle,
            mark size=2pt,
            mark options={
                draw=black,
                fill=white,
            },
            nodes near coords={2TOPS},
            every node near coord/.style={text=black, anchor=south east, font=\tiny},
            only marks,
            every mark/.append style={xshift=2, rotate=90},
        ]
        table {%
        140 2048
        };
                
        \node [text width=10em, anchor=north] at (axis description cs:0.5, 1) {\subcaption{$2\times 4$ herd\label{plot:matmul_bf16_2x4}}};
            
        \nextgroupplot
        
        \addplot [thick, only marks, mark=x, mark options={scale=1.0, color=red}, opacity=0.8] table [y=K256TP, x=K256OPS] {data/bf16_4x4.txt}; \label{plt:matmul_bf16_4x4_k256}
        \addplot [thick, only marks, mark=+, mark options={scale=1.0, color=blue}, opacity=0.8] table [y=K1024TP, x=K1024OPS] {data/bf16_4x4.txt}; \label{plt:matmul_bf16_4x4_k1024}
        \addplot [thick, only marks, mark=o, mark options={scale=1.0, color=black!25!green}, opacity=0.8] table [y=K4096TP, x=K4096OPS] {data/bf16_4x4.txt}; 
        \label{plt:matmul_bf16_4x4_k4096}

        \draw [thick, dashed] (axis cs:0.1,4096) -- (axis cs:140,4096);
        \draw [thick, ->, green!25!black] (axis cs:34,1991.19) -- node[left, font=\tiny]{max. eff.=\%48.6} (axis cs:34,4096); 
        \addplot [mark=triangle,
            mark size=2pt,
            mark options={
                draw=black,
                fill=white,
            },
            nodes near coords={4TOPS},
            every node near coord/.style={text=black, anchor=south east, font=\tiny},
            only marks,
            every mark/.append style={xshift=2, rotate=90},
        ]
        table {%
        140 4096
        };
                
        \node [text width=10em, anchor=north] at (axis description cs:0.5, 1) {\subcaption{$4\times 4$ herd\label{plot:matmul_bf16_4x4}}};
    
    \end{groupplot}

\end{tikzpicture}
	
    \caption{
        Throughput versus compute workload for \texttt{bfloat16} matrix multiplications, with shapes up to $M=N=K=4k$, for AIE tile herd sized (\subref{plot:matmul_bf16_4x4}) $2\times2$, (\subref{plot:matmul_bf16_2x4}) $2\times4$ and (\subref{plot:matmul_bf16_2x2}) $4\times4$, respectively.
        Compute workload increases along the x-axis.
        Each point represents the maximum throughput achieved across 20 random tests, to filter out any random system and DDR access latency injected at runtime.
        Each color/shape reflects a distinct $K$, with (\ref{plt:matmul_bf16_4x4_k256}), (\ref{plt:matmul_bf16_4x4_k1024}) and (\ref{plt:matmul_bf16_4x4_k4096}) annotating tests using $K=256$, $1024$ and $4096$, respectively.
        Dotted line marks the theoretical peak compute throughput\protect\footnotemark achievable for each herd of AIE cores at \texttt{bfloat16} precision, and the maximum compute efficiency achieved against it is annotated with an arrow.
    }
    \label{plot:matmul_sweep_groupplot_bf16}
\end{figure*}

\footnotetext{Theoretical peak compute throughput is calculated as the maximum compute speed achievable by the specified compute units, with the assumption of infinite data movement bandwidth and zero control overhead.}

Figure~\ref{plot:matmul_sweep_groupplot_bf16} shows the performance of \tech-compiled matrix multiplication kernels generated from a generic loop nest of the form shown in Listing~\ref{lst:matmul_loop_nest}, plotted as throughput (GOP/s) versus compute workload (GOPs). 
Three spatial hardware configurations were evaluated: $4\times4$ tile array (4 TOP/s peak) $2\times4$ tile array (2 TOP/s peak), and $2\times2$ tile array\footnote{Herd is reshaped to occupy a single column of four AIE tiles.} (1 TOP/s peak).

In this plot, the tiling sizes were fixed to $M = N = K = 64$, using the \texttt{bfloat16} (bf16) data format for both input and output buffers; this tile size is chosen to fit entirely within the local memory of a single NPU tile sized 64KB.
Note that this tile size was chosen heuristically and not fine-tuned; higher performance may be possible by adjusting tiling factors based on memory hierarchy and DMA burst sizes.

The three subplots (\subref{plot:matmul_bf16_2x2}---\subref{plot:matmul_bf16_4x4}) compare performance as the \herd dimensions increase.
The \herd dimensions are easily tunable in source code via tiling factors (Section~\ref{subsec:tiling_opt}).
The peak throughput achieved scales proportionately with the tile count, demonstrating that \tech is able to leverage increased spatial compute automatically, analyzed from the structured parallelism in the input program.

Larger problem sizes increas the computational intensity (OPs per memory access), which lead to better utilization of compute tiles in the dataflow pipeline.
As we sweep across increasing problem sizes (larger $K$ values), the throughput consistently improves.
Higher $K$ reduces the start-up effect of the dataflow pipeline by reducing the frequency of flushes as the scheduler refills accumulators with zeros, leading to increased overall performance.
This trend reflects \air's ability to schedule larger compute tiles effectively, reducing the relative overhead of data transfers and synchronization.
        
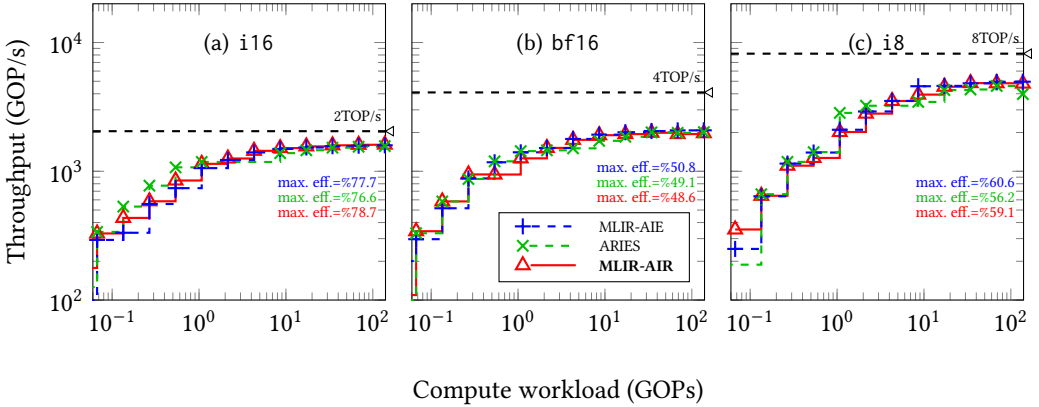
\begin{figure*}
    \centering
    \begin{tikzpicture}
    
    \begin{groupplot}[
        scale only axis,
		width=0.28\textwidth,
		height=0.28\textwidth,
		group style={group size=3 by 1, ylabels at=edge left, yticklabels at=edge left, horizontal sep=1em},
        ylabel near ticks,
        ylabel={Throughput (GOP/s)},
        xtick scale label code/.code={},
        xmin=0.06,
        xmax=140,
        ymin=100,
        ymax=20000,
        xmode=log,
        ymode=log
	]
        
        \nextgroupplot
        
        \addplot [thick, only marks, mark=triangle, mark options={scale=1.5, color=red}, opacity=1.0] table [y=TP, x=OPS] {data/i16_4x4_front.txt}; \label{plt:matmul_i16_4x4_pareto}
        \addplot [thick, color=red] table [y=TP, x=OPS, opacity=0.5] {data/i16_4x4_front_connected.txt}; 
        \addplot [thick, only marks, mark=+, mark options={scale=1.5, color=blue}, opacity=1.0] table [y=TP, x=OPS] {data/i16_4x4_iron_front.txt}; \label{plt:matmul_i16_4x4_iron_pareto}
        \addplot [thick, color=blue, dashed] table [y=TP, x=OPS, opacity=0.5] {data/i16_4x4_iron_front_connected.txt}; 
        \addplot [thick, only marks, mark=x, mark options={scale=1.5, color=black!25!green}, opacity=1.0] table [y=TP, x=OPS] {data/i16_4x4_aries_front.txt}; \label{plt:matmul_i16_4x4_aries_pareto}
        \addplot [thick, color=black!25!green, dashed] table [y=TP, x=OPS, opacity=0.5] {data/i16_4x4_aries_front_connected.txt}; 
        \draw [thick, dashed] (axis cs:0.06,2048) -- (axis cs:140,2048);
        \addplot [mark=triangle,
            mark size=2pt,
            mark options={
                draw=black,
                fill=white,
            },
            nodes near coords={2TOP/s},
            every node near coord/.style={text=black, anchor=south east, font=\tiny},
            only marks,
            every mark/.append style={xshift=2, rotate=90},
        ]
        table {%
        140 2048
        };
        \node[right, font=\tiny, color=blue] at (axis description cs:0.60,0.4) {max. eff.=\%77.7}; 
        \node[right, font=\tiny, color=black!25!green] at (axis description cs:0.60,0.35) {max. eff.=\%76.6}; 
        \node[right, font=\tiny, color=red] at (axis description cs:0.60,0.30) {max. eff.=\%78.7}; 
                
        \node [text width=10em, anchor=north] at (axis description cs:0.5, 1) {\subcaption{\texttt{i16}\label{plot:matmul_i16_4x4_pareto}}};
            
        \nextgroupplot [
            x label style={at={(axis description cs:0.5,-0.1)},anchor=north},
            xlabel={Compute workload (GOPs)}
        ]
        
        \addplot [thick, only marks, mark=triangle, mark options={scale=1.5, color=red}, opacity=1.0] table [y=TP, x=OPS] {data/bf16_4x4_front.txt}; \label{plt:matmul_bf16_4x4_pareto}
        \addplot [thick, color=red] table [y=TP, x=OPS, opacity=0.5] {data/bf16_4x4_front_connected.txt}; \label{plt:matmul_bf16_4x4_pareto_connected} 
        \addplot [thick, only marks, mark=+, mark options={scale=1.5, color=blue}, opacity=1.0] table [y=TP, x=OPS] {data/bf16_4x4_iron_front.txt}; \label{plt:matmul_bf16_4x4_iron_pareto}
        \addplot [thick, color=blue, dashed] table [y=TP, x=OPS, opacity=0.5] {data/bf16_4x4_iron_front_connected.txt}; 
        \label{plt:matmul_bf16_4x4_iron_pareto_connected}
        \addplot [thick, only marks, mark=x, mark options={scale=1.5, color=black!25!green}, opacity=1.0] table [y=TP, x=OPS] {data/bf16_4x4_aries_front.txt}; \label{plt:matmul_bf16_4x4_aries_pareto}
        \addplot [thick, color=black!25!green, dashed] table [y=TP, x=OPS, opacity=0.5] {data/bf16_4x4_aries_front_connected.txt}; 
        \label{plt:matmul_bf16_4x4_aries_pareto_connected}
        \draw [thick, dashed] (axis cs:0.06,4096) -- (axis cs:140,4096);
        \addplot [mark=triangle,
            mark size=2pt,
            mark options={
                draw=black,
                fill=white,
            },
            nodes near coords={4TOP/s},
            every node near coord/.style={text=black, anchor=south east, font=\tiny},
            only marks,
            every mark/.append style={xshift=2, rotate=90},
        ]
        table {%
        140 4096
        };
        \node[right, font=\tiny, color=blue] at (axis description cs:0.60,0.45) {max. eff.=\%50.8}; 
        \node[right, font=\tiny, color=black!25!green] at (axis description cs:0.60,0.40) {max. eff.=\%49.1}; 
        \node[right, font=\tiny, color=red] at (axis description cs:0.60,0.35) {max. eff.=\%48.6}; 
                
        \node [text width=10em, anchor=north] at (axis description cs:0.5, 1) {\subcaption{\texttt{bf16}\label{plot:matmul_bf16_4x4_pareto}}};
        
        \node[anchor=south east,draw,fill=white,inner sep=0.2em] at (axis description cs:0.95, 0.06) {\tiny
        \begin{tabular}{cl}
            \makecell{\ref{plt:matmul_bf16_4x4_iron_pareto}\ref{plt:matmul_bf16_4x4_iron_pareto_connected}} & MLIR-AIE \\
            \makecell{\ref{plt:matmul_bf16_4x4_aries_pareto}\ref{plt:matmul_bf16_4x4_aries_pareto_connected}} & ARIES \\
            \makecell{\ref{plt:matmul_bf16_4x4_pareto}\ref{plt:matmul_bf16_4x4_pareto_connected}} & \textbf{MLIR-AIR}
        \end{tabular}};
            
        \nextgroupplot
        
        \addplot [thick, only marks, mark=triangle, mark options={scale=1.5, color=red}, opacity=1.0] table [y=TP, x=OPS] {data/i8_4x4_front.txt}; \label{plt:matmul_i8_4x4_pareto}
        \addplot [thick, color=red] table [y=TP, x=OPS, opacity=0.5] {data/i8_4x4_front_connected.txt}; 
        \addplot [thick, only marks, mark=+, mark options={scale=1.5, color=blue}, opacity=1.0] table [y=TP, x=OPS] {data/i8_4x4_iron_front.txt}; \label{plt:matmul_i8_4x4_iron_pareto}
        \addplot [thick, color=blue, dashed] table [y=TP, x=OPS, opacity=0.5] {data/i8_4x4_iron_front_connected.txt}; 
        \addplot [thick, only marks, mark=x, mark options={scale=1.5, color=black!25!green}, opacity=1.0] table [y=TP, x=OPS] {data/i8_4x4_aries_front.txt}; \label{plt:matmul_i8_4x4_aries_pareto}
        \addplot [thick, color=black!25!green, dashed] table [y=TP, x=OPS, opacity=0.5] {data/i8_4x4_aries_front_connected.txt}; 
        \draw [thick, dashed] (axis cs:0.06,8192) -- (axis cs:140,8192);
        \addplot [mark=triangle,
            mark size=2pt,
            mark options={
                draw=black,
                fill=white,
            },
            nodes near coords={8TOP/s},
            every node near coord/.style={text=black, anchor=south east, font=\tiny},
            only marks,
            every mark/.append style={xshift=2, rotate=90},
        ]
        table {%
        140 8192
        };
        \node[right, font=\tiny, color=blue] at (axis description cs:0.60,0.4) {max. eff.=\%60.6}; 
        \node[right, font=\tiny, color=black!25!green] at (axis description cs:0.60,0.35) {max. eff.=\%56.2}; 
        \node[right, font=\tiny, color=red] at (axis description cs:0.60,0.30) {max. eff.=\%59.1}; 
                
        \node [text width=10em, anchor=north] at (axis description cs:0.5, 1) {\subcaption{\texttt{i8}\label{plot:matmul_i8_4x4_pareto}}};
    
    \end{groupplot}

\end{tikzpicture}
	
    \caption{
        Throughput versus compute workload Pareto frontiers for \texttt{bfloat16}, \texttt{i16} and \texttt{i8} matrix multiplications, with shapes swept up to $M=N=K=4k$, for AIE tile herd sized $4\times 4$.
        Output data width was kept consistent at 16 bits for all tests---\texttt{bfloat16} outputs for \texttt{bfloat16} inputs, and \texttt{i16} outputs for \texttt{i8} and \texttt{i16} inputs, respectively.
        The dotted line marks the theoretical peak compute throughput\protect\footnotemark[\value{footnote}] achievable for each herd of AIE cores at the specified precision.
    }
    \label{plot:matmul_pareto_groupplot}
\end{figure*}

Across all configurations, throughput is lower at small\gs{er} workloads (left side of each plot) due to underutilization of compute resources and startup latency at runtime.
As the compute workload increases, throughput rises and asymptotically approaches the device peak.
This indicates the transition from memory-bound throughput at small sizes to compute-bound throughput at larger sizes.
The shape of the performance curve confirms that \tech introduces minimal runtime overhead and supports efficient scaling into the compute-bound region.
\tech achieves up to 48.6\% of peak on the 4TOP/s herd, 65.0\% of peak on the 2TOP/s herd, and 48.2\% of peak on the 1TOP/s herd.

To evaluate the QoR achievable by \tech, we benchmark the performance of matrix multiplication workloads across three common AI data types supported by AMD NPU's vector engine: \texttt{i16}, \texttt{bf16}, and \texttt{i8}.
MLIR-AIE's hand-optimized implementations serve as baselines.
Figure~\ref{plot:matmul_pareto_groupplot} shows that in all three precision settings, \tech's compiler-generated designs achieve throughput closely tracking the Pareto frontier established by the manually optimized designs written in MLIR-AIE. 
This confirms \tech's effectiveness in generating near-optimal performance without requiring handcrafted code.

 
Designs generated by \tech achieve 78.7\%, 48.6\% and 59.1\% maximum efficiencies against the theoretical peak throughput for \texttt{i16}, \texttt{bf16} and \texttt{i8}, respectively, which fall within 5~pp from the MLIR-AIE hand-optimized designs.

We also compare \tech against ARIES, which demonstrates strong performance on smaller GEMM shapes, particularly for \texttt{i16} and \texttt{i8}. 
However, ARIES exhibits lower peak throughputs on these two data formats when saturated, indicating trade-offs between early-stage performance and scalability. 
These results further underscore \tech's ability to combine generality, analyzability, and performance within a unified spatial compilation flow.

\subsection{Kernel Merging: LLaMA 2 Multi-Head Attention}
\label{subsec:llama2}

\begin{figure}
    \centering
    \includegraphics[width=\linewidth]{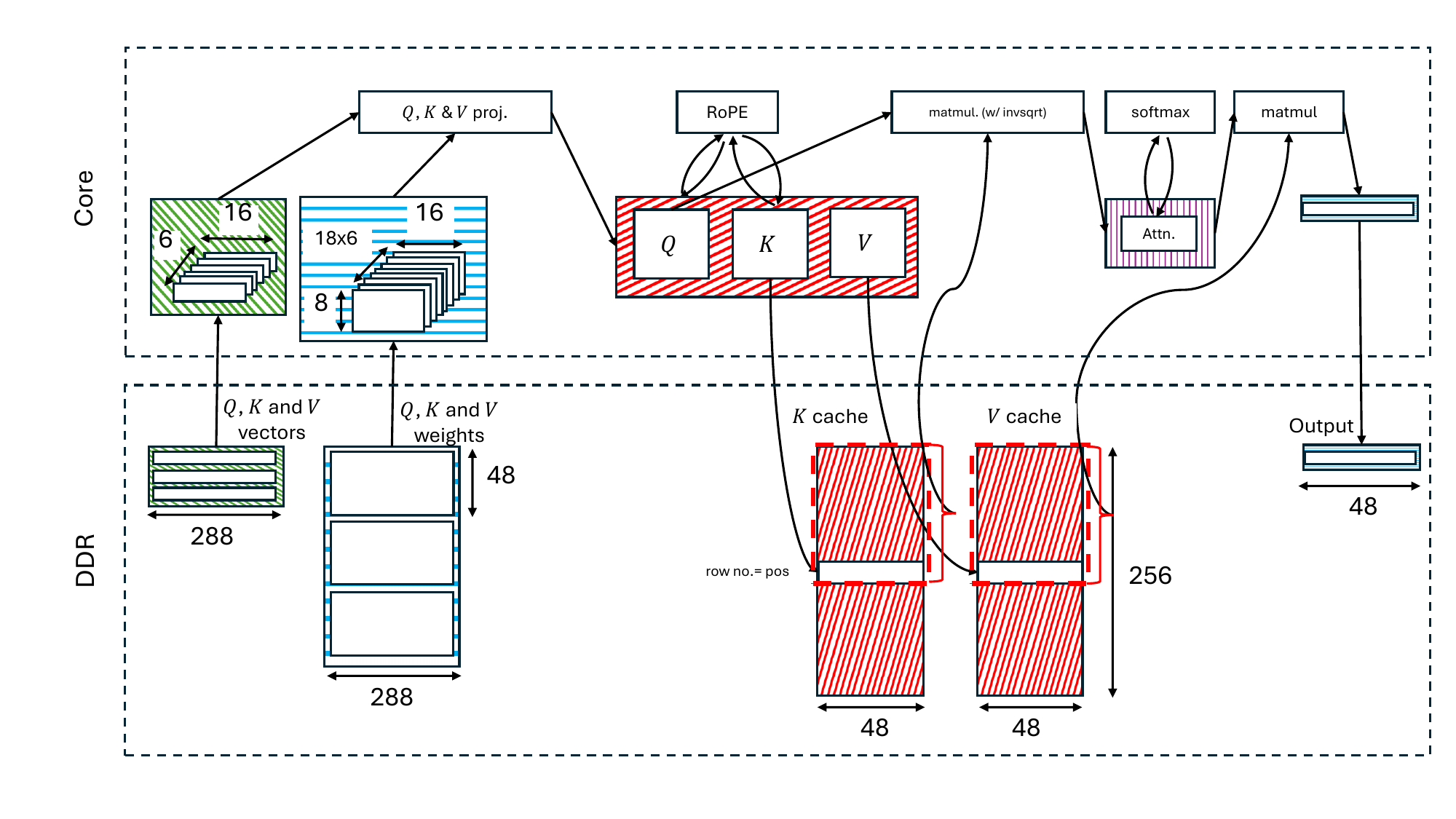}
    \caption{
        An overview of the fused LLaMA 2 MHA schedule.
        The dataflow through memory tile is omitted for simplicity.
    }
    \label{fig:MHA_OVERVLEW}
\end{figure}


To evaluate \tech's ability to express and optimize fused AI kernels, we implement a 
prototype of the LLaMA 2 MHA~\cite{LLAMA2} block on an AMD NPU using a single AIE core. 
The model uses a head size of 48 and a sequence length of 256, with 6 heads multiplexed in time. 

The MHA block includes projection ($Q$, $K$, $V$), rotary positional encoding (RoPE), softmax, and two matrix multiplications, separated by key-value ($KV$) caching~\cite{ROPE}. 
Each operation is implemented as a generic function, composed using structured \scffor and \scfpar loop nests. 
\tech compiles this into a hardware-aware schedule, by mapping operations to NPU components such DMA channels, BDs, and NPU compute tiles.

$KV$ caching is implemented as loop-nested \channel operations targeting persistent DDR memories, holding a cache size of $48\times 256$ data for $K$ and $V$, respectively.
DMA channel and BD reuse opportunities are captured via \channel merging described previously in Section~\ref{subsubsec:channel_fusion}.
Correctness in placement and buffer management is enforced via \tech's dependency analysis, which generates proper synchronization via \token values.

Table~\ref{tab:mha_fusion} profiles the end-to-end latency of each head, implemented with kernels dispatched both individually and fused together, with host and runtime dispatch overheads included.
When each component executes as an independent kernel, total latency is 834$\mu$s. 
With all kernels fused as one, latency is reduced to 373$\mu$s---achieving a 2.24$\times$ speedup by eliminating dispatch overheads, amortizing reconfiguration cost, and leveraging data locality within the NPU tile's local memory.

While this prototype does not yet exploit spatial parallelism across multiple AIE cores, it highlights \tech's ability to concisely represent and optimize non-trivial transformer blocks.
The full implementation is written in 155 lines of high-level MLIR, demonstrating the expressiveness of AIR abstractions for modeling modern AI computation and communication patterns.

\begin{table*}
    \centering
    \caption{
        LLaMA2 MHA evaluation.
        }
    \begin{threeparttable}
        \begin{tabular}{l S S S}
            \toprule
            Component & {Lines of code} & {Latency ($\mu$s)} & {Speedup} \\
            \midrule
            $Q$, $K$ and $V$ vector projection & 16 & 169 & {--} \\
            \texttt{RoPE} & 37 & 121 & {--} \\
            \texttt{matmul} (w/ \texttt{invsqrt}) & 39 & 283 & {--} \\
            \texttt{softmax} & 26 & 115 & {--} \\
            \texttt{matmul} & 37 & 146 & {--} \\
            \textbf{Total} & 155 & 834 & {--} \\
            \midrule
            \textbf{Fused} & 155 & 373 & {2.24$\times$} \\
            \bottomrule
        \end{tabular}
    \end{threeparttable}
    \label{tab:mha_fusion}
\end{table*}

\section{Future Directions}

We identify three key directions for extending \tech's applicability and automation in spatial compiler workflows:

\head{Multi-Target Hardware Support}
 \tech envisages support for multiple hardware platforms beyond NPUs, including compilation to GPUs using long-running persistent kernels and integration with user-developed accelerators implemented in FPGAs. This requires extending our backend abstractions and lowering pipelines to target architecture-specific runtimes, scheduling models and memory hierarchies. 

\head{Support for Heterogeneous Runtime Coordination}
As modern systems increasingly include CPUs, GPUs, and NPUs on a shared die, \tech can be a common abstraction supporting runtime coordination across heterogeneous devices. This includes lowering AIR to multiple backend runtimes (e.g., ROCr, XRT) and managing inter-accelerator data movement and synchronization.


\head{Cross-Device Launch Semantics}
Some \tech features such as data movement over explicit channels, and resource management using segments may have applicability in scaling beyond a single device, we plan to explore how an appropriate runtime might use \launch to enable multi-device dispatch. This includes scaling launches dynamically across available hardware based on runtime resource availability, allowing for coordinated execution across multiple accelerators on one host, and multiple hosts within a compute cluster. 

\section{Conclusion}

\tech~introduces a structured, extensible compiler abstraction for mapping high-level AI programs onto spatial architectures. 
By providing explicit constructs for asynchronous parallelism, data movement, and compute scheduling, AIR enables platform-agnostic, analyzable code generation without sacrificing performance. 
Our extensive evaluation demonstrates that \air provides both high expressiveness and efficiency, while maintaining a low abstraction overhead.  
We believe \tech provides a strong foundation for future spatial compiler infrastructures.


\bibliographystyle{ACM-Reference-Format}
\bibliography{bibliography}


\begin{thebibliography}{48}


\ifx \showCODEN    \undefined \def \showCODEN     #1{\unskip}     \fi
\ifx \showISBNx    \undefined \def \showISBNx     #1{\unskip}     \fi
\ifx \showISBNxiii \undefined \def \showISBNxiii  #1{\unskip}     \fi
\ifx \showISSN     \undefined \def \showISSN      #1{\unskip}     \fi
\ifx \showLCCN     \undefined \def \showLCCN      #1{\unskip}     \fi
\ifx \shownote     \undefined \def \shownote      #1{#1}          \fi
\ifx \showarticletitle \undefined \def \showarticletitle #1{#1}   \fi
\ifx \showURL      \undefined \def \showURL       {\relax}        \fi
\providecommand\bibfield[2]{#2}
\providecommand\bibinfo[2]{#2}
\providecommand\natexlab[1]{#1}
\providecommand\showeprint[2][]{arXiv:#2}

\bibitem[Abts et~al\mbox{.}(2022)]%
        {GROQCHIP}
\bibfield{author}{\bibinfo{person}{Dennis Abts}, \bibinfo{person}{John Kim}, \bibinfo{person}{Garrin Kimmell}, \bibinfo{person}{Matthew Boyd}, {et~al\mbox{.}}} \bibinfo{year}{2022}\natexlab{}.
\newblock \showarticletitle{{The Groq Software-defined Scale-out Tensor Streaming Multiprocessor : From chips-to-systems architectural overview}}. In \bibinfo{booktitle}{\emph{IEEE Hot Chips 34 Symposium (HCS)}}.
\newblock


\bibitem[Agostini et~al\mbox{.}(2022)]%
        {agostini2022mlir}
\bibfield{author}{\bibinfo{person}{Nicolas~Bohm Agostini}, \bibinfo{person}{Serena Curzel}, \bibinfo{person}{Vinay Amatya}, \bibinfo{person}{Cheng Tan}, {et~al\mbox{.}}} \bibinfo{year}{2022}\natexlab{}.
\newblock \showarticletitle{{An MLIR-based Compiler Flow for System-level Design and Hardware Acceleration}}. In \bibinfo{booktitle}{\emph{IEEE/ACM International Conference on Computer-Aided Design}}.
\newblock


\bibitem[Authors(2025)]%
        {PERFETTO}
\bibfield{author}{\bibinfo{person}{The~Chromium Authors}.} \bibinfo{year}{2025}\natexlab{}.
\newblock \bibinfo{booktitle}{\emph{{Perfetto}}}.
\newblock
\urldef\tempurl%
\url{https://perfetto.dev/docs/}
\showURL{%
\tempurl}


\bibitem[Authors(2019)]%
        {IREE}
\bibfield{author}{\bibinfo{person}{The~IREE Authors}.} \bibinfo{year}{2019}\natexlab{}.
\newblock \bibinfo{booktitle}{\emph{{IREE}}}.
\newblock
\urldef\tempurl%
\url{https://iree.dev/}
\showURL{%
\tempurl}


\bibitem[Baskaran et~al\mbox{.}(2008)]%
        {POLYOPT_GPGPU}
\bibfield{author}{\bibinfo{person}{Muthu~Manikandan Baskaran}, \bibinfo{person}{Uday Bondhugula}, \bibinfo{person}{Sriram Krishnamoorthy}, \bibinfo{person}{J. Ramanujam}, {et~al\mbox{.}}} \bibinfo{year}{2008}\natexlab{}.
\newblock \showarticletitle{{A Compiler Framework for Optimization of Affine Loop Nests for GPGPUs}}. In \bibinfo{booktitle}{\emph{International Conference on Supercomputing}}.
\newblock


\bibitem[Benabderrahmane et~al\mbox{.}(2010)]%
        {polyhedral_model}
\bibfield{author}{\bibinfo{person}{Mohamed-Walid Benabderrahmane}, \bibinfo{person}{Louis-No\"{e}l Pouchet}, \bibinfo{person}{Albert Cohen}, {and} \bibinfo{person}{C\'{e}dric Bastoul}.} \bibinfo{year}{2010}\natexlab{}.
\newblock \showarticletitle{{The Polyhedral Model is More Widely Applicable than You Think}}. In \bibinfo{booktitle}{\emph{Proceedings of the 19th Joint European Conference on Theory and Practice of Software, International Conference on Compiler Construction}}.
\newblock


\bibitem[Bondhugula et~al\mbox{.}(2008)]%
        {PLUTO}
\bibfield{author}{\bibinfo{person}{Uday Bondhugula}, \bibinfo{person}{Albert Hartono}, \bibinfo{person}{J. Ramanujam}, {and} \bibinfo{person}{P. Sadayappan}.} \bibinfo{year}{2008}\natexlab{}.
\newblock \showarticletitle{{A Practical Automatic Polyhedral Program Optimization System}}. In \bibinfo{booktitle}{\emph{ACM SIGPLAN Conference on Programming Language Design and Implementation (PLDI)}}.
\newblock


\bibitem[Cerebras(2025)]%
        {CEREBRAS_WAFER_SCALE_ENGINE}
\bibfield{author}{\bibinfo{person}{Cerebras}.} \bibinfo{year}{2025}\natexlab{}.
\newblock \bibinfo{booktitle}{\emph{{Cerebras Wafer Scale Engine}}}.
\newblock
\urldef\tempurl%
\url{https://www.cerebras.ai/chip}
\showURL{%
\tempurl}


\bibitem[Chandra et~al\mbox{.}(2001)]%
        {openmp}
\bibfield{author}{\bibinfo{person}{Rohit Chandra}, \bibinfo{person}{Leo Dagum}, \bibinfo{person}{David Kohr}, \bibinfo{person}{Ramesh Menon}, \bibinfo{person}{Dror Maydan}, {and} \bibinfo{person}{Jeff McDonald}.} \bibinfo{year}{2001}\natexlab{}.
\newblock \bibinfo{booktitle}{\emph{{Parallel programming in OpenMP}}}.
\newblock \bibinfo{publisher}{Morgan kaufmann}.
\newblock


\bibitem[Chatarasi et~al\mbox{.}(2021)]%
        {chatarasi2021marvel}
\bibfield{author}{\bibinfo{person}{Prasanth Chatarasi}, \bibinfo{person}{Hyoukjun Kwon}, \bibinfo{person}{Angshuman Parashar}, \bibinfo{person}{Michael Pellauer}, \bibinfo{person}{Tushar Krishna}, {and} \bibinfo{person}{Vivek Sarkar}.} \bibinfo{year}{2021}\natexlab{}.
\newblock \showarticletitle{{Marvel: A Data-centric Approach for Mapping Deep Learning Operators on Spatial Accelerators}}.
\newblock \bibinfo{journal}{\emph{ACM Transactions on Architecture and Code Optimization (TACO)}} \bibinfo{volume}{19}, \bibinfo{number}{1} (\bibinfo{year}{2021}), \bibinfo{pages}{1--26}.
\newblock


\bibitem[Cong et~al\mbox{.}(2011)]%
        {AUTOPILOT}
\bibfield{author}{\bibinfo{person}{Jason Cong}, \bibinfo{person}{Bin Liu}, \bibinfo{person}{Stephen Neuendorffer}, \bibinfo{person}{Juanjo Noguera}, \bibinfo{person}{Kees Vissers}, {and} \bibinfo{person}{Zhiru Zhang}.} \bibinfo{year}{2011}\natexlab{}.
\newblock \showarticletitle{{High-Level Synthesis for FPGAs: From Prototyping to Deployment}}.
\newblock \bibinfo{journal}{\emph{IEEE Transactions on Computer-Aided Design of Integrated Circuits and Systems}} \bibinfo{volume}{30}, \bibinfo{number}{4} (\bibinfo{year}{2011}), \bibinfo{pages}{473--491}.
\newblock


\bibitem[Corporation(2025)]%
        {TRITON_SHARED}
\bibfield{author}{\bibinfo{person}{Microsoft Corporation}.} \bibinfo{year}{2025}\natexlab{}.
\newblock \bibinfo{booktitle}{\emph{{Triton-shared}}}.
\newblock
\urldef\tempurl%
\url{https://github.com/microsoft/triton-shared}
\showURL{%
\tempurl}


\bibitem[Devices(2025a)]%
        {AMD_AI_ENGINE}
\bibfield{author}{\bibinfo{person}{Advanced~Micro Devices}.} \bibinfo{year}{2025}\natexlab{a}.
\newblock \bibinfo{booktitle}{\emph{{AI Engine}}}.
\newblock
\urldef\tempurl%
\url{https://www.amd.com/en/products/adaptive-socs-and-fpgas/technologies/ai-engine.html}
\showURL{%
\tempurl}


\bibitem[Devices(2025b)]%
        {MLIR_AIE}
\bibfield{author}{\bibinfo{person}{Advanced~Micro Devices}.} \bibinfo{year}{2025}\natexlab{b}.
\newblock \bibinfo{booktitle}{\emph{{MLIR-AIE}}}.
\newblock
\urldef\tempurl%
\url{https://xilinx.github.io/mlir-aie/}
\showURL{%
\tempurl}


\bibitem[Devices(2025c)]%
        {ROCR}
\bibfield{author}{\bibinfo{person}{Advanced~Micro Devices}.} \bibinfo{year}{2025}\natexlab{c}.
\newblock \bibinfo{booktitle}{\emph{{ROCr}}}.
\newblock
\urldef\tempurl%
\url{https://rocm.docs.amd.com/projects/ROCR-Runtime/en/latest/}
\showURL{%
\tempurl}


\bibitem[Devices(2025d)]%
        {AMD_RYZEN_7_7840U}
\bibfield{author}{\bibinfo{person}{Advanced~Micro Devices}.} \bibinfo{year}{2025}\natexlab{d}.
\newblock \bibinfo{booktitle}{\emph{{Ryzen 7 7840U}}}.
\newblock
\urldef\tempurl%
\url{https://www.amd.com/en/products/processors/laptop/ryzen/7000-series/amd-ryzen-7-7840u.html}
\showURL{%
\tempurl}


\bibitem[Devices(2025e)]%
        {XRT}
\bibfield{author}{\bibinfo{person}{Advanced~Micro Devices}.} \bibinfo{year}{2025}\natexlab{e}.
\newblock \bibinfo{booktitle}{\emph{{XRT}}}.
\newblock
\urldef\tempurl%
\url{https://xilinx.github.io/XRT/master/html/index.html}
\showURL{%
\tempurl}


\bibitem[Elango et~al\mbox{.}(2018)]%
        {DIESEL_POLYHEDRAL}
\bibfield{author}{\bibinfo{person}{Venmugil Elango}, \bibinfo{person}{Norm Rubin}, \bibinfo{person}{Mahesh Ravishankar}, \bibinfo{person}{Hariharan Sandanagobalane}, {and} \bibinfo{person}{Vinod Grover}.} \bibinfo{year}{2018}\natexlab{}.
\newblock \showarticletitle{{Diesel: DSL for Linear Algebra and Neural Net Computations on GPUs}}. In \bibinfo{booktitle}{\emph{ACM SIGPLAN International Workshop on Machine Learning and Programming Languages}}.
\newblock


\bibitem[Hariprasad et~al\mbox{.}(2017)]%
        {halstead_metrics}
\bibfield{author}{\bibinfo{person}{T Hariprasad}, \bibinfo{person}{G Vidhyagaran}, \bibinfo{person}{K Seenu}, {and} \bibinfo{person}{Chandrasegar Thirumalai}.} \bibinfo{year}{2017}\natexlab{}.
\newblock \showarticletitle{{Software Complexity Analysis using Halstead Metrics}}. In \bibinfo{booktitle}{\emph{International Conference on Trends in Electronics and Informatics (ICEI)}}.
\newblock


\bibitem[Hunhoff et~al\mbox{.}(2025)]%
        {IRON_FCCM}
\bibfield{author}{\bibinfo{person}{Erika Hunhoff}, \bibinfo{person}{Joseph Melber}, \bibinfo{person}{Kristof Denolf}, \bibinfo{person}{Andra Bisca}, \bibinfo{person}{Samuel Bayliss}, \bibinfo{person}{Stephen Neuendorffer}, \bibinfo{person}{Jeff Fifield}, \bibinfo{person}{Jack Lo}, \bibinfo{person}{Pranathi Vasireddy}, \bibinfo{person}{Phil James-Roxby}, {and} \bibinfo{person}{Eric Keller}.} \bibinfo{year}{2025}\natexlab{}.
\newblock \showarticletitle{{ Efficiency, Expressivity, and Extensibility in a Close-to-Metal NPU Programming Interface }}. In \bibinfo{booktitle}{\emph{IEEE International Symposium on Field-Programmable Custom Computing Machines (FCCM)}}.
\newblock


\bibitem[INTEL(2025)]%
        {INTEL_NPU}
\bibfield{author}{\bibinfo{person}{INTEL}.} \bibinfo{year}{2025}\natexlab{}.
\newblock \bibinfo{booktitle}{\emph{{Intel NPU}}}.
\newblock
\urldef\tempurl%
\url{https://edc.intel.com/content/www/us/en/design/products/platforms/details/arrow-lake-s/core-ultra-200s-series-processors-datasheet-volume-1-of-2/intel-neural-processing-unit-intel-npu/}
\showURL{%
\tempurl}


\bibitem[Jeong et~al\mbox{.}(2021)]%
        {jeong2021union}
\bibfield{author}{\bibinfo{person}{Geonhwa Jeong}, \bibinfo{person}{Gokcen Kestor}, \bibinfo{person}{Prasanth Chatarasi}, \bibinfo{person}{Angshuman Parashar}, \bibinfo{person}{Po-An Tsai}, \bibinfo{person}{Sivasankaran Rajamanickam}, \bibinfo{person}{Roberto Gioiosa}, {and} \bibinfo{person}{Tushar Krishna}.} \bibinfo{year}{2021}\natexlab{}.
\newblock \showarticletitle{{Union: A Unified HW-SW Co-design Ecosystem in MLIR for Evaluating Tensor Operations on Spatial Accelerators}}. In \bibinfo{booktitle}{\emph{International Conference on Parallel Architectures and Compilation Techniques (PACT)}}.
\newblock


\bibitem[Jouppi et~al\mbox{.}(2023)]%
        {GOOGLE_TPU_V4}
\bibfield{author}{\bibinfo{person}{Norm Jouppi}, \bibinfo{person}{George Kurian}, \bibinfo{person}{Sheng Li}, \bibinfo{person}{Peter Ma}, {et~al\mbox{.}}} \bibinfo{year}{2023}\natexlab{}.
\newblock \showarticletitle{{TPU v4: An Optically Reconfigurable Supercomputer for Machine Learning with Hardware Support for Embeddings}}. In \bibinfo{booktitle}{\emph{Annual International Symposium on Computer Architecture}}.
\newblock


\bibitem[Lacchia(2025)]%
        {RADON}
\bibfield{author}{\bibinfo{person}{Michele Lacchia}.} \bibinfo{year}{2025}\natexlab{}.
\newblock \bibinfo{booktitle}{\emph{{Radon}}}.
\newblock
\urldef\tempurl%
\url{https://github.com/rubik/radon/tree/master}
\showURL{%
\tempurl}


\bibitem[Lattner et~al\mbox{.}(2021)]%
        {mlir}
\bibfield{author}{\bibinfo{person}{Chris Lattner}, \bibinfo{person}{Mehdi Amini}, \bibinfo{person}{Uday Bondhugula}, \bibinfo{person}{Albert Cohen}, \bibinfo{person}{Andy Davis}, \bibinfo{person}{Jacques Pienaar}, \bibinfo{person}{River Riddle}, \bibinfo{person}{Tatiana Shpeisman}, \bibinfo{person}{Nicolas Vasilache}, {and} \bibinfo{person}{Oleksandr Zinenko}.} \bibinfo{year}{2021}\natexlab{}.
\newblock \showarticletitle{{{MLIR}}: Scaling Compiler Infrastructure for Domain Specific Computation}. In \bibinfo{booktitle}{\emph{{{IEEE/ACM}} International Symposium on Code Generation and Optimization (CGO)}}.
\newblock


\bibitem[Lattner and Pienaar(2019)]%
        {lattner2020mlircompilerinfrastructureend}
\bibfield{author}{\bibinfo{person}{Chris Lattner} {and} \bibinfo{person}{Jacques Pienaar}.} \bibinfo{year}{2019}\natexlab{}.
\newblock \bibinfo{title}{MLIR Primer: A Compiler Infrastructure for the End of Moore’s Law}.
\newblock


\bibitem[Moses et~al\mbox{.}(2021)]%
        {POLYGEIST}
\bibfield{author}{\bibinfo{person}{William~S. Moses}, \bibinfo{person}{Lorenzo Chelini}, \bibinfo{person}{Ruizhe Zhao}, {and} \bibinfo{person}{Oleksandr Zinenko}.} \bibinfo{year}{2021}\natexlab{}.
\newblock \showarticletitle{{Polygeist: Raising C to Polyhedral MLIR}}. In \bibinfo{booktitle}{\emph{International Conference on Parallel Architectures and Compilation Techniques (PACT)}}.
\newblock


\bibitem[Mutlu et~al\mbox{.}(2022)]%
        {comet}
\bibfield{author}{\bibinfo{person}{Erdal Mutlu}, \bibinfo{person}{Ruiqin Tian}, \bibinfo{person}{Bin Ren}, \bibinfo{person}{Sriram Krishnamoorthy}, \bibinfo{person}{Roberto Gioiosa}, \bibinfo{person}{Jacques Pienaar}, {and} \bibinfo{person}{Gokcen" Kestor}.} \bibinfo{year}{2022}\natexlab{}.
\newblock \showarticletitle{{COMET: A Domain-Specific Compilation of High-Performance Computational Chemistry}}. In \bibinfo{booktitle}{\emph{Languages and Compilers for Parallel Computing}}.
\newblock


\bibitem[NVIDIA(2020)]%
        {cuda}
\bibfield{author}{\bibinfo{person}{NVIDIA}.} \bibinfo{year}{2020}\natexlab{}.
\newblock \bibinfo{title}{CUDA, release: 10.2.89}.
\newblock
\urldef\tempurl%
\url{https://developer.nvidia.com/cuda-toolkit}
\showURL{%
\tempurl}


\bibitem[OpenAI(2025)]%
        {TRITON}
\bibfield{author}{\bibinfo{person}{OpenAI}.} \bibinfo{year}{2025}\natexlab{}.
\newblock \bibinfo{booktitle}{\emph{{TRITON}}}.
\newblock
\urldef\tempurl%
\url{https://triton-lang.org/main/index.html}
\showURL{%
\tempurl}


\bibitem[Prabhakar et~al\mbox{.}(2024)]%
        {SAMBANOVA}
\bibfield{author}{\bibinfo{person}{Raghu Prabhakar}, \bibinfo{person}{Ram Sivaramakrishnan}, \bibinfo{person}{Darshan Gandhi}, \bibinfo{person}{Yun Du}, {et~al\mbox{.}}} \bibinfo{year}{2024}\natexlab{}.
\newblock \showarticletitle{{ SambaNova SN40L: Scaling the AI Memory Wall with Dataflow and Composition of Experts }}. In \bibinfo{booktitle}{\emph{IEEE/ACM International Symposium on Microarchitecture (MICRO)}}.
\newblock


\bibitem[Qualcomm(2025)]%
        {QUALCOMM_AI_ENGINE}
\bibfield{author}{\bibinfo{person}{Qualcomm}.} \bibinfo{year}{2025}\natexlab{}.
\newblock \bibinfo{booktitle}{\emph{{Qualcomm AI Engine}}}.
\newblock
\urldef\tempurl%
\url{https://www.qualcomm.com/products/technology/processors/ai-engine}
\showURL{%
\tempurl}


\bibitem[Rico et~al\mbox{.}(2024)]%
        {RYZEN_XDNA}
\bibfield{author}{\bibinfo{person}{Alejandro Rico}, \bibinfo{person}{Satyaprakash Pareek}, \bibinfo{person}{Javier Cabezas}, \bibinfo{person}{David Clarke}, {et~al\mbox{.}}} \bibinfo{year}{2024}\natexlab{}.
\newblock \showarticletitle{{AMD} {XDNA™ NPU} in {Ryzen™ AI} Processors}.
\newblock \bibinfo{journal}{\emph{IEEE Micro}} \bibinfo{volume}{44}, \bibinfo{number}{6} (\bibinfo{year}{2024}), \bibinfo{pages}{73--82}.
\newblock


\bibitem[R\"osti and Franz(2025)]%
        {NPU_ML_TRAINING}
\bibfield{author}{\bibinfo{person}{Andr\'e R\"osti} {and} \bibinfo{person}{Michael Franz}.} \bibinfo{year}{2025}\natexlab{}.
\newblock \showarticletitle{{Unlocking the AMD Neural Processing Unit for ML Training on the Client Using Bare-Metal-Programming Tools}}. In \bibinfo{booktitle}{\emph{IEEE International Symposium on Field-Programmable Custom Computing Machines (FCCM)}}.
\newblock


\bibitem[Rucker et~al\mbox{.}(2024)]%
        {rucker2024revet}
\bibfield{author}{\bibinfo{person}{Alexander~C Rucker}, \bibinfo{person}{Shiv Sundram}, \bibinfo{person}{Coleman Smith}, \bibinfo{person}{Matthew Vilim}, \bibinfo{person}{Raghu Prabhakar}, \bibinfo{person}{Fredrik Kj{\o}lstad}, {and} \bibinfo{person}{Kunle Olukotun}.} \bibinfo{year}{2024}\natexlab{}.
\newblock \showarticletitle{{Revet: A Language and Compiler for Dataflow Threads}}. In \bibinfo{booktitle}{\emph{International Symposium on High-Performance Computer Architecture (HPCA)}}.
\newblock


\bibitem[Sabne(2020)]%
        {sabne2020xla}
\bibfield{author}{\bibinfo{person}{Amit Sabne}.} \bibinfo{year}{2020}\natexlab{}.
\newblock \bibinfo{title}{{XLA: Compiling Machine Learning for Peak Performance}}.
\newblock \bibinfo{howpublished}{\url{https://research.google/pubs/xla-compiling-machine-learning-for-peak-performance/}}.
\newblock


\bibitem[Singh(2022)]%
        {singh2022designing}
\bibfield{author}{\bibinfo{person}{Gagandeep Singh}.} \bibinfo{year}{2022}\natexlab{}.
\newblock \showarticletitle{Designing, modeling, and optimizing data-intensive computing systems}.
\newblock \bibinfo{journal}{\emph{arXiv preprint arXiv:2208.08886}} (\bibinfo{year}{2022}).
\newblock


\bibitem[Singh et~al\mbox{.}(2021)]%
        {singh2021fpga}
\bibfield{author}{\bibinfo{person}{Gagandeep Singh}, \bibinfo{person}{Mohammed Alser}, \bibinfo{person}{Damla~Senol Cali}, \bibinfo{person}{Dionysios Diamantopoulos}, \bibinfo{person}{Juan G{\'o}mez-Luna}, \bibinfo{person}{Henk Corporaal}, {and} \bibinfo{person}{Onur Mutlu}.} \bibinfo{year}{2021}\natexlab{}.
\newblock \showarticletitle{FPGA-based near-memory acceleration of modern data-intensive applications}.
\newblock \bibinfo{journal}{\emph{IEEE Micro}} \bibinfo{volume}{41}, \bibinfo{number}{4} (\bibinfo{year}{2021}), \bibinfo{pages}{39--48}.
\newblock


\bibitem[Singh et~al\mbox{.}(2024)]%
        {singh2024rubicon}
\bibfield{author}{\bibinfo{person}{Gagandeep Singh}, \bibinfo{person}{Mohammed Alser}, \bibinfo{person}{Kristof Denolf}, \bibinfo{person}{Can Firtina}, \bibinfo{person}{Alireza Khodamoradi}, \bibinfo{person}{Meryem~Banu Cavlak}, \bibinfo{person}{Henk Corporaal}, {and} \bibinfo{person}{Onur Mutlu}.} \bibinfo{year}{2024}\natexlab{}.
\newblock \showarticletitle{RUBICON: a framework for designing efficient deep learning-based genomic basecallers}.
\newblock \bibinfo{journal}{\emph{Genome Biology}} \bibinfo{volume}{25}, \bibinfo{number}{1} (\bibinfo{year}{2024}), \bibinfo{pages}{49}.
\newblock


\bibitem[Singh et~al\mbox{.}(2020)]%
        {singh2020nero}
\bibfield{author}{\bibinfo{person}{Gagandeep Singh}, \bibinfo{person}{Dionysios Diamantopoulos}, \bibinfo{person}{Christoph Hagleitner}, \bibinfo{person}{Juan Gomez-Luna}, \bibinfo{person}{Sander Stuijk}, \bibinfo{person}{Onur Mutlu}, {and} \bibinfo{person}{Henk Corporaal}.} \bibinfo{year}{2020}\natexlab{}.
\newblock \showarticletitle{NERO: A near high-bandwidth memory stencil accelerator for weather prediction modeling}. In \bibinfo{booktitle}{\emph{2020 30th International Conference on Field-Programmable Logic and Applications (FPL)}}. IEEE, \bibinfo{pages}{9--17}.
\newblock


\bibitem[Singh et~al\mbox{.}(2023)]%
        {singh2023sparta}
\bibfield{author}{\bibinfo{person}{Gagandeep Singh}, \bibinfo{person}{Alireza Khodamoradi}, \bibinfo{person}{Kristof Denolf}, \bibinfo{person}{Jack Lo}, \bibinfo{person}{Juan Gomez-Luna}, \bibinfo{person}{Joseph Melber}, \bibinfo{person}{Andra Bisca}, \bibinfo{person}{Henk Corporaal}, {and} \bibinfo{person}{Onur Mutlu}.} \bibinfo{year}{2023}\natexlab{}.
\newblock \showarticletitle{{SPARTA: Spatial Acceleration for Efficient and Scalable Horizontal Diffusion Weather Stencil Computation}}. In \bibinfo{booktitle}{\emph{Proceedings of the 37th International Conference on Supercomputing}}. \bibinfo{pages}{463--476}.
\newblock


\bibitem[Singha et~al\mbox{.}(2022)]%
        {singha2022leaper}
\bibfield{author}{\bibinfo{person}{Gagandeep Singha}, \bibinfo{person}{Dionysios Diamantopoulosb}, \bibinfo{person}{Juan G{\'o}mez-Lunaa}, \bibinfo{person}{Sander Stuijkc}, \bibinfo{person}{Henk Corporaalc}, {and} \bibinfo{person}{Onur Mutlua}.} \bibinfo{year}{2022}\natexlab{}.
\newblock \showarticletitle{LEAPER: Fast and accurate FPGA-based system performance prediction via transfer learning}. In \bibinfo{booktitle}{\emph{2022 IEEE 40th International Conference on Computer Design (ICCD)}}. IEEE, \bibinfo{pages}{499--508}.
\newblock


\bibitem[Su et~al\mbox{.}(2024)]%
        {ROPE}
\bibfield{author}{\bibinfo{person}{Jianlin Su}, \bibinfo{person}{Murtadha Ahmed}, \bibinfo{person}{Yu Lu}, \bibinfo{person}{Shengfeng Pan}, \bibinfo{person}{Wen Bo}, {and} \bibinfo{person}{Yunfeng Liu}.} \bibinfo{year}{2024}\natexlab{}.
\newblock \showarticletitle{{RoFormer: Enhanced Transformer with Rotary Position Embedding}}.
\newblock \bibinfo{journal}{\emph{Neurocomput.}} \bibinfo{volume}{568}, \bibinfo{number}{C} (\bibinfo{year}{2024}).
\newblock


\bibitem[Touvron et~al\mbox{.}(2023)]%
        {LLAMA2}
\bibfield{author}{\bibinfo{person}{Hugo Touvron}, \bibinfo{person}{Louis Martin}, \bibinfo{person}{Kevin Stone}, \bibinfo{person}{Peter Albert}, {et~al\mbox{.}}} \bibinfo{year}{2023}\natexlab{}.
\newblock \bibinfo{title}{{Llama 2: Open Foundation and Fine-Tuned Chat Models}}.
\newblock
\showeprint[arxiv]{2307.09288}~[cs.CL]
\urldef\tempurl%
\url{https://arxiv.org/abs/2307.09288}
\showURL{%
\tempurl}


\bibitem[Wang et~al\mbox{.}(2021)]%
        {AUTOSA}
\bibfield{author}{\bibinfo{person}{Jie Wang}, \bibinfo{person}{Licheng Guo}, {and} \bibinfo{person}{Jason Cong}.} \bibinfo{year}{2021}\natexlab{}.
\newblock \showarticletitle{{AutoSA: A Polyhedral Compiler for High-Performance Systolic Arrays on FPGA}}. In \bibinfo{booktitle}{\emph{ACM/SIGDA International Symposium on Field-Programmable Gate Arrays}}.
\newblock


\bibitem[Ye et~al\mbox{.}(2024)]%
        {ye2024hida}
\bibfield{author}{\bibinfo{person}{Hanchen Ye}, \bibinfo{person}{Hyegang Jun}, {and} \bibinfo{person}{Deming Chen}.} \bibinfo{year}{2024}\natexlab{}.
\newblock \showarticletitle{{HIDA: A Hierarchical Dataflow Compiler for High-level Synthesis}}. In \bibinfo{booktitle}{\emph{ACM International Conference on Architectural Support for Programming Languages and Operating Systems}}.
\newblock


\bibitem[Zheng et~al\mbox{.}(2022)]%
        {zheng2022amos}
\bibfield{author}{\bibinfo{person}{Size Zheng}, \bibinfo{person}{Renze Chen}, \bibinfo{person}{Anjiang Wei}, \bibinfo{person}{Yicheng Jin}, \bibinfo{person}{Qin Han}, \bibinfo{person}{Liqiang Lu}, \bibinfo{person}{Bingyang Wu}, \bibinfo{person}{Xiuhong Li}, \bibinfo{person}{Shengen Yan}, {and} \bibinfo{person}{Yun Liang}.} \bibinfo{year}{2022}\natexlab{}.
\newblock \showarticletitle{{AMOS: Enabling Automatic Mapping for Tensor Computations on Spatial Accelerators with Hardware Abstraction}}. In \bibinfo{booktitle}{\emph{Annual International Symposium on Computer Architecture}}.
\newblock


\bibitem[Zhuang et~al\mbox{.}(2025)]%
        {zhuang2025aries}
\bibfield{author}{\bibinfo{person}{Jinming Zhuang}, \bibinfo{person}{Shaojie Xiang}, \bibinfo{person}{Hongzheng Chen}, \bibinfo{person}{Niansong Zhang}, \bibinfo{person}{Zhuoping Yang}, \bibinfo{person}{Tony Mao}, \bibinfo{person}{Zhiru Zhang}, {and} \bibinfo{person}{Peipei Zhou}.} \bibinfo{year}{2025}\natexlab{}.
\newblock \showarticletitle{{ARIES: An Agile MLIR-Based Compilation Flow for Reconfigurable Devices with AI Engines}}. In \bibinfo{booktitle}{\emph{ACM/SIGDA International Symposium on Field Programmable Gate Arrays}}.
\newblock


\end{thebibliography}

\newpage
\appendix

\section{Input IR to \tech's Vector-add Example}
\label{appendix:vecadd_expanded}

\begin{minipage}{\textwidth}
\begin{lstlisting}[language=mlir, caption={Element-wise vector add, described in upstream MLIR dialects.}, label={lst:vecadd_input}, basicstyle=\ttfamily\tiny, xleftmargin=1em, xrightmargin=2em] 
func.func @eltwise_add(%arg0: memref<65536xf32>, %arg1: memref<65536xf32>, %arg2: memref<65536xf32>) {
  %c65536 = arith.constant 65536 : index
  %c2048 = arith.constant 2048 : index
  %c1024 = arith.constant 1024 : index
  %c1 = arith.constant 1 : index
  %c2 = arith.constant 2 : index
  %c0 = arith.constant 0 : index
  scf.parallel (%arg3) = (%c0) to (%c2) step (%c1) {
    %alloc = memref.alloc() : memref<1024xf32, 2>
    %alloc_0 = memref.alloc() : memref<1024xf32, 2>
    %alloc_1 = memref.alloc() : memref<1024xf32, 2>
    scf.for %arg4 = %c0 to %c65536 step %c2048 {
      %subview = memref.subview %arg0[0] [1024] [1] : memref<65536xf32> to memref<1024xf32>
      memref.copy %subview, %alloc : memref<1024xf32> to memref<1024xf32, 2>
      %subview_2 = memref.subview %arg1[0] [1024] [1] : memref<65536xf32> to memref<1024xf32>
      memref.copy %subview_2, %alloc_0 : memref<1024xf32> to memref<1024xf32, 2>
      scf.for %arg5 = %c0 to %c1024 step %c1 {
        %0 = memref.load %alloc[%arg5] : memref<1024xf32, 2>
        %1 = memref.load %alloc_0[%arg5] : memref<1024xf32, 2>
        %2 = arith.addf %0, %1 : f32
        memref.store %2, %alloc_1[%arg5] : memref<1024xf32, 2>
      }
      %subview_3 = memref.subview %arg2[0] [1024] [1] : memref<65536xf32> to memref<1024xf32>
      memref.copy %alloc_1, %subview_3 : memref<1024xf32, 2> to memref<1024xf32>
      memref.dealloc %alloc : memref<1024xf32, 2>
      memref.dealloc %alloc_0 : memref<1024xf32, 2>
      memref.dealloc %alloc_1 : memref<1024xf32, 2>
    }
    scf.reduce 
  }
  return
}
\end{lstlisting}
\end{minipage}

\newpage
\section{\air Python Bindings to \tech's Vector-add Example}
\label{appendix:vecadd_pybind}

\begin{minipage}{\textwidth}
\begin{lstlisting}[language=python, caption={Element-wise vector add, described in \air's Python bindings.}, label={lst:vecadd_pybind}, basicstyle=\ttfamily\tiny, xleftmargin=1em, xrightmargin=2em] 
@module_builder
def build_module(n, tile_n, np_dtype_in):
    a_size = [n]
    b_size = a_size
    out_size = a_size
    xrt_dtype_in = type_mapper(np_dtype_in)
    num_tiles = 2
    assert n % (tile_n * num_tiles) == 0
    # L3 MemRefTypes
    l3memrefTy = MemRefType.get(a_size, xrt_dtype_in)
    # L1 MemRefTypes
    l1MemrefTy = MemRefType.get(
        shape=[tile_n],
        element_type=xrt_dtype_in,
        memory_space=IntegerAttr.get(T.i32(), MemorySpace.L1),
    )
    @FuncOp.from_py_func(l3memrefTy, l3memrefTy, l3memrefTy)
    def eltwise_add(arg0, arg1, arg2):
        @herd(
            name="herd_0",
            sizes=[1, num_tiles],
            operands=[arg0, arg1, arg2],
        )
        def herd_body(_tx, _ty, _sx, _sy, _l3_a, _l3_b, _l3_c):
            l1_a_data = AllocOp(l1MemrefTy, [], [])
            l1_b_data = AllocOp(l1MemrefTy, [], [])
            l1_out_data = AllocOp(l1MemrefTy, [], [])
            for _l_ivx in range_(0, n, tile_n * num_tiles):
                offset_map = AffineMap.get(0, 2,
                    [
                        AffineExpr.get_add(
                            AffineSymbolExpr.get(0),
                            AffineExpr.get_mul(
                                AffineSymbolExpr.get(1),
                                AffineConstantExpr.get(tile_n),
                            ),
                        )
                    ],
                )
                offset = affine_apply(offset_map, [_l_ivx, _ty])
                dma_memcpy_nd(l1_a_data, _l3_a,
                    src_offsets=[offset],
                    src_sizes=[tile_n],
                    src_strides=[1],
                )
                dma_memcpy_nd(l1_b_data, _l3_b,
                    src_offsets=[offset],
                    src_sizes=[tile_n],
                    src_strides=[1],
                )
                for i in range_(tile_n):
                    val_a = load(l1_a_data, [i])
                    val_b = load(l1_b_data, [i])
                    val_out = arith.addf(val_a, val_b)
                    store(val_out, l1_out_data, [i])
                    yield_([])
                dma_memcpy_nd(_l3_c, l1_out_data,
                    dst_offsets=[offset],
                    dst_sizes=[tile_n],
                    dst_strides=[1],
                )
                DeallocOp(l1_a_data)
                DeallocOp(l1_b_data)
                DeallocOp(l1_out_data)
                yield_([])
\end{lstlisting}
\end{minipage}

\end{document}